# Autonomous Chemical Mechanistic Discovery through Agentic Reasoning and Validation

Dong Li[1,2], Sixuan Mi[1,4], Zihao Ye[1,5], Huan Xiong[2], Tao XU[7], Tong Zhu[4], Aijia Zhang[3], Junqi Gao[3], Kaiyan Zhang[6], Shijie Wang[1], Bowen Zhou[1,6*], Yuqiang Li[1*], Biqing Qi[1*]

[1]Shanghai Artificial Intelligence Laboratory, Shanghai, China.

[2]Institute for Advanced Study in Mathematics, Harbin Institute of Technology, Harbin, China.

[3]School of Mathematics, Harbin Institute of Technology, Harbin, China.

[4]School of Chemistry and Molecular Engineering, East China Normal University, Shanghai, China.

[5]Department of Chemistry, Fudan University, Shanghai, China.

[6]Department of Electronic Engineering, Tsinghua University, Beijing, China.

[7]School of Chemical Science and Engineering, Tongji University, Shanghai, China.

[*]Corresponding author(s): zhoubowen@tsinghua.edu.cn; liyuqiang@pjlab.org.cn; qibiqing@pjlab.org.cn;

## Abstract

**Unraveling reaction mechanisms is central to modern chemistry, yet automating these investigations remains challenging because computational workflows still rely heavily on expert intervention. Here we introduce ARCHE, an autonomous agentic system that integrates a general-purpose reasoning model, a domain-specialized computational chemistry model, and a structured tool registry to transform mechanistic inquiry into a scalable, self-validating process. ARCHE interprets scientific questions, generates and prioritizes mechanistic hypotheses, orchestrates computational workflows, and iteratively refines conclusions based on computed evidence within a closed loop. We validate its capabilities across three increasingly demanding scenarios: reconstructing stereocontrolling transition states and validating the corresponding reaction mechanism in a previously reported asymmetric catalytic reaction; proposing and validating a plausible radical pathway through iterative hypothesis refinement for a recently discovered but unpublished $\alpha$-iodoboronate C–I cleavage reaction; and identifying a chemically interpretable descriptor that governs selectivity in nickel-catalysed migratory cross-coupling reactions. By coupling agentic reasoning with rigorous computational validation, ARCHE advances autonomous mechanistic discovery and establishes a foundation for broader machine-assisted chemical research. The code for ARCHE is publicly available at `https://github.com/JetAstra/Arche-Harness`.**

# Introduction

The core mission of chemical science is to uncover the fundamental principles governing matter transformation, with the elucidation of reaction mechanisms serving as the pivotal endeavor to achieve this goal (*1–3*). Mechanistic insights drive the field from merely observing phenomena to precisely controlling processes, underpinning rational catalyst design, stereochemical control, and the interpretation of experimental reactivity (*4–6*). Despite significant advances in computational chemistry software and automated workflows, transforming experimental observations into robust mechanistic insights remains a formidable challenge. Existing tools can automate predefined computational procedures, including error handling and restart routines, but they fall short of the closed-loop process required for autonomous discovery, in which mechanistic hypotheses are generated, computationally validated and iteratively revised on the basis of the resulting evidence (*7–10*).

This process traditionally relies on the intuition of computational chemists to propose plausible pathways based on experimental observations, involves laborious calculation procedures, demands rigorous analysis of output data, and frequently requires diagnosing failures and refining strategies when necessary (*11–13*). Recent advances in Large Language Models (LLMs) have enabled agentic systems capable of coordinating scientific reasoning with external computational tools (*14–17*). In chemistry, emerging frameworks show that language-driven agents can automate predefined workflows for well-specified computational tasks, with error handling and restart mechanisms (*18–21*). These efforts establish that generalist LLM-based agents can translate well-defined objectives into executable computational task loops. However, their domain knowledge is typically injected via prompting, retrieval, or rule-based modules, which confines them to executing fixed protocols rather than engaging in open-ended scientific reasoning (*22–24*).

Consequently, the autonomous exploration of unresolved reaction mechanisms remains highly constrained. First, existing systems are organized around predefined workflows or isolated computational stages. When confronted with open-ended mechanistic questions, they struggle to actively generate multiple competing hypotheses, design discriminative calculations, and iteratively refine conclusions (*25,26*). Second, computational workflows require consistent specification of methods, basis sets, solvation models and convergence criteria, together with checks that the resulting inputs and outputs satisfy the requirements of the calculation. Integrating these checks with hypothe-sis generation and workflow execution remains a central challenge for autonomous mechanistic investigation (*27–29*). Bridging this gap requires an integrated framework that unifies agentic reasoning, domain-specialized computational expertise, and reliable tool orchestration to enable truly autonomous, self-validating mechanistic inquiry.

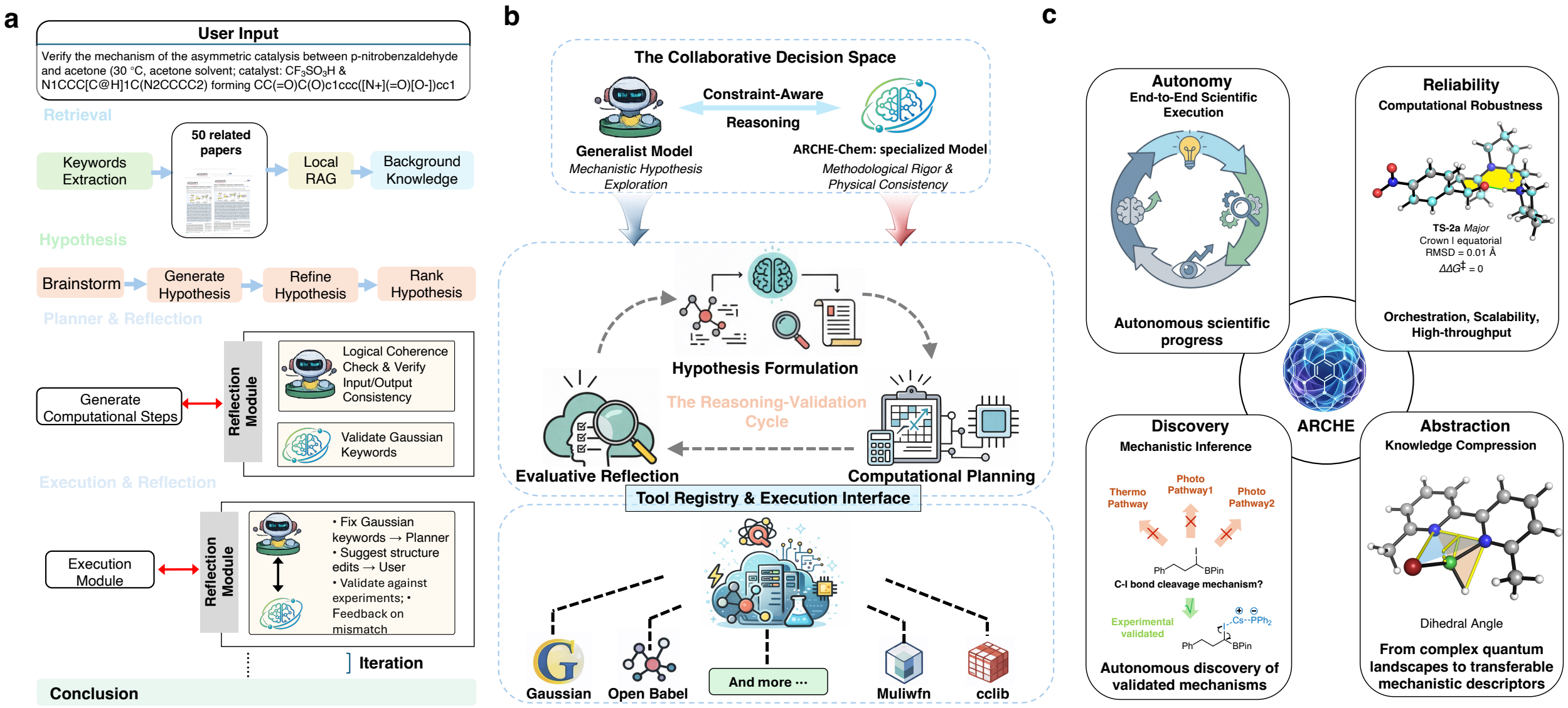

**Figure 1**: **The ARCHE framework for autonomous mechanistic discovery in computational chemistry. a,** From a scientific question, ARCHE retrieves relevant knowledge, generates and ranks mechanistic hypotheses, and translates prioritized hypotheses into executable quantum-chemical workflows. Reflection modules verify logical coherence, validate Gaussian inputs and analyse computational outputs, enabling iterative refinement toward a consistent mechanistic conclusion. **b,** A unified decision space couples a generalist reasoning model for mechanistic exploration with ARCHE-Chem for Gaussian input refinement, output interpretation and error diagnosis. These components interact through hypothesis formulation, computational planning and evaluative reflection, and are connected to external tools through a structured tool registry and execution interface. **c,** ARCHE enables reliable orchestration of quantum-chemical computation steps, autonomous mechanistic discovery through hypothesis competition, and abstraction of transferable mechanistic descriptors from computed reaction landscapes.

Here, we present **ARCHE**, an autonomous agentic system an autonomous agentic system designed to transform mechanistic discovery from a manual, expert-dependent process into a self-driving, closed-loop scientific endeavor. ARCHE synergizes a generalist reasoning model with a domain-adapted specialist LLM for Gaussian-based computations (ARCHE-Chem) and a structured tool registry, bridging the gap between flexible LLM reasoning and rigorous quantum-chemical execution (*30*). In this architecture, the generalist model acts as a scientific conductor, formulating diverse mechanistic hypotheses and designing high-level computational workflows. Complementing this, ARCHE-Chem serves as a domain-specific validator, enforcing physical and methodological consistency, refining computational parameters, and autonomously diagnosing failure modes (e.g., transition-state verification via frequency and IRC analysis). These components operate within a dynamic closed loop, where intermediate results are continuously evaluated against mechanistic expectations, enabling iterative hypothesis refinement without human intervention. By unifying open-ended scientific reasoning with high-fidelity computational validation, ARCHE establishes a robust framework for autonomous, interpretable mechanistic exploration (Fig. 1).

We demonstrate ARCHE's capabilities across three increasingly demanding scenarios that span methodological reliability, mechanistic discovery, and the abstraction of chemical principles. First, in a benchmark asymmetric organocatalytic reaction (*31*), the system accurately reconstructs stereocontrolling transition states and reproduces experimental selectivity trends. Second, in a recently discovered but unpublished photochemical radical transformation, ARCHE iteratively evaluates and refines competing mechanistic hypotheses, ultimately converging on a plausible mechanistic explanation. Third, in nickel-catalyzed migratory cross-coupling reactions, a rapidly emerging research frontier, the system distills a compact, chemically interpretable descriptor that governs ligand-controlled regioselectivity, resolving a long-standing mechanistic ambiguity. Together, these studies establish that ARCHE can seamlessly navigate open-ended mechanistic landscapes, transforming fragmented computational workflows into a unified, self-validating discovery platform (Fig. 1).

# Results

## ARCHE System Architecture

ARCHE is an architecture for autonomous mechanistic discovery in computational chemistry, built around the widely used quantum chemistry software Gaussian and organized as a dynamic closed loop of proposal, execution and reflection (Fig. 1a, b). The system adopts a hierarchical framework that tightly couples a dual-model collaborative reasoning layer with a structured tool registry equipped with an execution interface, jointly supporting an iterative reasoning–validation cycle. Within a unified decision space, a generalist language model explores competing mechanistic pathways and plans computational workflows, while ARCHE-Chem reviews Gaussian-specific settings, interprets computed outputs and diagnoses execution failures. To connect this reasoning process with computational chemistry tools, ARCHE employs a structured tool registry and execution interface that translate planned workflows into executable Gaussian tasks. This interface enables quantum-chemical calculations to be invoked and analysed in a manner that preserves alignment between scientific intent, mechanistic reasoning and high-fidelity computational evaluation.

In operation, ARCHE realizes this architecture as a dynamic closed-loop workflow through iterative cycles of proposal, execution and reflection (Fig. 1a). Competing mechanistic hypotheses are first generated and prioritized through scientific reasoning. The general computational workflows required to evaluate these hypotheses are then constructed by the generalist model, including task decomposition and calculation planning. When workflow steps involve Gaussian-based quantum-chemical calculations, ARCHE-Chem, a domain-specialized computational chemistry model, is invoked to refine the Gaussian-specific components of the workflow, such as route construction and input configuration. The resulting workflows are subsequently executed through quantum-chemical calculations. The computed outputs are then jointly assessed through scientific reasoning and methodology-aware analysis to examine logical coherence, computational consistency and mechanistic plausibility, with ARCHE-Chem contributing domain-specific evaluation of calculation validity and results, including checks on convergence behaviour and the reliability of key mechanistic characterizations when needed. Detected inconsistencies trigger targeted refinements in either the reasoning process or the computational configuration, leading to revision of mechanistic hypotheses, parameter adjustment, or workflow updates. Through this iterative process, ARCHE enables systematic competition among mechanistic hypotheses while maintaining methodological rigor and computational validity. Detailed descriptions of individual agents, tool definitions and the complete system workflow are provided in Supplementary Section A1.1.

## ARCHE-Chem: training and performance of Gaussian expert model

Computational chemistry workflows require theoretical principles to be translated into appropriate calculation settings, followed by interpretation of outputs and diagnosis of execution failures (*32–34*). For Gaussian-based calculations, these tasks involve selecting compatible methods and keywords, assessing convergence and the nature of stationary points, and identifying corrective actions when calculations fail. ARCHE incorporates a domain-specialized model, ARCHE-Chem, to support these tasks alongside the generalist reasoning model. Within the reasoning–validation loop, ARCHE-Chem refines Gaussian inputs, reviews computational outputs and diagnoses errors, informing revisions to calculation settings and subsequent workflow steps. Representative examples of input refinement and error diagnosis are shown in Fig. 2e,f).

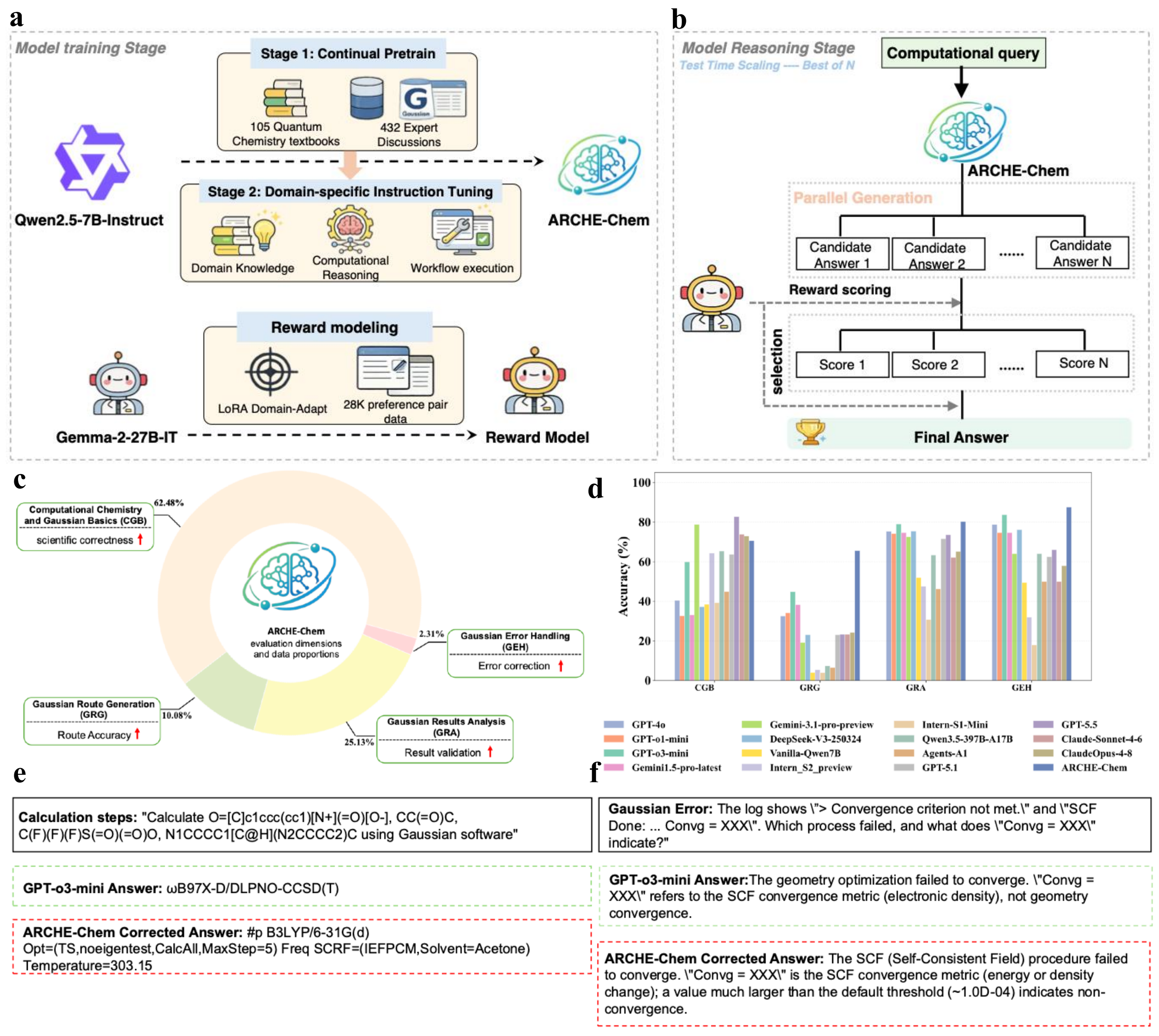

**Figure 2**: **Training, inference, evaluation dimensions and practical behaviour of ARCHE-Chem. a,** ARCHE-Chem was developed through continual pretraining and domain-specific instruction tuning, alongside reward-model training for response selection. **b,** At inference, ARCHE-Chem uses Best-of-*N* test-time scaling for computational queries. **c,** Benchmark composition across four evaluation dimensions: computational chemistry and Gaussian basics (CGB), Gaussian route generation (GRG), Gaussian results analysis (GRA), and Gaussian error handling (GEH). **d,** Performance across four benchmark categories. GPT-5.5 achieves the highest CGB score, whereas ARCHE-Chem achieves the highest scores in GRG, GRA and GEH under the reported evaluation configurations. **e,** Representative comparison with the GPT-o3-mini in Gaussian keywords generation, showing that GPT-o3-mini proposes an inappropriate method combination, whereas ARCHE-Chem produces chemically and procedurally valid keywords. **f,** Representative comparison with GPT-o3-mini in Gaussian error handling, showing that GPT-o3-mini misidentifies the failed process, whereas ARCHE-Chem correctly attributes the error to SCF non-convergence.

ARCHE-Chem is trained to encode both conceptual knowledge of computational chemistry and operational proficiency in Gaussian-based computational workflows. The training corpus comprises 105 textbooks and manuals covering quantum chemistry and Gaussian software, together with 432 expert-level forum discussions documenting workflow conventions, troubleshooting strategies and methodological preferences. From these materials, we constructed an instruction-style dataset organized into four task categories: computational chemistry and Gaussian basics, Gaussian route generation, Gaussian result analysis and Gaussian error handling. All samples were manually curated and expert-validated to ensure consistency with Gaussian conventions and to reflect authentic computational practice. Evaluation tasks and case studies reported in this work were not included in the training corpus. Further details of the training dataset are provided in Supplementary A.2.

Model training followed a two-stage procedure (Fig. 2a). Continual pretraining first adapted a base instruction-tuned model (Qwen2.5-7B-Instruct (*35*)) to the terminology, conventions and practical knowledge embedded in computational chemistry and Gaussian-related corpora, followed by supervised fine-tuning to align the model with task-specific instruction formats. To improve robustness against plausible but failure-prone configurations, we trained a reward model (Gemma-2-27B-IT (*36*)) to score outputs based on syntactic validity and chemical consistency. During inference (Fig. 2b), multiple candidates are generated and ranked, and the highest-scoring output is selected to improve robustness (*37*). Further details of model training and hyperparameter settings are provided in Supplementary Section A.2.

We evaluated ARCHE-Chem alongside 15 other models on held-out benchmarks covering four task categories aligned with its functions within ARCHE: computational chemistry and Gaussian basics (CGB), Gaussian route generation (GRG), Gaussian result analysis (GRA), and Gaussian error handling (GEH) (Fig. 2d). ARCHE-Chem achieved accuracies of 70.65%, 65.60%, 80.30%, and 87.60%, respectively. GPT-5.5 achieved the highest accuracy on CGB at 82.71%, whereas ARCHE-Chem achieved the highest accuracies on GRG, GRA, and GEH. The largest difference was observed in route generation: ARCHE-Chem achieved 65.60% accuracy, compared with 44.80% for GPT-o3-mini and 3.91–38.30% for the remaining models. Its advantages in result analysis and error handling were smaller, exceeding the next-highest accuracies by 1.30 and 3.90 percentage points, respectively. GPT-5.5's higher CGB accuracy may reflect broader command of computational chemistry concepts and basic Gaussian knowledge. Route generation, output interpretation and error diagnosis require this knowledge to be applied to specific calculation settings, keyword combinations and diagnostic details. These requirements closely match ARCHE-Chem's training on computational chemistry textbooks, software documentation and expert discussions, followed by task-oriented instruction tuning. Such training may help connect computational objectives with appropriate Gaussian configurations and the interpretation of outputs and errors, while reward-guided Best-of-N selection may favour responses that better satisfy these requirements.

## Case Study 1 | Autonomous reproduction of stereocontrolling transition states in a reported vicinal diamine-catalysed aldol reaction

To test the reliability of ARCHE in performing autonomous mechanistic investigation, we revisited the vicinal diamine-catalysed asymmetric aldol reaction analysed by the Houk group in 2016 (*31*). This previous study employed density functional calculations to locate stereocontrolling transition states for vicinal diamine-catalysed aldol reactions, including the acetone and p-nitrobenzaldehyde reaction catalysed by vicinal diamines (*38*). The authors demonstrated that stereocontrolling cyclic hydrogen-bonded transition states, particularly the crown conformations, account for the observed stereoselectivity.

ARCHE was given only a structured reaction specification comprising the reactants, product, catalyst, solvent and temperature (Fig. 3a); experimental selectivity, mechanistic interpretation, transition-state structures, computational protocols and energetic results from the reference study were withheld and were not represented in the training data. This setup therefore provided a stringent test of reliability by asking whether the system could recover the published mechanistic analysis from reaction-level information alone.

Given only this limited reaction specification, ARCHE autonomously executed the full multi-agent reasoning and computational workflow (Fig. 3a). The retrieval agent first extracted mecha-nistically relevant concepts from the input scientific question and used them to retrieve background knowledge on asymmetric aldol catalysis and hydrogen-bond-mediated stereocontrol, enabling the hypothesis agent to formulate competing enantio-determining transition-state models. The planning agent then converted these hypotheses into complete computational workflows, including conformational sampling, geometry optimisation, transition-state verification and thermochemical analysis, which were translated by the execution module into executable tasks and run to generate structured structural and energetic results for evaluation. No manual intervention was involved throughout the process.

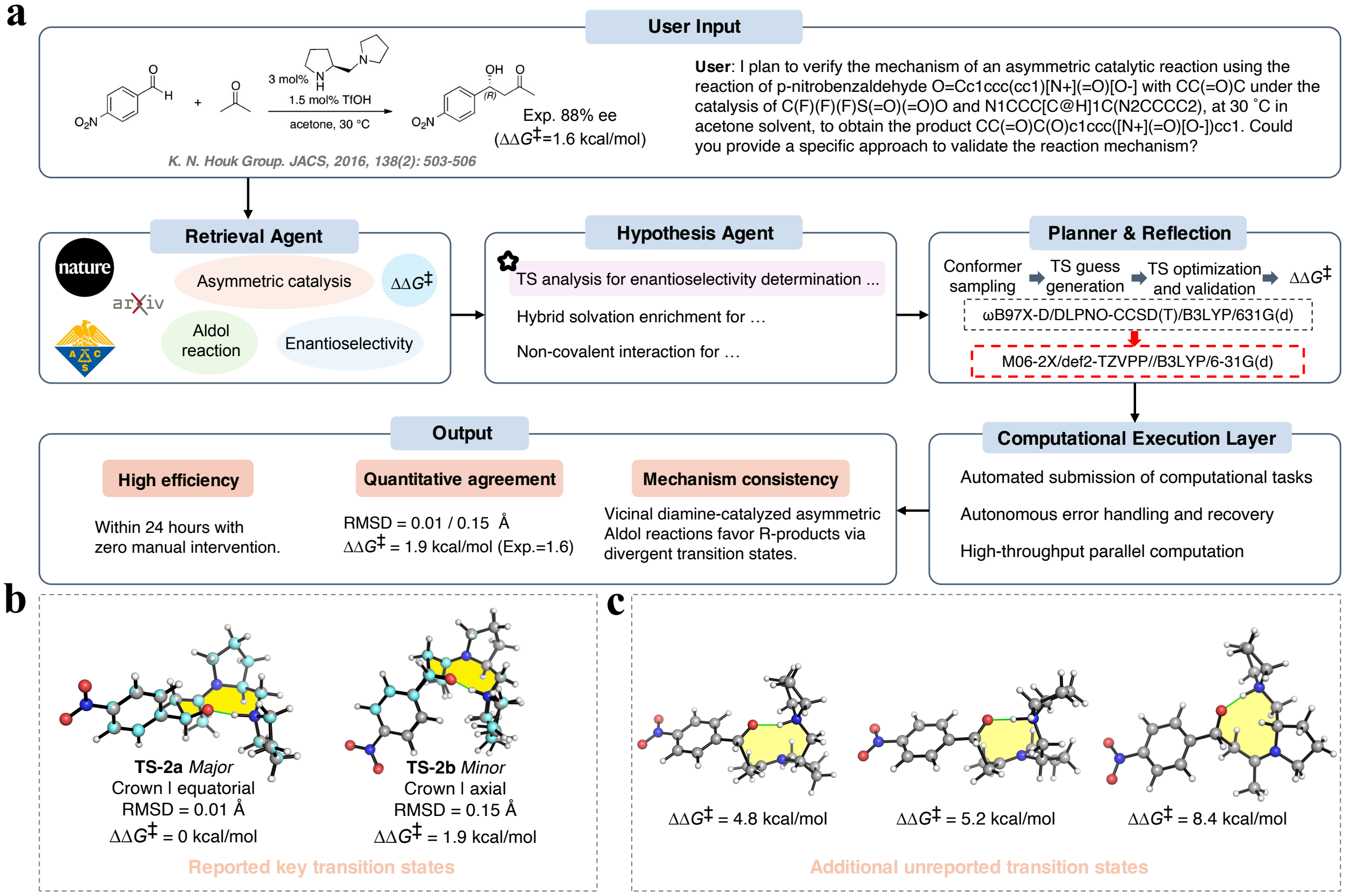

**Figure 3**: **Autonomous mechanistic analysis of an asymmetric aldol reaction. a,** Autonomous workflow for literature retrieval, hypothesis generation, computational planning, transition-state search and enantioselectivity evaluation. **b,** Literature-reported enantiodetermining transition-state structures recovered by ARCHE. Hydrogen-bonding interactions are highlighted by solid green lines, and the reaction core is highlighted in yellow. TS-2a and TS-2b correspond to the most important major and minor enantiomeric transition states, respectively. **c,** Previously unreported higher-energy transition-state structures identified by ARCHE. The reaction core is highlighted in yellow.

ARCHE identified 12 transition-state structures, including the nine reported in the reference study and three additional candidates not reported there (Fig. 3c). Details of all 12 structures are provided in Supplementary Section B.1.1. The additional structures indicate systematic exploration of the stereochemical and conformational landscape and, although higher in energy, represent chemically reasonable alternatives that were correctly identified as less favourable than the dominant pathways. Among the explored structures, the lowest-energy transition states leading to the two enantiomeric products (Fig. 3b) corresponded to the competing enantio-determining pathways described by the Houk group. Structural superposition between the automatically identified and literature transition states gave RMSD values (backbone only) of 0.01 Å for **TS-2a** and 0.15 Å for **TS-2b**, indicating close agreement in the stereochemistry-defining arrangements. The computed difference in activation free energies between the two pathways was 1.9 kcal/mol, close to the 1.6 kcal/mol value inferred from the reported experimental enantiomeric excess of 88% at 303.15 K.

All stages of the analysis were completed within 24 hours without manual intervention or post hoc adjustment. Together, these results show that ARCHE can reliably reconstruct the transition-state model and its stereochemical outcome from reaction-level input alone. This reliability benchmark provides the foundation for the subsequent validation cases.

## Case Study 2 | Autonomous mechanistic analysis of a newly discovered $\alpha$-iodoboronate C–I cleavage reaction

To test whether ARCHE can unravel previously unknown reaction mechanisms, we investigated a new blue-light-driven reaction experimentally discovered by our collaborators in the XU group. In this reaction, irradiation of $\alpha$-iodoboronate **1** with $HPPh_2$ and $Cs_2CO_3$ under 450 nm light affords alkylboronate **2** in 83% yield. Control experiments (Fig. 4c) showed that product formation requires blue light, base and phosphine, and is suppressed by radical scavenger TEMPO, indicating the involvement of radical intermediates, while the underlying activation mode and overall reaction mechanism remain unresolved.

Given only this experimental description and without prespecifying the reaction pathway, ARCHE generated a set of plausible mechanistic hypotheses (Fig. 4a). ARCHE first retrieved literature precedents relevant to the transformation, including halogen activation, $\alpha$-boryl radical stabilisation and phosphine-mediated redox processes. On this basis, the system formulated six representative activation modes and subjected them to quantitative evaluation through correspond-ing computational workflows. Through these calculations, ARCHE excluded concerted pathways, whose Gibbs free energy barriers reached 46.0 kcal·mol$^{-1}$ and were therefore incompatible with room-temperature reactivity. ARCHE then examined direct photoexcitation of isolated **1** and its $PPh_2$-associated counterpart. Excited-state calculations showed that their lowest singlet excitation energies are 102.8 and 75.8 kcal·mol$^{-1}$, respectively, both well above the 63.6 kcal·mol$^{-1}$ photon energy provided by 450 nm irradiation. Three representative cases are discussed below, with evaluation results summarised in Supplementary Fig. S3. The corresponding prompts and intermediate outputs are provided in Supplementary Section C.

These negative results triggered a reflective step within ARCHE, during which the system reconsidered the possible roles of the reaction components and proposed an alternative activation scenario. ARCHE then evaluated this revised hypothesis computationally. The resulting calculations showed that the revised model markedly alters the photophysical properties of the system, giving rise to an electronic excitation of approximately 64.0 kcal·mol$^{-1}$, in close agreement with the energy delivered by 450 nm irradiation. On this basis, ARCHE identified a plausible initiation pathway for the deiodination process, in which photoinduced activation enables homolytic C–I bond cleavage and generates radical intermediates (Fig. 4b).

Together, these calculations support a refined mechanistic scenario in which formation of the $CsPPh_2$-complex is thermodynamically favourable, its calculated excitation energy matches that of 450 nm irradiation, and the ensuing radical pathway is consistent with the available experimental evidence, including TEMPO inhibition. This activation mode was not specified a priori, but emerged through iterative hypothesis generation, computational evaluation and reflective revision. The case therefore shows that ARCHE can move beyond literature reconstruction to identify a chemically consistent mechanism for a previously unresolved reaction.

## Case Study 3 | Abstracting mechanistic descriptors for migratory cross-coupling reactions

Chemically meaningful descriptors condense complex mechanistic behavior into compact and interpretable principles, allowing selectivity trends to be understood in a form that extends beyond individual computed structures or energies (*39–41*). Identifying such descriptors is substantially more demanding than executing a computational workflow, because it requires isolating the controlling factor of a reaction outcome from many plausible geometric and electronic features and thus depends heavily on expert mechanistic judgment. We therefore selected descriptor discovery as a stringent test of whether ARCHE could move beyond automated calculation to mechanistic abstraction, namely, whether it could extract a concise and chemically meaningful principle from computed evidence in a mechanistically complex catalytic system.

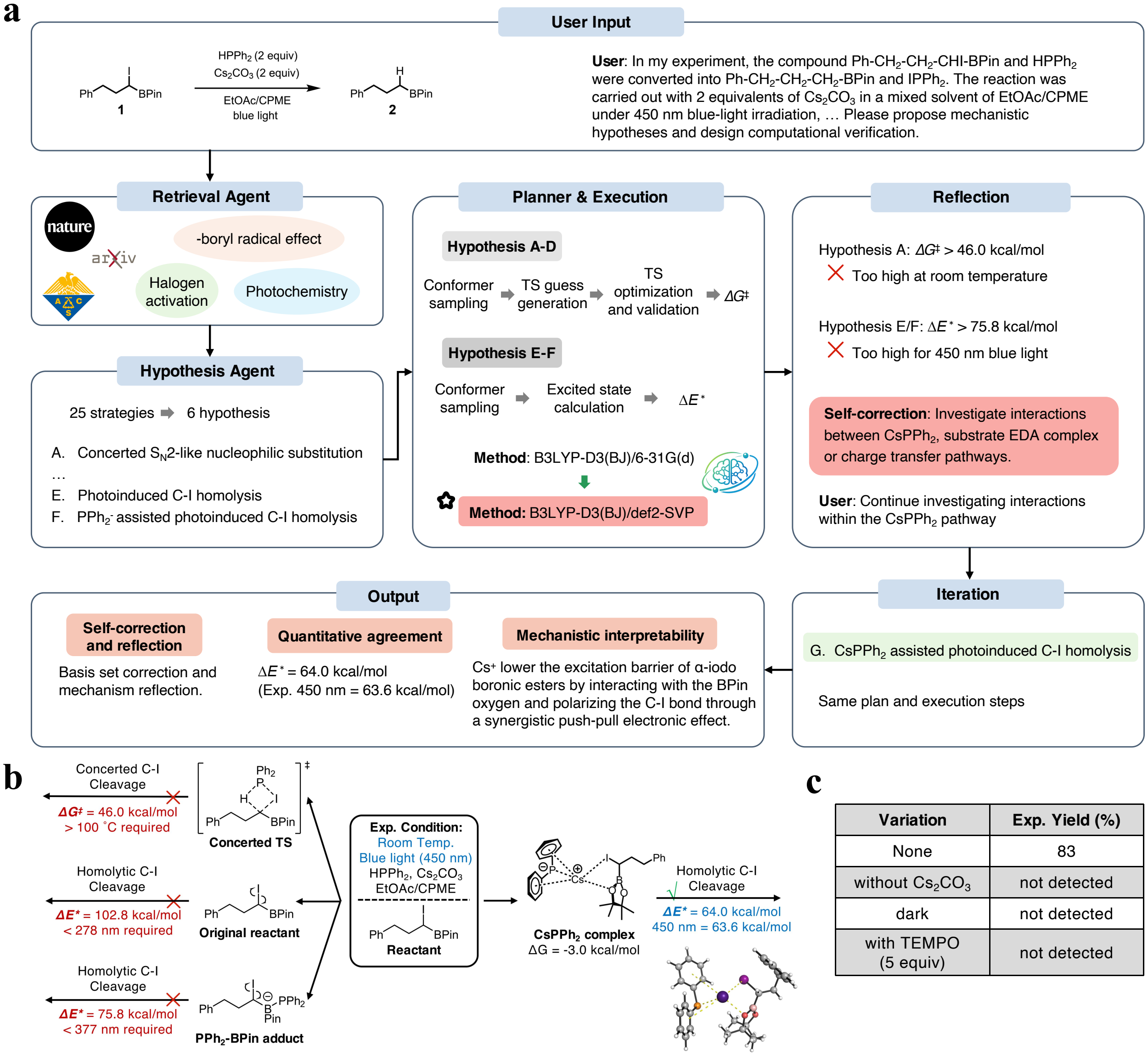

| Variation | Exp. Yield (%) |
|---|---|
| None | 83 |
| without $Cs_2CO_3$ | not detected |
| dark | not detected |
| with TEMPO (5 equiv) | not detected |

**Figure 4**: **Autonomous identification of a photoinduced radical pathway by ARCHE under blue-light irradiation. a,** Overview of the value-validation workflow, showing how an initially standard, hypothesis-enumeration approach fails to reconcile experimental constraints, prompting reflection, contradiction detection, and autonomous hypothesis refinement. The system transitions from broad mechanistic exploration to a targeted photochemical complex hypothesis through iterative retrieval, planning, execution, and reflection. **b,** Computed reaction pathways illustrating that concerted C-I cleavage, direct and $PPh_2^-$-assisted C–I homolysis are all energetically inaccessible under 450 nm irradiation. Only the $CsPPh_2$-complex matches the blue-light energy profile, facilitating the radical pathway. The three-dimensional structure of the $CsPPh_2$ complex are presented below. **c,** Control experiments (base removal, dark conditions, TEMPO) demonstrating suppression of product formation, consistent with the refined activation mechanism identified by the system.

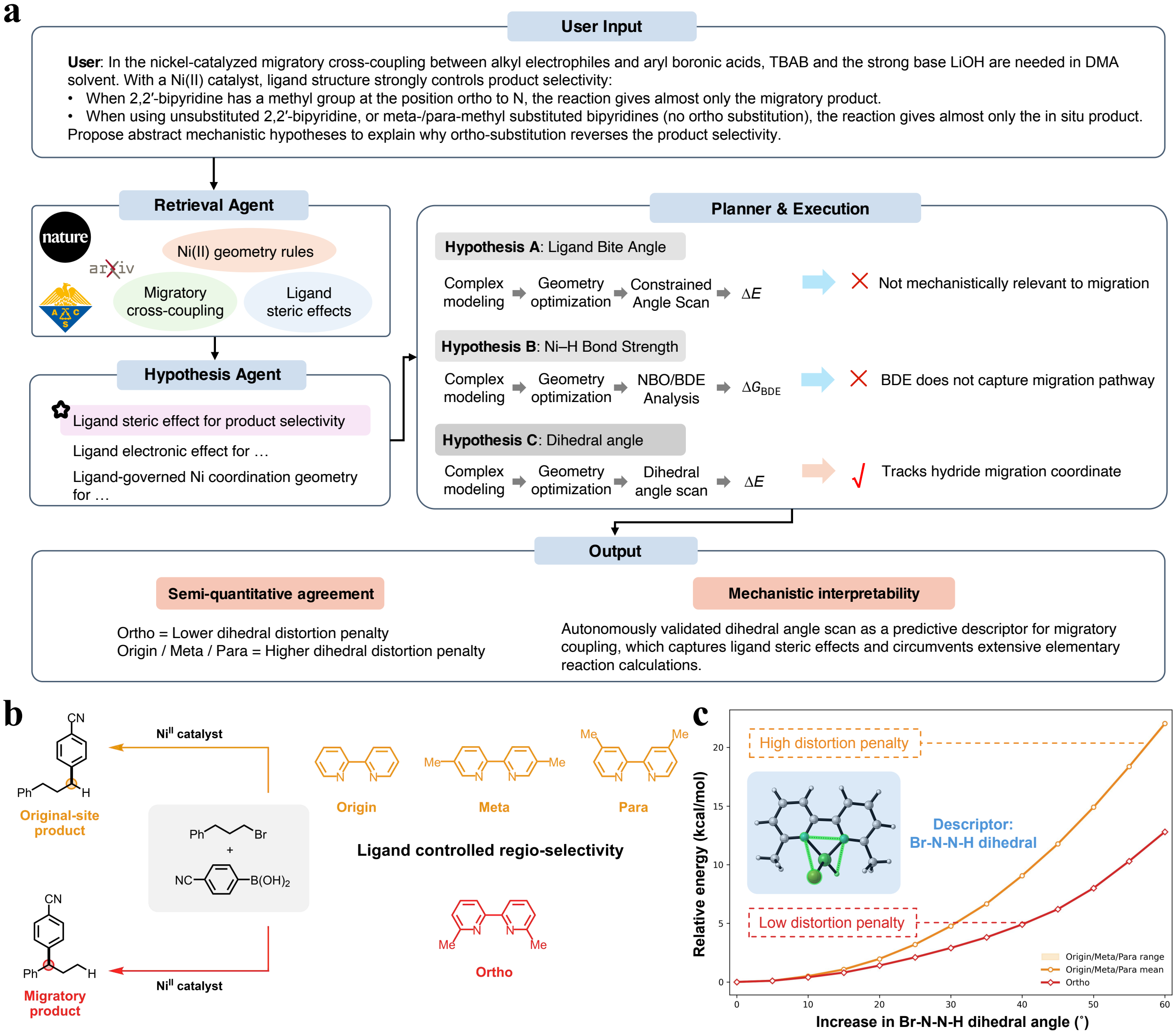

**Figure 5**: **Abstraction of mechanistic descriptors for migratory cross-coupling reaction by ARCHE. a,** Workflow for mechanistic abstraction. The system compresses a complex reaction network into a single descriptor—the Br–N–N–H dihedral torsional energy—and verifies that ortho substitution lowers the rotational barrier, explaining the observed migratory selectivity. **b,** Ligand-controlled regioselectivity in Ni-catalysed cross-coupling. Ortho-methyl substitution promotes the migratory product, whereas the unsubstituted parent and meta-/para-methyl variants favour the original-site product. **c,** Distorsion–energy profiles derived from the ARCHE identified dihedral descriptor. The ortho-methyl ligand shows a lower energy penalty for distortion of the Br–N–N–H dihedral angle.

Nickel-catalysed migratory cross-coupling reactions represents a high-impact and extensively studied transformation in modern synthetic chemistry, with numerous mechanistic investigations documented in the literature (*42–46*). In these systems, unsubstituted and meta- or para-substituted N,N-bidentate ligands typically favour original-site coupling, whereas ortho-substituted analogues selectively promote migratory pathways. Distilling a general mechanistic descriptor for this reaction class would provide substantial value to the field, as it would capture shared selectivity principles across diverse substrates rather than remaining confined to isolated examples. A representative system reported in 2020 (*47*) exemplifies this challenge: although the ligand-dependent regioselectivity is experimentally well established, the underlying structural or electronic factor that governs this switch has not yet been discovered. This reaction therefore provides a stringent test of whether ARCHE can autonomously extract a single, interpretable chemical descriptor from a complex mechanistic landscape.

Starting from a natural-language description of the observed selectivity trend, ARCHE generated several candidate structural hypotheses, including the rotation of the Br–N–N–H dihedral angle, modulation of ligand bite angle and electronic regulation of the Ni–H bond strength (see Supplementary Section B.1.3). Rather than committing to a detailed mechanistic sequence, each proposal was treated as a testable structural abstraction. The planner agent translated these hypotheses into targeted computational chemistry tasks, including relaxed dihedral scans and controlled geometric distortions of a square-planar Ni(II)(bpy)–Br–H intermediate. After verification of the computational setup by the ARCHE-Chem, the calculations were executed through the execution module to obtain energetic profiles along defined structural coordinates.

Across the tested features, the system consistently converged on a single geometric quantity: the energetic cost of rotating the Ni–H bond out of the coordination plane, namely the distortion energy of Br–N–N–H dihedral angle (Fig. 5c). Relaxed dihedral scans reveal a clear ligand-dependent trend.

Ortho-methyl substitution destabilises the coplanar geometry through steric repulsion, lowering the torsional energy penalty and favouring non-planar conformations that facilitate hydride migration. In contrast, unsubstituted and meta- or para-substituted ligands maintain a near-planar coordination environment, incur a higher torsional energy cost and correspondingly favour the non-migratory pathway. This torsional descriptor reproduces the experimentally observed selectivity trend across the ligand series and is consistent with conclusions previously obtained from extensive transition-state analyses.

By identifying a chemically intuitive Br–N–N–H torsional coordinate as a key mechanistic descriptor, ARCHE distills intricate reaction mechanisms into an interpretable, descriptor-guided analytical framework. This case validates that ARCHE can autonomously distinguish mechanistically critical features from complex reaction systems. More broadly, it establishes a generalizable strategy to uncover fundamental selectivity-governing principles in nickel-catalysed migratory cross-coupling systems. This fills a longstanding field gap by providing compact, chemically meaningful descriptors for this widely studied reaction class.

# Discussion

Mechanistic research in chemistry requires not only performing calculations, but also generating competing hypotheses, quantitatively discriminating among them and iteratively refining explanations under chemical and experimental constraints. Here we show that these processes can be organized within an autonomous framework that couples a general reasoning model for hypothesis exploration with a domain-specialized computational chemistry model that enforces methodological validity. Through an iterative reasoning–validation cycle, this integration enables systematic exploration of mechanistic possibilities while maintaining computational reliability.

Across three increasingly demanding scenarios spanning reliability, discovery and mechanistic abstraction, the results highlight how embedding domain-level expertise improves the generation and evaluation of mechanistic hypotheses. The system reproduces established stereocontrolling transition states with quantitative agreement to experiment, identifies a previously unreported radical pathway consistent with experimental observations, and extracts a minimal torsional descriptor governing migratory selectivity. Together, these results show that integrating domain-specific com-putational expertise enables more reliable and chemically grounded mechanistic analysis than approaches relying solely on general-purpose models or workflow automation.

More broadly, this work speaks to a question that is becoming increasingly important across the sciences: what it would take for AI systems to contribute genuine scientific value in domains where progress depends on specialised knowledge, rigorous validation and scientifically interpretable conclusions. As AI systems are applied to increasingly larger parts of the research process, the central issue is not simply whether individual tasks can be automated, but whether such systems can support reliable, evidence-grounded inquiry under the real methodological and practical constraints of scientific research. In this sense, our study is not only about reaction mechanisms. Rather, it illustrates how autonomous systems may begin to contribute to scientific inquiry when mechanistic reasoning is tightly coupled to domain-specific expertise and explicit computational validation. While ARCHE currently operates within a Gaussian-centric workflow, its modular tool registry and dual-model architecture provide a transferable blueprint for extending self-validating mechanistic exploration to other domain-specific scientific software and even closed-loop autonomous laboratory systems that integrate robotic execution with real-time analytical feedback.

# Methods

## ARCHE system overview

ARCHE is a closed-loop system for reaction-mechanism discovery that couples generalist–specialist collaborative reasoning with explicit computational execution (Fig. 1). From a scientific question, it retrieves relevant knowledge, generates and ranks competing mechanistic hypotheses, translates prioritized hypotheses into executable computational workflows, and carries them out autonomously through external computational chemistry tools (Fig. 1a). This process is coordinated through a collaborative decision space in which a generalist model supports mechanistic exploration and the domain-specialized ARCHE-Chem model provides methodological rigour and physical consistency during computational planning and evaluative reflection (Fig. 1b). A structured tool registry and execution interface connects these reasoning components to external tools, enabling calculated results to be returned in structured form for iterative assessment and revision. In this way, ARCHE implements mechanistic elucidation as a reasoning–validation cycle in which hypotheses are formulated, computationally evaluated and refined on the basis of calculated evidence. Detailed descriptions of the individual agents and their roles are provided in Supplementary Section A1.1.

## Reasoning–validation circle

ARCHE implements reaction-mechanism elucidation through an iterative reasoning–validation cycle that couples a generalist reasoning model with the domain-specialized computational chemistry model ARCHE-Chem. Starting from a scientific question, the system first retrieves relevant chemical knowledge to ground subsequent analysis. The generalist model then formulates competing mechanistic hypotheses and translates them into computational workflows designed to test the evidence most decisive for distinguishing among them. Within this process, ARCHE-Chem provides the methodological and technical judgement required for execution-critical computational chemistry tasks, including refining Gaussian keywords and related computational settings, diagnosing calculation failures and interpreting computed outputs. Validated workflows are passed to the execution module, which performs the corresponding computational chemistry calculations and returns structured results for assessment. Reflection is triggered when calculations fail, when computed evidence does not support the proposed mechanistic claim or when inconsistencies arise between calculated results and observed reactivity. In such cases, the system revises the hypothesis, the workflow or both, and initiates further evaluation until a chemically consistent mechanistic conclusion is reached. Further details of the individual agents and their roles are provided in Supplementary Section A1.

## ARCHE-Chem training and inference

ARCHE-Chem is a domain-specialized model derived from Qwen2.5-7B-Instruct and developed to support Gaussian input refinement, output interpretation and error diagnosis within ARCHE. The model underwent continual pretraining and supervised fine-tuning, followed by reward-guided response selection at inference.

For continual pretraining, the base model was trained on a curated corpus of computational chemistry literature and technical documentation, including textbooks, official Gaussian manuals and documentation, and expert-level discussions and articles from online forums and the scientific literature. The corpus comprised 105 textbooks and manuals together with 432 expert discussions and articles, corresponding to approximately 286 MB of text. Full-parameter optimization was used to adapt the model to domain-specific terminology, theoretical concepts and practical computational conventions.

The model was then supervised fine-tuned on an instruction dataset of approximately 50,000 question–answer pairs constructed from the curated corpus. These data covered major categories of computational chemistry tasks, including computational chemistry and Gaussian basics, result interpretation, Gaussian route generation, output-file analysis and diagnosis of common computational errors. This stage was used to convert domain knowledge into task-oriented expert behaviour.

No task-specific information from the evaluation scenarios was included in the training data. In particular, the curated corpus and fine-tuning set did not contain the evaluated mechanistic hypotheses, transition-state structures, computational procedures, energetic results or benchmark instances used in this work. The reported mechanistic conclusions were therefore derived from explicit computational chemistry calculations performed during inference rather than retrieved from memorized task-specific content.

To improve response selection during inference, we trained a reward model based on Gemma-2-27B-IT using 28,000 preference pairs. ARCHE-Chem was deployed with reward-guided test-time scaling, in which multiple candidate responses were sampled, scored by the reward model and ranked accordingly. A Best-of-N strategy with $N = 8$ was used, and the highest-scoring response was selected as the final output. For details of model training, see Supplementary Section A2.

## Benchmark design for ARCHE-Chem evaluation

ARCHE-Chem was evaluated on held-out benchmarks spanning four task categories covering computational chemistry and Gaussian basics, Gaussian route generation, result analysis and error handling. These categories were designed to distinguish between descriptive methodological familiarity and execution-critical capabilities required for practical computational chemistry support. Representative task types included interpretation of Gaussian keywords and method settings, generation of executable route sections, analysis of Gaussian outputs and diagnosis of failed calculations. Full benchmark construction details, category definitions and example tasks are provided in Supplementary Section A2.3.

## Tool registry and executable workflow construction

To translate mechanistic hypotheses into executable computational studies, ARCHE uses a structured tool registry that defines the executable interface between reasoning and computation. For each operation, the registry specifies its computational function together with the required inputs and expected outputs, so that mechanistic hypotheses can be converted into schema-constrained workflows that are executable in practice. The registry includes both basic utilities, such as molecular structure conversion and parsing of Gaussian output files, and workflow-level tools that encapsulate recurring computational chemistry operations. During workflow construction, candidate operations are selected and assembled to satisfy the computational objective, with the declared input–output relations ensuring that each step provides the information required by the next. If the output of one step cannot be used directly as the input of the following step, the system inserts the necessary conversion operation automatically, for example to transform a molecular structure from one file format or coordinate representation to another. This conversion-aware assembly is required for workflow compositionality, ensuring that multi-step workflows remain executable as complete procedures rather than only as conceptual plans. The resulting workflow therefore provides a structured, traceable and executable program through which mechanistic reasoning is connected to computational evaluation. For details of the tool registry and representative tool examples, see Supplementary Section A3.

## Evaluation framework

Evaluation was performed at two levels. At the model level, ARCHE-Chem was assessed on benchmark tasks covering core capabilities required for computational chemistry support, including computational chemistry and Gaussian basics, Gaussian route generation, output analysis and error handling. At the system level, ARCHE was evaluated through three value-demonstration scenarios that tested end-to-end reaction-mechanism elucidation, including reproduction of established mechanistic conclusions, resolution of an unresolved mechanistic question and extraction of an interpretable mechanistic descriptor from a complex catalytic system.

The model-level benchmarks were designed to provide a reliable assessment of ARCHE-Chem's expert capabilities. Benchmark instances were curated to cover the major task types encountered in computational chemistry practice and were manually reviewed for technical correctness, task relevance and consistency of the reference answers. To reduce the risk of contamination, these benchmark data were kept separate from the continual pretraining corpus, supervised fine-tuning dataset and reward-model preference data used for model development.

At the system level, each scientific hypothesis was evaluated through its corresponding computational workflow generated within the ARCHE framework. In this setting, evaluation was based on whether the resulting computational evidence supported the proposed mechanistic claim and whether the full reasoning-to-execution process could produce chemically consistent conclusions for the problem under study. The three case studies scenarios were kept separate from model development data, and no task-specific mechanistic hypotheses, structures, computational procedures or target results from these scenarios were included in training or preference data. The reported system-level conclusions therefore arose from explicit computational chemistry calculations performed during inference. The system-generated mechanistic hypotheses, computational workflows, their refined versions after iterative revision, and the computational results extracted by the model from Gaussian log files for all three value demonstrations are provided in Supplementary C.

## Use of proprietary models

In the present implementation of ARCHE, GPT-5 was used through an API to support the Retrieval Agent, Hypothesis Agent and Planner Agent. As these components depend on a proprietary model, exact reproduction of some agent-level outputs may be affected by model availability and version changes. We report this explicitly for transparency.

## Computational chemistry protocols

All electronic-structure calculations were performed using Gaussian 16 (*30*). Molecular structure preparation, format conversion and data analysis were assisted by external computational tools, including RDKit (*48*), Open Babel (*49*), ASE (*50*), cclib (*51*) and Multiwfn (*52*), among others (*53–55*). Detailed Gaussian route specifications for each case study are provided in Supplementary Tab.5.

For the asymmetric catalysis case study, geometry optimizations and frequency calculations were performed at the B3LYP/6-31G(d) level (*56–58*) with the IEFPCM solvation model (*59*) for acetone. Transition states were verified using intrinsic reaction coordinate (IRC) calculations, and electronic energies were refined using M06-2X/def2-TZVPP single-point calculations (*60, 61*).

For the radical reaction system, geometry optimizations and frequency calculations were carried out at the B3LYP-D3(BJ)/def2-SVP level (*62*) with the SMD solvation model (*63*) in ethyl acetate. Time-dependent density functional theory (TD-DFT) calculations (*64*) were used to evaluate excited states when required.

For descriptor-based mechanistic analysis, geometry optimizations were performed using $\omega$B97X-D (*65*) with def2-SVP basis set and the SMD solvation model in dimethylacetamide (DMA). All stationary points were confirmed by vibrational frequency analysis. The full Gaussian keywords used in the value demonstrations are provided in Supplementary Section A4.

## Runtime environment and implementation

ARCHE was implemented in a runtime environment that separates language-model reasoning from computational chemistry calculations. Model training and inference were performed on GPU-based servers, whereas computational chemistry calculations were executed on CPU resources within the ARCHE execution environment. This resource separation allowed semantic reasoning and chemistry calculations to operate in a coordinated but decoupled manner.

ARCHE-Chem was trained on a GPU cluster equipped with eight NVIDIA A100 GPUs (80 GB each) running Ubuntu 22.04 LTS, using PyTorch 2.2 and DeepSpeed with mixed-precision (FP16) training. Continual pretraining and supervised fine-tuning required approximately 21 wall-clock hours in total (approximately 168 GPU-hours). During inference, the generalist reasoning model and ARCHE-Chem were deployed on GPU servers within the ARCHE framework.

All computational chemistry calculations were performed using Gaussian 16 (Revision B.01). Production runs for the value-demonstration scenarios were executed on a dedicated node equipped with dual Intel Xeon Platinum 8352Y CPUs and 512 GB RAM, with Gaussian jobs run on multi-core CPU threads. This production node was used for the end-to-end studies reported in the main text. Computational parameters for each case study, including density functional methods, basis sets, solvation models and key Gaussian settings, are summarized in Supplementary Tab.S5.

The ARCHE tools stack comprised the language-model orchestration framework together with computational chemistry utilities for molecular structure handling, format conversion, workflow preparation and output parsing, including RDKit, Open Babel, ASE, cclib and Multiwfn and related utilities.

## Code availability

Code for the ARCHE framework is publicly available at GitHub.

## Data availability

The data underlying this study, including the training data used for ARCHE-Chem, are publicly available at GitHub.

## Acknowledgments

This work was conducted during Dong Li's internship at Shanghai Artificial Intelligence Laboratory. The project was jointly led by Dr. Biqing Qi and Dr. Yuqiang Li. We sincerely thank Dr. Yuqiang Li and Prof. Tao Xu for providing three validation scenarios for the evaluation and testing of our system. We would also like to express our special gratitude to Prof. Bowen Zhou for his continuous support and interest in this project. This work was supported by the National Natural Science Foundation of China (Grant No. 6250076080) and New Generation Artificial Intelligence-National Science and Technology Major Project(2025ZD0121802) and Intern-Discovery.

**Competing interests:** There are no competing interests to declare.

## References and Notes

1. E. Kraka, D. Cremer, Computational analysis of the mechanism of chemical reactions in terms of reaction phases: hidden intermediates and hidden transition states. *Accounts of chemical research* **43** (5), 591–601 (2010).

2. S. Chen, R. Babazade, T. Kim, S. Han, Y. Jung, A large-scale reaction dataset of mechanistic pathways of organic reactions. *Scientific Data* **11** (1), 863 (2024).

3. J. F. Joung, *et al.*, Reproducing Reaction Mechanisms with Machine-Learning Models Trained on a Large-Scale Mechanistic Dataset. *Angewandte Chemie International Edition* **63** (43), e202411296 (2024).

4. G. J. Hutchings, Spiers Memorial Lecture: Understanding reaction mechanisms in heterogeneously catalysed reactions. *Faraday Discussions* **229**, 9–34 (2021).

5. S. Ahn, M. Hong, M. Sundararajan, D. H. Ess, M.-H. Baik, Design and optimization of catalysts based on mechanistic insights derived from quantum chemical reaction modeling. *Chemical reviews* **119** (11), 6509–6560 (2019).

6. C. N. Prieto Kullmer, *et al.*, Accelerating reaction generality and mechanistic insight through additive mapping. *Science* **376** (6592), 532–539 (2022).

7. P. C. St. John, Y. Guan, Y. Kim, S. Kim, R. S. Paton, Prediction of organic homolytic bond dissociation enthalpies at near chemical accuracy with sub-second computational cost. *Nature communications* **11** (1), 2328 (2020).

8. A. J. Ribeiro, I. G. Riziotis, J. D. Tyzack, N. Borkakoti, J. M. Thornton, EzMechanism: an automated tool to propose catalytic mechanisms of enzyme reactions. *Nature Methods* **20** (10), 1516–1522 (2023).

9. Z. Zhang, G. Piccini, Exploring chemistry and catalysis by biasing skewed distributions via deep learning. *Nature Communications* (2026).

10. M. Steiner, M. Reiher, A human-machine interface for automatic exploration of chemical reaction networks. *Nature Communications* **15** (1), 3680 (2024).

11. Y. Ruan, *et al.*, An automatic end-to-end chemical synthesis development platform powered by large language models. *Nature communications* **15** (1), 10160 (2024).

12. K. D. Vogiatzis, *et al.*, Computational approach to molecular catalysis by 3d transition metals: challenges and opportunities. *Chemical reviews* **119** (4), 2453–2523 (2018).

13. J. A. Gauthier, *et al.*, Challenges in modeling electrochemical reaction energetics with polarizable continuum models. *Acs Catalysis* **9** (2), 920–931 (2018).

14. A. Ghafarollahi, M. J. Buehler, SciAgents: automating scientific discovery through bioinspired multi-agent intelligent graph reasoning. *Advanced Materials* **37** (22), 2413523 (2025).

15. Z. Lai, Y. Pu, PriM: Principle-Inspired Material Discovery through Multi-Agent Collaboration. *ArXiv* **abs/2504.08810** (2025).

16. A. Ghafarollahi, M. J. Buehler, Sparks: Multi-Agent Artificial Intelligence Model Discovers Protein Design Principles. *ArXiv* **abs/2504.19017** (2025).

17. K. Swanson, W. Wu, N. L. Bulaong, J. E. Pak, J. Zou, The Virtual Lab of AI agents designs new SARS-CoV-2 nanobodies. *Nature* **646** (8085), 716–723 (2025).

18. D. A. Boiko, R. MacKnight, B. Kline, G. Gomes, Autonomous chemical research with large language models. *Nature* **624** (7992), 570–578 (2023).

19. A. M. Bran, *et al.*, Augmenting large language models with chemistry tools. *Nature Machine Intelligence* **6** (5), 525–535 (2024).

20. Y. Zou, *et al.*, El Agente: An autonomous agent for quantum chemistry. *Matter* **8** (7) (2025).

21. J. Liu, *et al.*, VASPilot: MCP-facilitated multi-agent intelligence for autonomous VASP simulations. *Chinese Physics B* **34** (11), 117106 (2025).

22. A. J. Goodell, S. N. Chu, D. Rouholiman, L. F. Chu, Large language model agents can use tools to perform clinical calculations. *npj Digital Medicine* **8** (1), 163 (2025).

23. Y. Kang, J. Kim, ChatMOF: an artificial intelligence system for predicting and generating metal-organic frameworks using large language models. *Nature communications* **15** (1), 4705 (2024).

24. I. Mandal, *et al.*, Evaluating large language model agents for automation of atomic force microscopy. *Nature Communications* **16** (1), 9104 (2025).

25. T. D. Pham, A. Tanikanti, M. Keçeli, ChemGraph as an agentic framework for computational chemistry workflows. *Communications Chemistry* (2026).

26. S. X. Leong, S. Pablo-García, B. Wong, A. Aspuru-Guzik, MERMaid: Universal multimodal mining of chemical reactions from PDFs using vision-language models. *Matter* **8** (12) (2025).

27. Z. Zhao, *et al.*, Developing ChemDFM as a large language foundation model for chemistry. *Cell Reports Physical Science* **6** (4) (2025).

28. Y. Zhang, *et al.*, Exploring the role of large language models in the scientific method: from hypothesis to discovery. *npj Artificial Intelligence* **1** (1), 14 (2025).

29. Y. Yang, *et al.*, BatGPT-Chem: a foundation large model for chemical engineering. *Research* **8**, 0827 (2025).

30. M. J. Frisch, *et al.*, Gaussian 16 Revision C.01 (2016), gaussian Inc. Wallingford CT.

31. A. Simon, Y.-h. Lam, K. Houk, Transition states of vicinal diamine-catalyzed aldol reactions. *Journal of the American Chemical Society* **138** (2), 503–506 (2016).

32. M. C. Ramos, C. J. Collison, A. D. White, A review of large language models and autonomous agents in chemistry. *Chemical science* (2025).

33. J. Van Herck, *et al.*, Assessment of fine-tuned large language models for real-world chemistry and material science applications. *Chemical science* **16** (2), 670–684 (2025).

34. P. F. Jacobs, R. Pollice, Developing large language models for quantum chemistry simulation input generation. *Digital Discovery* **4** (3), 762–775 (2025).

35. A. Yang, *et al.*, Qwen2 technical report. *arXiv preprint arXiv:2407.10671* (2024).

36. G. Team, Gemma (2024), doi:10.34740/KAGGLE/M/3301, https://www.kaggle.com/m/3301.

37. C. Snell, J. Lee, K. Xu, A. Kumar, Scaling llm test-time compute optimally can be more effective than scaling model parameters. *arXiv preprint arXiv:2408.03314* (2024).

38. M. Nakadai, S. Saito, H. Yamamoto, Diversity-based strategy for discovery of environmentally benign organocatalyst: diamine–protonic acid catalysts for asymmetric direct aldol reaction. *Tetrahedron* **58** (41), 8167–8177 (2002).

39. L.-C. Xu, *et al.*, Enantioselectivity prediction of pallada-electrocatalysed C–H activation using transition state knowledge in machine learning. *Nature Synthesis* **2** (4), 321–330 (2023).

40. G.-M. Chen, Z.-H. Ye, Z.-M. Li, J.-L. Zhang, Universal descriptors of quasi transition states for small-data-driven asymmetric catalysis prediction in machine learning model. *Cell Reports Physical Science* **5** (7) (2024).

41. A. E. Cuomo, *et al.*, Feed-forward neural network for predicting enantioselectivity of the asymmetric Negishi reaction. *ACS Central Science* **9** (9), 1768–1774 (2023).

42. F. Juliá-Hernández, T. Moragas, J. Cornella, R. Martin, Remote carboxylation of halogenated aliphatic hydrocarbons with carbon dioxide. *Nature* **545**, 84–88 (2017), letter.

43. F. Chen, *et al.*, Remote Migratory Cross-Electrophile Coupling and Olefin Hydroarylation Reactions Enabled by in Situ Generation of NiH. *Journal of the American Chemical Society* **139** (39), 13929–13935 (2017).

44. Y. He, Y. Cai, S. Zhu, Mild and Regioselective Benzylic C–H Functionalization: Ni-Catalyzed Reductive Arylation of Remote and Proximal Olefins. *Journal of the American Chemical Society* **139** (3), 1061–1064 (2017).

45. L. Peng, *et al.*, Ligand-Controlled Nickel-Catalyzed Reductive Relay Cross-Coupling of Alkyl Bromides and Aryl Bromides. *ACS Catalysis* **8** (1), 310–313 (2018).

46. L. Peng, Z. Li, G. Yin, Photochemical Nickel-Catalyzed Reductive Migratory Cross-Coupling of Alkyl Bromides with Aryl Bromides. *Organic Letters* **20** (7), 1880–1883 (2018).

47. Y. Li, *et al.*, Reaction scope and mechanistic insights of nickel-catalyzed migratory Suzuki–Miyaura cross-coupling. *Nature Communications* **11** (1), 417 (2020).

48. G. A. Landrum, RDKit: Open-source cheminformatics., `https://www.rdkit.org`.

49. N. M. O'Boyle, *et al.*, Open Babel: An open chemical toolbox. *Journal of cheminformatics* **3** (1), 33 (2011).

50. A. H. Larsen, *et al.*, The atomic simulation environment—a Python library for working with atoms. *Journal of Physics: Condensed Matter* **29** (27), 273002 (2017).

51. N. M. O'boyle, A. L. Tenderholt, K. M. Langner, Cclib: a library for package-independent computational chemistry algorithms. *Journal of computational chemistry* **29** (5), 839–845 (2008).

52. T. Lu, F. Chen, Multiwfn: A multifunctional wavefunction analyzer. *Journal of computational chemistry* **33** (5), 580–592 (2012).

53. C. Bannwarth, S. Ehlert, S. Grimme, GFN2-xTB—an Accurate and Broadly Parametrized Self-Consistent Tight-Binding Quantum Chemical Method with Multipole Electrostatics and Density-Dependent Dispersion Contributions. *Journal of Chemical Theory and Computation* **15** (3), 1652–1671 (2019).

54. R. Bjornsson, ASH: a multi-scale, multi-theory modelling program. *ChemRxiv* (2026), preprint (v1, Jan 28, 2026).

55. T. A. Young, J. J. Silcock, A. J. Sterling, F. Duarte, autodE: Automated Calculation of Reaction Energy Profiles – Application to Organic and Organometallic Reactions. *Angewandte Chemie International Edition* **60** (8), 4266–4274 (2021).

56. P. J. Stephens, F. J. Devlin, C. F. Chabalowski, M. J. Frisch, Ab initio calculation of vibrational absorption and circular dichroism spectra using density functional force fields. *The Journal of physical chemistry* **98** (45), 11623–11627 (1994).

57. W. J. Hehre, R. Ditchfield, J. A. Pople, Self—Consistent Molecular Orbital Methods. XII. Further Extensions of Gaussian—Type Basis Sets for Use in Molecular Orbital Studies of Organic Molecules. *The Journal of Chemical Physics* **56** (5), 2257–2261 (1972).

58. P. C. Hariharan, J. A. Pople, The influence of polarization functions on molecular orbital hydrogenation energies. *Theoretica chimica acta* **28** (3), 213–222 (1973).

59. V. Barone, M. Cossi, Continuous surface charge polarizable continuum models of solvation. I. General formalism. *The Journal of Chemical Physics* **132** (11), 114110 (2010).

60. Y. Zhao, D. G. Truhlar, The M06 suite of density functionals for main group thermochemistry, thermochemical kinetics, noncovalent interactions, excited states, and transition elements: two new functionals and systematic testing of four M06-class functionals and 12 other functionals. *Theoretical chemistry accounts* **120** (1), 215–241 (2008).

61. F. Weigend, R. Ahlrichs, Balanced basis sets of split valence, triple zeta valence and quadruple zeta valence quality for H to Rn: Design and assessment of accuracy. *Physical Chemistry Chemical Physics* **7** (18), 3297–3305 (2005).

62. S. Grimme, S. Ehrlich, L. Goerigk, Effect of the damping function in dispersion corrected density functional theory. *Journal of computational chemistry* **32** (7), 1456–1465 (2011).

63. A. V. Marenich, C. J. Cramer, D. G. Truhlar, Universal solvation model based on solute electron density and on a continuum model of the solvent defined by the bulk dielectric constant and atomic surface tensions. *The Journal of Physical Chemistry B* **113** (38), 12365–12376 (2009).

64. C. Adamo, D. Jacquemin, The calculations of excited-state properties with Time-Dependent Density Functional Theory. *Chemical Society Reviews* **42** (3), 845–856 (2013).

65. J.-D. Chai, M. Head-Gordon, Long-Range Corrected Hybrid Density Functionals with Damped Atom–Atom Dispersion Corrections. *Physical Chemistry Chemical Physics* **10** (44), 6615–6620 (2008).

# Supplementary Materials for

# Autonomous mechanistic discovery through agentic reasoning and validation

## Contents

# A Supplementary Methods

## A.1 Extended ARCHE architecture and execution flow

### A.1.1 Agent-level architecture

ARCHE is organized as a closed-loop reasoning–computation system in which a scientific question is progressively transformed into retrievable context, competing mechanistic propositions, executable computational workflows, and evidence-grounded conclusions (Supplementary Fig. S1). Starting from a user-provided query, the system first retrieves external and local background knowledge relevant to the chemical problem. This contextual information is then used to formulate multiple mechanistic hypotheses. The resulting hypotheses are converted into explicit computational workflows, which are executed through external chemistry tools and parsed into structured outputs. These outputs are subsequently examined in a reflection stage, where the system assesses whether the computed evidence supports, refines or rejects the proposed mechanism. In this way, ARCHE couples scientific reasoning to explicit computation in an iterative loop: retrieval informs hypothesis generation, hypotheses determine planning, planning enables execution, and execution supplies the evidence required for reflection and subsequent revision.

The key architectural principle is that these functions are distributed across components with distinct responsibilities rather than being handled by a single agent. The Retrieval Agent receives the scientific question as input and returns contextual objects, including literature-derived summaries, mechanistic precedents and methodological cues. The Hypothesis Agent takes these contextual objects as input and produces a set of competing mechanistic propositions formulated as testable scientific claims. The Planner Agent then maps each proposition to an executable workflow composed of ordered computational operations with defined inputs, outputs and dependencies. The Execution Module receives only the validated workflow specification and associated data objects, performs the requested calculations, and returns structured computational outputs such as geometries, energies, frequencies and diagnostic metadata. It does not perform scientific judgement. Reflection operates on the structured outputs together with the original propositions and determines whether the current evidence is sufficient, whether the workflow should be revised, or whether alternative mechanistic propositions should be prioritized. This separation of input and output types clarifies that ARCHE is not a monolithic language model, but a coordinated system in which reasoning, workflow construction, computation and interpretation are explicitly partitioned.

**Retrieval Agent.** The Retrieval Agent establishes the literature and knowledge context for subsequent reasoning and decision-making. Starting from the user's scientific question, and incorporating reaction descriptions, molecular representations and experimental conditions when available, it identifies the chemically relevant elements of the problem, such as substrates, catalysts, reaction classes, plausible intermediates and computation-related keywords. It then uses these cues to gather relevant literature and local knowledge records, which are organized into a searchable retrieval-augmented knowledge context for subsequent querying and use by other agents.

The Retrieval Agent provides the background knowledge on which later reasoning steps depend, including prior reaction patterns, mechanistic precedents, experimentally or computationally relevant observations, and commonly used computational strategies. This curated context supports subsequent hypothesis generation, workflow design and result interpretation, while remaining distinct from the hypotheses and conclusions derived from it.

**Hypothesis Agent.** The Hypothesis Agent formulates testable mechanistic hypotheses by integrating the retrieved literature with the assembled knowledge context. Taking as input the scientific question together with the background assembled by the Retrieval Agent, it examines the problem from multiple chemically meaningful perspectives and formulates alternative hypotheses that capture different ways the transformation might proceed. These alternatives may differ in the identity of the reactive species involved, the ordering of elementary steps, the nature of the key bond-forming or bond-breaking event, or the physical origin of the observed selectivity or reactivity.

The role of this stage is not merely to enumerate possibilities, but to organize them into a scientifically useful hypothesis space. Candidate hypotheses are expressed in forms that can be evaluated in later stages, such as proposed reaction pathways, competing activation modes, alternative transition-state models or distinct mechanistic scenarios. The agent also considers the relationships among hypotheses, for example whether two proposals are independent, partially complementary or genuinely contradictory. This comparison allows closely related ideas to be refined, merged or reformulated before downstream testing, rather than being treated as isolated and unstructured outputs.

The resulting hypothesis set is therefore a ranked and refined collection of candidate hypothese rather than a final conclusion. In prioritizing hypotheses, the agent considers criteria such as scientific soundness, chemical relevance, feasibility, clarity of logic and, where appropriate, novelty. The highest-priority hypotheses are then passed to subsequent stages for computational planning and evaluation. In this way, the Hypothesis Agent defines the space of plausible scientific hypothese that the system will test, while leaving the design of specific computational workflows to later components.

**Planner Agent.** The Planner Agent translates a selected scientific hypothesis into a concrete computational plan. Its inputs include the prioritized hypothesis, contextual data, and the structured tool registry that specifies the operations accessible to the system. It first identifies the scientific questions that must be resolved to evaluate the hypothesis, and then converts them into an ordered set of computational operations. Depending on the problem, these may include structure preparation, conformer sampling, transition-state search, geometry optimization, frequency analysis, IRC calculations and single-point energy evaluation.

The output of planning is an explicit workflow specification comprising the computational steps, their dependencies, the tools required for each operation, and the expected inputs and intermediate outputs across the workflow. The planner also checks whether all required tools are available in the current registry; if necessary capabilities are missing, these gaps are identified before execution, recorded as unmet requirements, and reported to the user so that the missing tools can be provided.

Before execution, the generated workflow is reviewed in two stages. A general-purpose reasoning model first checks the logical coherence of the workflow and verifies the consistency of its stepwise inputs, intermediate outputs and dependencies. ARCHE-Chem then performs a domain-specialized review of the proposed computational settings, assessing whether the selected methods and Gaussian keywords, where relevant, are chemically appropriate and technically well formed. The Planner Agent therefore serves as the bridge between scientific reasoning and executable computation by determining how a selected hypothesis should be tested.

**Execution Module.** The Execution Module executes the validated computational workflow by connecting the planned computational steps to external computational chemistry tools. Its input is the validated workflow specification, which defines the sequence of operations and the required data flow across the workflow. Guided by this specification, the module submits and monitors the requested calculations, manages execution status, and records the files generated at each step.

The direct outputs of execution are the computational results and software logs produced by the workflow. Depending on the calculation sequence, these may include optimized geometries, electronic energies, thermodynamic quantities, vibrational frequencies, IRC trajectories, convergence diagnostics and Gaussian log files. These outputs are subsequently parsed into structured, machine-readable records for downstream use.

Parsed results then enter a result-analysis and reflection process. ARCHE-Chem and the generalist model jointly examine the returned energies, frequencies, thermodynamic quantities and related outputs to assess their implications for the hypothesis under evaluation. When Gaussian calculations fail, a first reflection step diagnoses the error and supports correction, for example by revising computational settings or, where appropriate, suggesting structure-level changes for user intervention. When calculations complete successfully but the resulting evidence remains inconsistent with experiment, a second reflection step feeds this mismatch back to the Hypothesis Agent so that alternative hypothese can be reconsidered. The Execution Module is therefore embedded in an iterative loop of calculation, parsing, evaluation and revision, rather than serving as a terminal execution endpoint.

### A.1.2 Execution flow and feedback routing

This section describes how validated computational workflows are executed in practice and how execution outcomes are routed back to upstream components when revision is required. Communication across the closed loop is mediated through structured intermediate records representing hypotheses, workflow specifications and parsed computational results. After passing the logical and domain-specialized pre-execution reviews described above, a workflow is transferred to the Execution Module, where the corresponding input files are prepared, calculations are submitted to the computational environment, and execution status is monitored until completion, interruption or failure. Gaussian log files and related outputs are then parsed to extract structured results, including geometries, energies, vibrational frequencies, thermodynamic quantities, convergence diagnostics and other calculation-specific metadata. These parsed results are subsequently passed to the downstream analysis and reflection process.

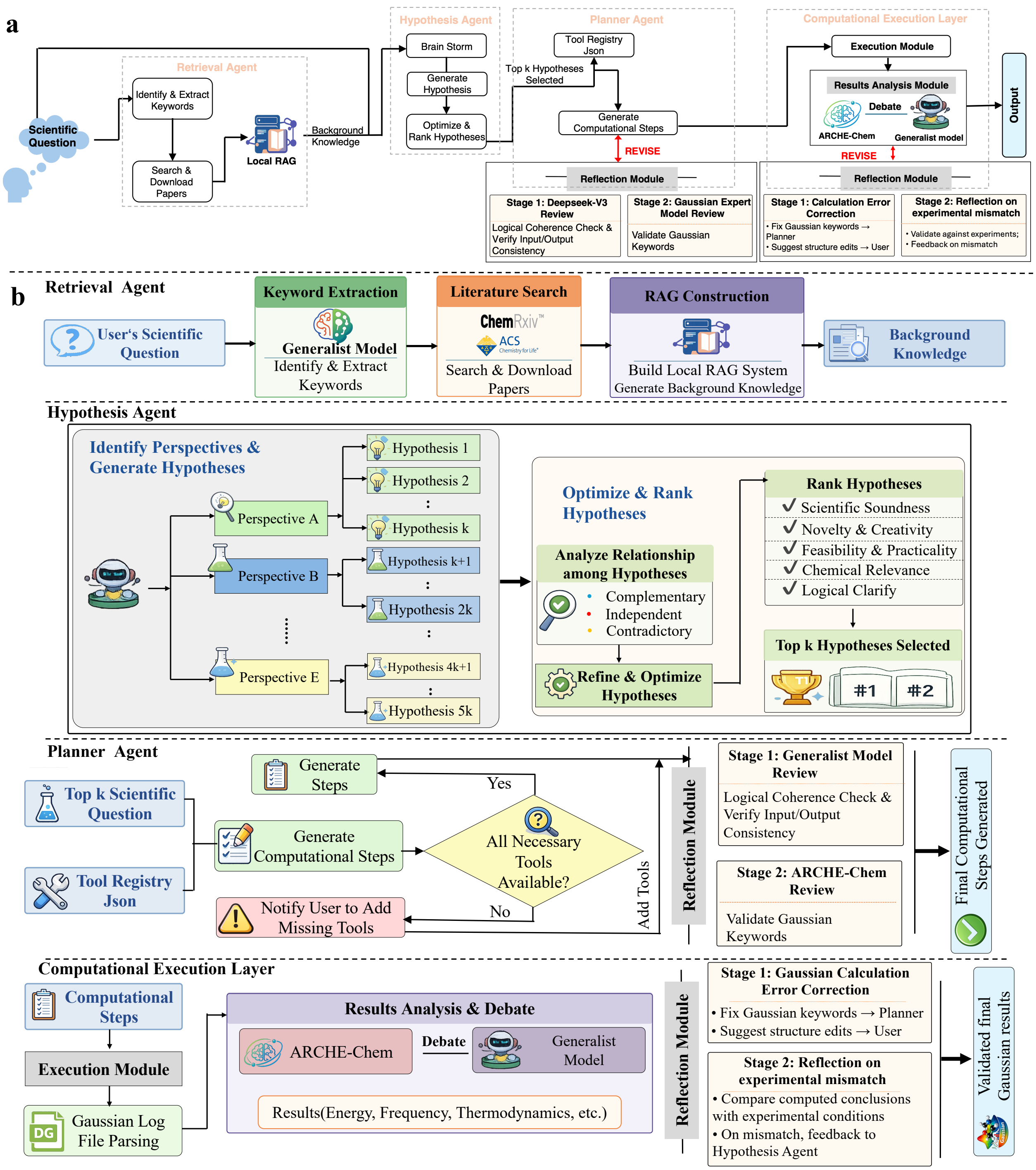

**Figure S1**: **Dual-model, tool-aware multi-agent architecture for quantum-chemistry workflows.** (a) System overview; (b) Agent-level workflow.

The main feedback decisions arise after computational outputs have been produced and examined. One class of outcomes involves technical failure, in which a valid and interpretable computational result cannot be obtained. This includes, for example, invalid Gaussian keywords, inappropriate computational settings, malformed input generation, failed convergence, missing required calculations, or other execution problems that prevent the workflow from producing usable evidence. In such cases, the system identifies whether the problem originates from the computational procedure or from the supplied structural inputs. Problems associated with the computational procedure are routed back to the planning component so that the workflow and calculation settings can be revised. By contrast, failures caused by incomplete, defective or chemically unsuitable starting structures are reported for user intervention, because continued autonomous execution would not be reliable without corrected structural input.

A second class of outcomes arises when calculations complete successfully and yield analyzable results, but the interpreted evidence remains inconsistent with the available experimental observations or with the candidate explanation under evaluation. In these cases, the parsed results and their interpretation are returned directly to the Hypothesis Agent. The purpose of this feedback is to allow the system to reconsider the current explanation, reformulate alternative mechanistic possibilities and reprioritize competing hypotheses in light of the new computational evidence.

More generally, the closed loop distinguishes between technical correction and scientific revision. Feedback to the planning component is used when the current computational workflow is invalid, incomplete or otherwise unable to generate reliable computational evidence. Feedback to the Hypothesis Agent is used when the workflow executes successfully but the resulting evidence does not support the candidate explanation being tested. User input is requested only when the system lacks the structural completeness or tool support required to proceed autonomously. In this way, the execution layer functions not only as the site of calculation, but also as the stage at which computed evidence is classified and routed back to the appropriate upstream component for further revision.

### A.1.3 Structured inter-agent communication

Communication across ARCHE is organized through structured intermediate representations that transmit scientific hypotheses, computational plans and interpreted calculation results between components. Each stage of the reasoning–computation cycle is captured in a representation with a defined scientific role, including a hypothesis representation, a workflow specification, an execution record, a parsed result summary and a reflection summary. Together, these representations provide an explicit record of what has been proposed, what has been executed, what evidence has been obtained and how that evidence informs subsequent steps.

The hypothesis representation records a candidate mechanistic hypothesis in a structured form suitable for downstream computational evaluation. It typically contains a textual description of the proposed mechanism, the underlying scientific question, relevant evidence from retrieval, the specific mechanistic proposition to be tested and the computational objectives defined for its assessment. The output of the Hypothesis Agent is therefore not a final conclusion, but a structured candidate explanation that defines the scientific possibility under consideration. The corresponding workflow specification, produced by the Planner Agent, translates this candidate explanation into an explicit computational plan. It defines the ordered computational steps, the tools associated with each step, the required inputs and expected intermediate outputs, and the dependencies linking one operation to the next.

During execution, communication remains similarly structured. The execution record captures the operational status of each step, including the selected tool, generated input files, job status, produced outputs and any reported errors. After execution, Gaussian log files and related outputs are converted into a parsed result summary containing standardized scientific information such as optimized geometries, electronic energies, free energies, vibrational frequencies, thermodynamic quantities, convergence information and provenance details linking the result to the originating workflow step. These structured outputs provide the basis for downstream analysis. The reflection summary then records the outcome of post-execution evaluation, including whether the returned evidence supports the intended computational test, whether the computational plan should be revised, whether the candidate explanation should be reconsidered or whether additional user input is required. In this way, reflection serves not only to interpret results, but also to determine how the next cycle of revision should proceed.

This communication scheme is important because ARCHE coordinates multiple components with distinct scientific roles and technical functions. The general-purpose reasoning model, ARCHE-Chem, the Planner Agent and the Execution Module each contribute different forms of reasoning and evaluation. By exchanging hypotheses, workflow specifications, execution records and parsed results in a common structured form, the system ensures that all components operate on the same evolving scientific context even though their responsibilities differ. This reduces ambiguity when information is passed between stages and helps distinguish whether a problem arises from an unsupported hypothesis, an inadequate computational plan, an execution failure or a mismatch between computed evidence and the explanation under evaluation.

Structured communication is also important for traceability and reproducibility. Because each stage receives and produces explicit intermediate representations, the progression from scientific question to candidate hypotheses, from hypotheses to computational workflow, and from workflow to computational evidence can be reconstructed and examined afterwards. This makes it possible to identify which hypothesis led to a given calculation, which workflow step produced a particular Gaussian result and which reflection outcome triggered a rerun or hypothesis revision. Such traceability is valuable not only for diagnosing errors, but also for ensuring scientific reproducibility, because the reasoning pathway and the computational pathway remain explicitly connected rather than being reduced to an opaque sequence of generated text. In this way, structured inter-agent communication provides the basis for modular coordination, transparent iteration and evidence-grounded scientific analysis.

#### A.1.4 Cross-stage consistency

ARCHE maintains cross-stage consistency by enforcing explicit links between the scientific question, the mechanistic hypotheses, the computational workflows used to evaluate them, the executed results returned by those workflows, and the final conclusion. No downstream stage is allowed to introduce a scientific claim that cannot be traced to an upstream hypothesis object and to either a defined computational test or executed computational evidence. This constrains the system to remain on the original scientific target rather than drifting to a related but unevaluated question.

Each hypothesis must map to a defined computational test. The planner enforces this requirement by converting a selected mechanistic proposition into a workflow intended to evaluate that specific claim. Reflection then checks consistency at two levels. Before or during execution, it assesses whether the workflow remains a coherent and chemically valid test of the original proposition. After execution, it assesses whether the returned evidence supports, weakens or fails to resolve that proposition. Workflow revision therefore preserves the original hypothesis linkage unless the scientific proposition itself is explicitly changed.

Final conclusions are formed only by integrating evidence returned from executed workflows and by interpreting that evidence against the original scientific question and available experimental facts. Retrieval-derived context and model priors may guide hypothesis generation and workflow construction, but they are not treated as evidence in the final conclusion. If the available results are incomplete, technically invalid or non-diagnostic, the system records that limitation rather than promoting an unsupported interpretation.

Consistency is further maintained through provenance tracking across revisions. Revised workflows retain their link to the originating hypothesis, and revised hypotheses are annotated with the reason for change, such as workflow invalidation, calculation failure, contradictory evidence or insufficient support. This preserves an auditable chain from question to conclusion and prevents unsupported intermediate reasoning from being presented as an established mechanistic finding.

## A.2 ARCHE-Chem development, evaluation and role in the system

### A.2.1 Training pipeline overview

ARCHE-Chem was developed through a three-stage training pipeline designed not only to improve standalone performance on computational chemistry tasks, but also to provide a reliable expert layer within the full ARCHE system. Starting from the Qwen2.5-7B-Instruct base model, the model was first adapted to the computational chemistry domain through continual pretraining on a curated corpus of textbooks, software documentation and expert discussions. This stage equipped the model with the terminology, conceptual background and workflow conventions needed to interpret mechanistic problems and interact meaningfully with professional computational chemistry procedures.

Following continual pretraining, the model was further refined through supervised fine-tuning

(SFT) on a domain-specific instruction dataset constructed from the curated corpus. The dataset spans multiple categories that are directly relevant to ARCHE operation, including theoretical knowledge, software knowledge, Gaussian route generation, output interpretation and error diagnosis. These capabilities are important not only as benchmark tasks in their own right, but because they support the system functions required for reliable planning and reflection. Within ARCHE, ARCHE-Chem is used to audit whether Gaussian settings are appropriate, whether returned outputs are scientifically interpretable, and whether observed failures should be attributed to route design, execution conditions or the mechanistic proposal itself.

To improve response quality at inference time, a reward model was trained on preference pairs derived from candidate responses to domain-specific prompts and subsequently reviewed by human experts. This preference curation favors responses that reflect chemically valid reasoning, correct software practice and diagnostically useful interpretation. During inference, ARCHE-Chem uses a Best-of-N test-time scaling strategy ($N = 8$), in which multiple candidate responses are generated and ranked by the reward model. In the full ARCHE architecture, this selection mechanism is important because the model is not used simply to answer chemistry questions in isolation; rather, it improves the reliability of domain-specialized support during workflow review, execution-related analysis and post-computation reflection across the reasoning–computation loop.

### A.2.2 Pretraining corpus composition

The continual pretraining corpus was assembled from publicly available computational chemistry resources selected to cover complementary forms of domain knowledge (Tab. S1). The corpus combines textbooks, official Gaussian documentation and expert discussions from online forums, because these sources capture different but jointly necessary aspects of computational chemistry practice. Textbooks provide the conceptual and methodological foundation of the field, including electronic-structure theory, reaction analysis and standard computational protocols. Official Gaussian documentation supplies software-specific knowledge, such as keyword usage, input conventions and calculation settings. Expert forum discussions contribute practice-oriented knowledge that is often underrepresented in formal references, including troubleshooting strategies, workflow heuristics and common sources of calculation failure. All documents were deduplicated before training.

This composition was designed to support the role of ARCHE-Chem within the full ARCHE system. In ARCHE, the model must not only recognize domain terminology, but also evaluate workflow validity, propose appropriate computational settings, diagnose execution failures and interpret returned results in a chemically meaningful way. The pretraining corpus was therefore constructed to expose the model to both formal chemical knowledge and the procedural knowledge required for real computational workflows. This composition was designed to support the role of ARCHE-Chem within the full ARCHE system. In ARCHE, the model must not only recognize domain terminology, but also evaluate workflow validity, propose appropriate computational settings, diagnose execution failures and interpret returned results in a chemically meaningful way. The pretraining corpus was therefore constructed to expose the model to both formal chemical knowledge and the practical knowledge required for real computational workflows. By combining foundational knowledge, guidance on computational chemistry practice and troubleshooting materials, the corpus more closely reflects the knowledge needed for real computational chemistry reasoning.

**Table S1**: **Document sources used to construct the ARCHE-Chem corpus.** The training corpus integrates textbooks, software documentation and expert forum discussions covering computational chemistry workflows and troubleshooting practices.

| **Source type** | **Description** | **Documents** |
|---|---|---:|
| Textbooks | Computational chemistry textbooks | 105 |
| Manuals | Gaussian software documentation | 1 |
| Forum discussions | Expert troubleshooting and methodological discussions | 432 |

### A.2.3 Instruction dataset composition

The supervised fine-tuning (SFT) dataset was designed to support the expert functions of ARCHE-Chem within the full ARCHE framework, rather than to serve only as an isolated benchmark set. Accordingly, the dataset was organized around four task categories that reflect the main capabilities required across the reasoning–planning–execution–reflection loop: computational chemistry and Gaussian basics, Gaussian route generation, result analysis and error handling.

The *computational chemistry and Gaussian basics* category was constructed from computational chemistry textbooks, teaching materials, Gaussian official documentation and related reference resources. It covers foundational concepts in quantum chemistry and computational chemistry, together with the basic knowledge required to understand how calculations are specified in Gaussian. This includes theoretical principles, method characteristics, the meanings and intended uses of common Gaussian keywords, and the chemical significance of routine computational settings. This category therefore provides the conceptual basis for method selection, workflow construction and the chemically meaningful interpretation of computed results.

The *Gaussian route generation* category was constructed from Gaussian input examples provided in the official documentation together with curated keywords examples and templates accumulated by the authors through long-term computational chemistry research. This category focuses on the generation of chemically and technically appropriate Gaussian keywords sections for a broad range of computational tasks, including, but not limited to, geometry optimization, frequency analysis, transition-state search, intrinsic reaction coordinate calculations and single-point energy evaluation. It also covers the selection and combination of methods, basis sets and relevant control keywords required to define executable Gaussian calculations. Because this category directly affects whether a scientific hypothesis can be translated into a valid computational test, it represents one of the most execution-critical components of the SFT dataset.

The *result analysis* and *error handling* categories were compiled mainly from common issues encountered by the authors in practical computational chemistry studies, supplemented by representative discussions from expert forums and recurrent cases documented in Gaussian manuals and related technical resources. The result-analysis subset focuses on interpretation of Gaussian outputs, including energies, frequencies, convergence behaviour, structural information and reaction-path results. The error-handling subset focuses on diagnosis of failed calculations, identification of likely causes and proposal of corrective actions, such as keyword adjustment, changes in computational settings or revisions to the computational strategy.

Together, these four task categories were chosen to cover both foundational knowledge and execution-critical support. computational chemistry and Gaussian basics provide theoretical and methodological grounding, whereas Gaussian route generation, result analysis and error handling more directly support practical workflow reliability within ARCHE. Representative examples from each category are provided in Supplementary Table S4.

**Table S2**: **Task categories and sample distribution in the ARCHE-Chem training corpus.** The instruction dataset is organized into four task categories covering conceptual knowledge, reasoning, route generation, error handling and result analysis.

| **Task category** | **Description** | **Train** | **Test** |
|---|---|---|---|
| computational chemistry and Gaussian basics | Foundational concepts and terminology in computational chemistry, together with basic Gaussian knowledge and keyword interpretation | 24,267 | 3,000 |
| Gaussian route generation | Generation of valid Gaussian route sections | 3,899 | 500 |
| Gaussian error handling | Diagnosis and correction of computational errors | 807 | 200 |
| Gaussian results analysis | Interpretation of Gaussian output files | 9,968 | 1,000 |

#### A.2.4 Data independence and contamination control

To ensure a fair assessment, no information from the evaluation scenarios—including mechanistic hypotheses, transition-state structures, computational procedures or energetic results—was included in the training data of ARCHE-Chem or the reward model. The training corpus consists of general computational chemistry knowledge, textbooks, software documentation and expert discussions, rather than task-specific reaction systems or benchmark instances.

All reported mechanistic conclusions are derived from explicit quantum-chemical calculations performed during inference, rather than retrieved from training data. Comparisons with external models were conducted under identical input conditions to ensure consistency and fairness.

#### A.2.5 Training configuration

Continual pretraining and supervised fine-tuning were performed using PyTorch with mixed-precision training. Key training parameters are summarized in Supplementary Tab. S3.

**Table S3**: **Training hyperparameters used for ARCHE-Chem model development.**

| Training stage | Parameter | Value |
|---|---|---|
| Continual pretraining | Batch size (per device) | 1 |
| | Gradient accumulation | 4 |
| | Epochs | 2 |
| | Learning rate | $5 \times 10^{-5}$ |
| | Warmup steps | 100 |
| | Precision | FP16 |
| Supervised fine-tuning | Batch size (per device) | 4 |
| | Gradient accumulation | 4 |
| | Epochs | 2 |
| | Learning rate | $5 \times 10^{-5}$ |
| | Warmup steps | 100 |

#### A.2.6 Reward model training and Best-of-N inference

A reward model based on Gemma-2-27B-IT was trained using 28,000 preference pairs constructed from candidate responses generated under domain-specific prompts and subsequently curated through expert review. In each pair, responses were ranked not only according to whether they resembled expert chemical reasoning, but also according to whether they would function more reliably as execution-facing outputs within ARCHE. Preferred responses were therefore those that better preserved chemical validity, proposed more appropriate Gaussian workflows, and were less likely to produce technically invalid, internally inconsistent or non-executable inputs. During reward model training, optimization was performed using the AdamW optimizer with a cosine learning-rate schedule, a maximum sequence length of 4096 tokens, a learning rate of $2 \times 10^{-6}$, and a warmup ratio of 0.03 for three training epochs.

During inference, ARCHE-Chem employs a Best-of-N sampling strategy in which multiple candidate responses are generated and ranked by the reward model, and the highest-scoring candidate is selected as the final output. In the context of ARCHE, this procedure is important because it biases selection toward outputs that are more executable, more chemically coherent and less likely to induce downstream Gaussian failures. Best-of-N inference therefore serves as a reliability layer for planning and validation, reducing the probability that a superficially plausible but executionally fragile response is propagated into the workflow.

**Table S4**: **Task categories used in ARCHE-Chem training and evaluation.** The four categories were designed to cover both foundational knowledge and execution-critical capabilities required for practical computational chemistry support within ARCHE.

| Task category | Primary sources | Main content | Primary role in ARCHE |
| --- | --- | --- | --- |
| computational chemistry and Gaussian basics | Computational chemistry textbooks, teaching materials, Gaussian official documentation and related reference resources | Foundational concepts in quantum chemistry and computational chemistry; method characteristics; chemically meaningful interpretation of common computational approaches; meanings and intended uses of Gaussian keywords; route-section components; practical interpretation of common computational settings | Provides the conceptual basis for method selection, computational setup and the chemically meaningful interpretation of computed results. |
| Gaussian route generation | Gaussian official examples and curated input templates accumulated through the authors' research practice | keywords sections for common tasks such as optimization, frequency analysis, TS search, IRC and single-point refinement | Directly supports translation of mechanistic hypotheses into executable computational tests |

Continued on next page

| Task category | Primary sources | Main content | Primary role in ARCHE |
|---|---|---|---|
| Gaussian Result analysis | Author-curated cases from practical computational chemistry, supplemented by expert forums and Gaussian-related technical resources | Interpretation of Gaussian outputs, including energies, frequencies, convergence behaviour, structural results and reaction-path information | Supports evaluation of computed evidence during reflection and mechanistic assessment |
| Gaussian Error handling | Author-curated failure cases from practical computational chemistry, supplemented by expert forums and Gaussian-related technical resources | Diagnosis of failed calculations; identification of likely causes; proposal of corrective actions including keyword adjustment and strategy revision | Supports workflow robustness by enabling repair of failed or inconsistent calculations |

## A.3 Tool registry specification

### A.3.1 Tool registry schema

To translate candidate scientific hypotheses into executable computational procedures, ARCHE uses a structured tool registry that defines the computational operations available to the Planner Agent. This registry provides the practical link between scientific reasoning and computational chemistry execution by describing what each tool does, what inputs it requires and what outputs it produces.

Within the ARCHE architecture, the Hypothesis Agent proposes candidate hypotheses, and the Planner Agent converts selected hypotheses into computational plans. To support this process, the tool registry organizes available tools into two broad categories: atomic tools and pipeline tools. Atomic tools perform single, well-defined operations, such as file format conversion or Gaussian input generation. Pipeline tools corre-spond to relatively complex but single-purpose computational procedures that are sufficiently standardized to be defined in the registry as complete procedures. Examples include conformer exploration. Representing such procedures in this way helps improve the reliability and accuracy of tool selection during planning.

Each tool is represented in the registry through a standardized description that specifies its role, required inputs, produced outputs and any relevant operational parameters or constraints. This information allows the Planner Agent to assemble workflows whose individual steps are clear in purpose, compatible in data flow and executable in practice.

A simplified example of the schema used for registry entries is shown below:

***Example of Tool Schema***

```
[
  {
    "tool_name": "convert_sdf_to_xyz",
    "description": "Convert a molecular structure file from SDF format to XYZ
         format",
    "inputs": {
      "structure": "SDF"
    },
    "outputs": {
      "structure": "XYZ"
    },
    "parameters": {}
  }
]
```

### A.3.2 Representative registry entries

The ARCHE registry includes both atomic utilities and higher-order workflow tools. Atomic utilities carry out single, well-defined operations, such as structure conversion, Gaussian input construction, job execution and output parsing. Workflow tools package recurrent multi-step procedures that are frequently needed in mechanistic studies, including conformer preparation and transition-state initialization. This organization allows the planner to retain explicit control when fine-grained tool composition is needed, while also reusing stable multi-step routines for recurrent computational patterns. Representative entries from the registry are listed below.

**Table S5: Representative entries in the ARCHE tool registry.**

| Tool name | Type | Function |
|---|---|---|
| `smiles_to_sdf` | Atomic | Convert SMILES strings into three-dimensional molecular structures |
| `generate_conformers` | Atomic | Generate conformer ensembles for reactants, intermediates or candidate complexes |
| `sdf_to_xyz` | Atomic | Convert structure files into Cartesian coordinates for downstream quantum-chemical tasks |
| `xyz_to_gjf` | Atomic | Convert molecular coordinates into Gaussian input files |
| `generate_gaussian_route` | Atomic | Construct Gaussian route sections and related input settings for specific computational objectives |
| `run_gaussian` | Atomic | Execute Gaussian calculations |
| `parse_gaussian_log` | Atomic | Extract energies, geometries, frequencies and convergence information from Gaussian output files |
| `build_reactive_complex` | Pipeline | Assemble candidate reactive complexes or initial structures for downstream transition-state exploration |
| `ts_initialization` | Pipeline | Generate starting structures for transition-state searches from mechanistic hypotheses and structural inputs |

### A.3.3 Workflow construction mechanism

The Planner Agent constructs computational workflows by querying the registry for tools whose capabilities match the requirements of a given mechanistic hypothesis. Hypotheses are interpreted as computational objectives, which are decomposed into ordered operations such as structure preparation, geometry optimization or electronic structure analysis.

During workflow construction, the planner verifies compatibility between successive tool invocations using the input-output specifications defined in the registry. When representation mismatches occur between intermediate outputs and required inputs, appropriate conversion tools are automatically inserted to mediate between data formats. The resulting workflow is represented as a directed graph in which nodes correspond to tool invocations and edges represent data dependencies between computational steps.

### A.3.4 Example generated workflow

Representative examples of a workflow generated by the planner is shown below for a structural descriptor evaluation task.

***Workflow of Case Study 1***

```
[
    {
        "Step_number": 1,
        "Description": "Generate initial transition state guess structures
             for the C-C bond formation step between the enolate and
            aldehyde using the main pipeline tool. This tool handles atom
            mapping, force-field optimization, and generates initial TS
            structures from SMILES inputs.",
        "Tool": "TSPipline",
        "Input": "Reactant SMILES: O=Cc1ccc(cc1)[N+](=O)[O-], CC(=O)C, C(F
            )(F)(F)S(=O)(=O)O, N1CCC[C@H]1C(N2CCCC2); Product SMILES: CC(=
            O)C(O)c1ccc([N+](=O)[O-])cc1",
        "Output": "XYZ file containing initial TS guess structures for
            both enantiomeric pathways"
    },
    {
        "Step_number": 2,
        "Description": "Generate Gaussian keywords and route section for
            TS optimization and frequency calculation.",
        "Tool": "GenerateGaussianCode",
        "Input": "TS optimization and frequency calculation.",
        "Output": "Gaussian keywords and route section"
    },
    {
        "Step_number": 3,
```

```
        "Description": "Convert XYZ TS guess structures to Gaussian input
           files (.gjf) using the generated keywords and route section",
        "Tool": "xyz_to_gjf",
        "Input": "XYZ file from Step 1 and Gaussian keywords from Step 2",
        "Output": "GJF files for TS optimization of both enantiomeric
           pathways"
    },
    {
        "Step_number": 4,
        "Description": "Perform Gaussian TS optimization and frequency
           calculation for both enantiomeric transition states.",
        "Tool": "Gaussian",
        "Input": "GJF files from Step 3",
        "Output": "Optimized TS geometries with frequency data (one
           imaginary frequency expected) and energies for both
           enantiomeric pathways"
    },
    {
        "Step_number": 5,
        "Description": "Generate Gaussian keywords and route section for
           IRC calculations to confirm connectivity of TS structures",
        "Tool": "generate_gaussian_code",
        "Input": "IRC calculation",
        "Output": "Gaussian keywords and route section"
    },
    {
        "Step_number": 6,
        "Description": "Convert optimized TS structures to Gaussian input
           files for IRC calculations",
        "Tool": "xyz_to_gjf",
```

```
        "Input": "Optimized TS XYZ structures from Step 4 and Gaussian
            keywords from Step 5",
        "Output": "GJF files for IRC calculations of both TS structures"
    },
    {
        "Step_number": 7,
        "Description": "Perform IRC calculations in Gaussian to confirm TS
            connectivity to reactants and products",
        "Tool": "Gaussian",
        "Input": "GJF files from Step 6",
        "Output": "IRC trajectories confirming connectivity of TS
            structures to correct reactants and products"
    },
    {
        "Step_number": 8,
        "Description": "Generate Gaussian keywords and route section for
            high-level single-point energy calculations.",
        "Tool": "generate_gaussian_code",
        "Input": "Single-point energy calculation",
        "Output": "Gaussian keywords and route section"
    },
    {
        "Step_number": 9,
        "Description": "Convert optimized TS structures to Gaussian input
            files for single-point energy calculations",
        "Tool": "xyz_to_gjf",
        "Input": "Optimized TS XYZ structures from Step 4 and Gaussian
            keywords from Step 8",
        "Output": "GJF files for single-point energy calculations of both
            TS structures"
```

```
        },
        {
            "Step_number": 10,
            "Description": "Perform high-level single-point energy
                calculations on optimized TS structures.",
            "Tool": "Gaussian",
            "Input": "GJF files from Step 9",
            "Output": "High-level electronic energies for both enantiomeric
                transition states"
        },
        {
            "Step_number": 11,
            "Description": "Calculate Gibbs free energy barriers (ΔG‡) at 30°C
                 from frequency calculations and single-point energies, then
                compute enantiomeric excess from the energy difference between
                 competing TS pathways",
            "Tool": "Python script",
            "Input": "Thermochemical data from Step 4 frequency calculations
                and electronic energies from Step 10",
            "Output": "ΔG‡ values for both enantiomeric pathways and predicted
                 enantiomeric excess (ee%)"
        }
]
```

These structured workflows enable the system to translate mechanistic hypotheses into reproducible computational procedures while maintaining explicit data dependencies and execution order across all computational steps.

## A.4 Computational protocols for value demonstrations

This section summarizes the computational protocols used in the three value-demonstration scenarios investigated in this study. These protocols define the quantum-chemical calculations performed by the ARCHE execution module during mechanistic evaluation. For each case study, representative Gaussian input settings are provided to illustrate the computational procedures used to evaluate candidate mechanisms.

### A.4.1 Representative Gaussian protocols sections

Representative Gaussian computational protocols used in the study (Tab. S6).

**Table S6**: **Representative Gaussian computational settings used in different case studies.**

| Case study | Computational task | Gaussian route |
|---|---|---|
| Asymmetric catalysis | Reactant and product optimization | `#p opt freq B3LYP/6-31G(d) scrf=(iefpcm,solvent=acetone) temperature=303.15` |
| Asymmetric catalysis | Transition state optimization | `#p B3LYP/6-31G(d) opt=(ts,noeigentest,calcall,maxstep=5) freq scrf=(iefpcm,solvent=acetone) temperature=303.15` |
| Asymmetric catalysis | IRC verification | `#p B3LYP/6-31G(d) irc=(calcfc,LQA,maxpoints=50,recalc=5) scrf=(iefpcm,solvent=acetone)` |
| Asymmetric catalysis | Single-point energy refinement | `#p M062X/def2TZVPP sp scrf=(iefpcm,solvent=acetone) int=ultrafine scf=tight` |
| Photochemical reaction | Geometry optimization | `#p B3LYP/def2SVP empiricaldispersion=gd3bj opt(calcfc) freq integral=ultrafine scrf=(smd,solvent=ethylethanoate)` |
| Photochemical reaction | Excited-state calculation | `#p td(nstates=20) B3LYP/def2SVP empiricaldispersion=gd3bj scrf=(smd,solvent=ethylethanoate) int=ultrafine` |
| Descriptor analysis | Geometry optimization | `#p opt freq wB97XD scrf=(smd,solvent=n,n-dimethylacetamide) def2SVP` |
| Descriptor analysis | Relaxed scan | `#p opt=modredundant wB97XD scrf=(smd,solvent=n,n-dimethylacetamide) def2SVP nosymm` |

### A.4.2 Asymmetric catalysis case study

For the asymmetric catalysis system, the computational workflow was designed not only to identify the relevant reaction pathway and stereocontrolling transition states, but also to assess whether the computed results were consistent with the conclusions and transition-state structures reported in the original study. Geometry optimizations and vibrational frequency calculations were first performed for reactants and products to obtain stable stationary points. Transition states were then located and confirmed by frequency analysis, followed by intrinsic reaction coordinate (IRC) calculations to verify connectivity to the corresponding reactant and product minima. Electronic energies were subsequently refined through higher-level single-point calculations to obtain an energetically reliable profile for comparison among competing transition structures.

Agreement with the literature was assessed at three levels. First, optimized transition-state geometries were compared with the reported structures by structural similarity metrics, including RMSD where direct comparison was possible. Second, the relative energetic ordering of competing stereocontrolling transition states was evaluated to determine whether the computed free-energy differences reproduced the reported selectivity trend. Third, the optimized transition states were examined for qualitative consistency with the mechanistic interpretation presented in the original study, including the key hydrogen-bonding pattern, substrate–catalyst arrangement and stereodifferentiating transition-state motif. In this way, agreement with the literature was assessed not only by successful location of transition states, but also by combined geometric, energetic and mechanistic consistency with the published conclusions and reported structures.

### A.4.3 Photochemical reaction case study

For the photochemical system, the computational analysis was designed not only to characterize candidate intermediates and excited states, but also to eliminate unsupported mechanistic hypotheses and converge on a pathway consistent with the available evidence. Geometry optimizations and vibrational frequency calculations were first performed to identify stable intermediates and confirm the absence of imaginary frequencies for putative ground-state species. Excited-state energies were then evaluated using Time-Dependent Density Functional Theory (TDDFT) calculations for mechanistic scenarios that required direct photoexcitation. These calculations were interpreted together with the experimentally reported reaction conditions and control observations, so that each hypothesized pathway could be assessed against both computed energetics and observable reactivity constraints.

Unsupported pathways were excluded through an elimination logic based on whether their key mechanistic requirements were compatible with the computed and experimental evidence. Pathways that required energetically inaccessible elementary steps under the reported conditions were rejected on energetic grounds. Hypotheses invoking direct photoexcitation of isolated species were disfavoured when the calculated excitation energies were inconsistent with the energy available from 450 nm irradiation. Likewise, closed-shell two-electron pathways were excluded when the computed barriers were too high to account for the observed room-temperature reactivity. The retained pathway was therefore not selected simply because it could be computed, but because it remained the one most consistent with the full evidence chain, including feasible computed energetics, photophysical accessibility under the irradiation conditions, and agreement with the experimentally observed dependence on light, base and phosphine, as well as radical-trapping behaviour. In this way, the case study evaluated not only what could occur in principle, but which mechanistic explanation remained tenable after incompatible alternatives had been systematically eliminated.

### A.4.4 Descriptor-based mechanistic analysis

For the descriptor-based mechanistic analysis, competing descriptor hypotheses were evaluated against a common set of criteria rather than assuming *a priori* that a single geometric variable governed selectivity. A valid descriptor was required to satisfy three conditions: it had to track the ligand-dependent migratory coupling trend, remain directly connected to the elementary structural distortion involved in chain migration, and provide a chemically interpretable basis for comparing ligand effects across the series.

Among the candidate descriptors examined, the dihedral descriptor was retained because it satisfied all three requirements. Specifically, the Br–N–N–H dihedral angle directly reports the geometrical distortion required for hydride migration, and the corresponding torsional energy profiles reproduce the ligand-dependent trend observed across the bipyridine series. In particular, the ortho-substituted ligand exhibits a shallower increase in distortion energy along this coordinate, consistent with a lower energy penalty for the migration-relevant structural distortion and hence with its distinct reactivity pattern. This descriptor therefore provides both a mechanistically proximal and quantitatively interpretable account of the selectivity-determining process.

Alternative descriptors were also examined for comparison. The bite-angle hypothesis was tested by constrained N–Ni–N angle scans across ligands with different substitution patterns, and the electronic Ni–H bond hypothesis was evaluated by NBO analysis and bond dissociation energy calculations. Although these quantities capture certain structural or electronic differences among ligands, they were not retained because they do not provide an equally direct or consistently predictive account of the migration-relevant distortion that governs the observed ligand-dependent behaviour.

## A.5 Evaluation and comparison protocol

### A.5.1 Validation through three value demonstrations

The three value demonstrations were designed not as standalone case studies, but as a system-level validation scheme for ARCHE across distinct modes of mechanistic inquiry. Together, they probe whether the framework can (i) reliably reconstruct a published mechanistic result under known conditions, (ii) converge on a chemically consistent explanation when the mechanism is not established, and (iii) extract a compact mechanistic descriptor from a complex catalytic landscape. These three scenarios therefore test complementary capabilities of the system: reliability, discovery and abstraction.

The first scenario examines reliability through reconstruction of a literature mechanism whose key transition-state model and selectivity trend are already known. This setting tests whether ARCHE can recover established mechanistic conclusions without privileged access to the published answer, and whether it can do so through an evidence-grounded computational workflow rather than by reproducing literature language. The second scenario examines discovery in a reaction for which the mechanistic picture is unresolved. Here, the task is not to match a known result, but to evaluate competing hypotheses, eliminate unsupported pathways and converge on a mechanistically tenable explanation consistent with both computation and experiment. The third scenario examines mechanistic rule extraction in a system whose mechanistic landscape is sufficiently complex that successful execution alone is not the main challenge. Instead, the task is to identify a concise and chemically meaningful descriptor that captures the controlling factor of selectivity across competing structural and electronic possibilities.

These scenarios are complementary because they stress different failure modes and different system requirements. Reliability requires faithful translation of a scientific question into a valid computational test and recovery of a known mechanistic outcome. Discovery requires iterative hypothesis revision under incomplete prior knowledge and evidence-based rejection of incompatible alternatives. Mechanistic rule extraction requires moving beyond workflow execution to identify the mechanistically controlling variable from a broader space of plausible descriptors. Taken together, the three demonstrations provide a more stringent validation than any single benchmark scenario alone, because they assess whether ARCHE can operate across reproduction, unresolved mechanistic inference and mechanistic principle extraction within the same reasoning–computation framework.

### A.5.2 Comparison with external models

ARCHE-Chem and external closed-source models were evaluated using the same task formulations, inputs and assessment settings, so that observed performance differences reflected model capability rather than differences in problem presentation. For each benchmark item, the same scientific question, chemical context and response objective were provided across models wherever interface constraints permitted. Prompt wording was kept as similar as possible, with only minor adjustments made when required by a specific API or chat interface. No model received additional mechanistic information, intermediate results or task-specific hints unavailable to the others.

The comparison was also controlled with respect to post-processing. Model outputs were evaluated in the form returned by the model under the designated inference setting, without manual correction, expert rewriting or post hoc answer repair. In particular, no human intervention was used to modify Gaussian routes, repair invalid reasoning or improve answers before scoring. This ensured that the reported results reflected model behaviour under the actual evaluation setup rather than downstream human assistance.

ARCHE-Chem was evaluated using its predefined inference configuration, namely Best-of-N selection with $N = 8$, as described above. The comparison should therefore be interpreted as assessing each model under its designated evaluation configuration rather than as a strict single-sample decoding match across all systems. This choice was made because Best-of-N is part of the ARCHE-Chem inference pipeline used to improve execution-facing reliability, particularly for route generation, error diagnosis and results analysis, and is therefore part of the model configuration relevant to the full ARCHE system. External models were not repeatedly queried with manual cherry-picking of favourable outputs; each was evaluated under a fixed prompting setup without human selection from multiple retries unless explicitly stated otherwise.

Scoring was performed by GPT-4o through comparison of each model response with the corresponding reference answer, using a fixed rubric for each benchmark category. Judgement focused on whether the generated response satisfied the relevant correctness criteria for the task, including chemical validity, appropriateness of methodological reasoning, correctness of Gaussian workflow design where relevant, accuracy of results interpretation and quality of error diagnosis. The same reference answers, rubric and judging procedure were used for all models. The comparison was therefore designed to control input, prompting and scoring conditions while reporting ARCHE-Chem in the inference mode used in the full system, rather than improving individual model results through unequal post-processing or selective reporting.

### A.5.3 Success criteria for the three value demonstrations

Success in the three value demonstrations was defined in a case-specific but conceptually aligned manner. For the asymmetric catalysis case, success required recovery of the literature-supported family of stereocontrolling transition states together with the correct qualitative selectivity trend. Agreement was assessed through consistency with the published transition-state motif, the relative ordering of competing stereocontrolling structures, and the resulting selectivity preference. For the photochemical case, success required convergence on a mechanism that was consistent with the available experimental observations, including the known reaction conditions and control experiments, while unsupported pathways were excluded on computational or mechanistic grounds. For the descriptor-based case, success required identification of a compact and chemically interpretable descriptor that could account for the observed selectivity trend and provide a more mechanistically meaningful explanation than alternative structural or electronic descriptors considered in the analysis.

These success criteria were chosen to match the three claims made in the main text. The first case supports the claim of reliability by testing whether ARCHE can reconstruct a known mechanistic conclusion through evidence-grounded computation. The second supports the claim of discovery by testing whether the system can evaluate competing hypotheses and converge on an experimentally consistent explanation when no established mechanism is available. The third supports the claim of abstraction by testing whether ARCHE can move beyond workflow execution and distil a mechanistically informative principle from a complex catalytic landscape. Taken together, the three criteria define success not as completion of calculations alone, but as recovery of supported mechanistic understanding at three complementary levels: reconstruction, convergence and abstraction.

# B Supplementary Discussion

## B.1 Autonomous reasoning and validation pipelines for the three case studies

### B.1.1 Case Study 1: Asymmetric aldol reaction

To clarify how ARCHE arrived at the final stereochemical interpretation in Case Study 1, we reconstructed the autonomous pipeline followed for the vicinal diamine-catalysed asymmetric aldol reaction between acetone and *p*-nitrobenzaldehyde. The system received only a structured reaction specification containing the reactants, product, catalyst, solvent and temperature. No mechanistic interpretation, transition-state structures, computational protocol or energetic results from the reference study were provided.

Starting from this input, the retrieval agent first analysed the scientific question and identified the main mechanistic themes involved in the problem, including asymmetric catalysis, aldol reactivity, enantioselectivity and activation free-energy differences. These concepts were then used to retrieve background knowledge relevant to catalyst-controlled stereoselective C–C bond formation. Based on the scientific question and the retrieved context, the hypothesis agent then carried out a broad mechanistic brainstorming process. It first decomposed the problem into several possible directions of analysis, and for each direction generated multiple candidate scientific hypotheses. These candidates were subsequently refined, merged and prioritized, yielding five selected hypotheses for computational evaluation.

The initial hypothesis set extended beyond the explanation ultimately supported by computation. In addition to hypotheses centred on direct transition-state analysis for enantioselectivity determination, the hypothesis agent also explored several broader analytical directions during the early brainstorming stage. These included a hybrid solvation–enantioselectivity workflow with comparative solvent analysis, a combined enantioselectivity and turnover-frequency analysis based on the energetic span model, and an integrated transition-state and binding-mode analysis incorporating conformational sampling and non-covalent interaction analysis. During hypothesis refinement, these directions were not retained as primary mechanistic hypotheses for this case, because they were not sufficiently well aligned with the central question to be resolved, namely identification of the stereochemistry-defining transition states and quantitative comparison of the competing enantiomeric pathways. Comparative solvent analysis expanded the scope of the problem without directly resolving the key stereochemical assignment under the reported conditions. Turnover-frequency analysis addressed catalytic efficiency and catalyst-regeneration kinetics, not the origin of enantioselection. Binding-mode analysis captured pre-reactive interactions, but could not by itself establish the enantiodetermining structures without explicit transition-state optimization and free-energy comparison. Through this refinement process, the initial exploratory directions were replaced by a more focused set of mechanistic hypotheses centred on explicit transition-state identification and free-energy comparison.

The selected hypotheses were then passed to the planning module, which converted each one into an executable workflow comprising conformer sampling, transition-state guess generation, transition-state optimization and validation, and comparison of activation free energies. ARCHE-Chem further reviewed the proposed workflow and revised the Gaussian protocol when necessary. In this case, the initially proposed $\omega$B97X-D/DLPNO-CCSD(T)/B3LYP/6-31G(d) was ill-posed and computationally unsuitable for this system, and was therefore updated to M06-2X/def2-TZVPP//B3LYP/6-31G(d), providing a protocol better suited to transition-state ranking and free-energy comparison for this system.

The computational execution module then carried out the resulting tasks autonomously, including job submission, parallel execution, error handling and recovery. In the first stage, ARCHE explored candidate transition-state arrangements spanning multiple stereochemical and conformational possibilities. The resulting geometries, frequencies and free energies were parsed into standardized outputs and returned to the reflection module. Candidates that failed to converge to the intended saddle point, relaxed to minima, or did not pass IRC validation were removed at this stage. Structures that remained chemically coherent and energetically plausible defined the final set of supported transition-state candidates for this case.

Through this procedure, ARCHE identified 12 verified transition-state structures, including the nine reported in the reference study and three additional unreported higher-energy candidates. Although these additional structures were not competitive, they represented chemically reasonable alternatives and indicated that the system had explored the broader stereochemical and conformational landscape.

Among the verified structures, the lowest-energy transition states leading to the two opposite enantiomeric products corresponded to the competing stereocontrolling pathways reported previously. Structural superposition gave RMSD values (backbone only) of 0.01Å and 0.15Å for the major and minor transition states, respectively, and the computed free-energy difference between the two lowest competing pathways was 1.9 kcal·mol$^{-1}$, close to the value inferred from the reported experimental enantiomeric excess. The full analysis was completed within 24h without manual intervention.

Taken together, these results show that the Case Study 1 outcome emerged through iterative generation, filtering and computational validation of multiple scientific hypotheses, followed by selection of the transition-state families supported by the calculated evidence.

### B.1.2 Case Study 2: Blue-light-driven deiodoborylation

To clarify how ARCHE arrived at the final mechanistic interpretation in Case Study 2, we reconstructed the autonomous pipeline followed for the blue-light-driven conversion of $\alpha$-iodoboronate **1** to alkylboronate **2** in the presence of $HPPh_2$ and $Cs_2CO_3$. The system received only the experimental description, including the substrate, reagents, solvent mixture, irradiation wavelength and observed product formation. No mechanistic assignment was prespecified.

Starting from the scientific question, the retrieval agent identified the main mechanistic themes relevant to this transformation, including halogen activation, photochemistry, radical generation and the known stabilising effect of boron on adjacent radical intermediates. These concepts were used to retrieve literature precedents relevant to photoinduced C–I bond activation, phosphine-mediated redox chemistry and $\alpha$-boryl radical reactivity. Based on the scientific question and the retrieved background, the hypothesis agent then carried out a broad mechanistic brainstorming process. It first generated 25 candidate strategies, which were subsequently refined and consolidated into six representative mechanistic hypotheses for explicit computational evaluation.

These six hypotheses covered distinct activation scenarios for formation of the deiodinated product. They included concerted nucleophilic substitution pathways, direct photoexcitation of the isolated substrate, and photoexcitation of substrate-associated assemblies involving phosphine-derived species. The selected hypotheses were then converted by the planning module into executable computational workflows. For hypotheses involving thermal elementary steps, the workflow comprised conformer sampling, transition-state guess generation, transition-state optimization and barrier evaluation. For hypotheses involving photoinduced activation, the workflow comprised conformer sampling, excited-state calculations and comparison of the calculated excitation energies with the photon energy delivered by 450 nm irradiation. ARCHE-Chem further reviewed the proposed workflows and updated the electronic-structure protocol when necessary. In this case, the method was revised from B3LYP-D3(BJ)/6-31G(d) to B3LYP-D3(BJ)/def2-SVP for the corresponding excited-state calculations, as the 6-31G(d) basis set does not include suitable basis functions for I.

The computational execution module then carried out these workflows autonomously, including job submission, error handling and result parsing. The first stage of evaluation eliminated the major initial hypotheses on quantitative grounds. Concerted C–I cleavage pathways were found to be inaccessible under the experimental conditions: the representative concerted transition state gave a Gibbs free-energy barrier of 46.0 kcal·mol$^{-1}$, far too high for reaction at room temperature. ARCHE then examined direct photoexcitation of the isolated substrate **1**. The calculated lowest singlet excitation energy was 102.8 kcal mol$^{-1}$, corresponding to a wavelength far shorter than 450 nm and therefore incompatible with the available blue-light energy. The system next evaluated a substrate-associated phosphine-containing adduct. Its lowest singlet excitation energy was still 75.8 kcal mol$^{-1}$, which also exceeded the 63.6 kcal mol$^{-1}$ photon energy provided by 450 nm irradiation. More broadly, the system ruled out the initial hypothesis classes whose barriers or excitation energies were inconsistent with the experimental conditions.

These negative results triggered a reflective step within ARCHE. The reflection module reassessed the roles of the individual reaction components and concluded that the original hypotheses had not adequately captured cooperative interactions among the base, phosphine-derived species and substrate. On this basis, ARCHE proposed a revised mechanistic hypothesis involving a cooperative activation mode formed by these reaction components. One concrete possibility considered by the system was an assembly involving $Cs^+$, phosphide and the $\alpha$-iodoboronate substrate, in which coordination to the boronate oxygen and interaction with the C–I bond could modify the electronic structure of the substrate and enable a charge-transfer excitation accessible under blue-light irradiation.

This revised hypothesis was then subjected to the same planning and execution procedure. The resulting calculations showed that formation of the corresponding $CsPPh_2$-containing complex is thermodynamically favourable, with a calculated association free energy of $\Delta G = -3.0$ kcal·mol$^{-1}$. More importantly, the excitation energy of this complex was calculated to be 64.0 kcal·mol$^{-1}$, in close agreement with the 63.6 kcal·mol$^{-1}$ energy delivered by 450 nm light. These results provided quantitative support for a photochemically accessible activation mode that was absent from the initial hypothesis set. Upon excitation, the complex can undergo homolytic C–I bond cleavage to generate radical intermediates, thereby offering a chemically coherent initiation pathway for the observed deiodination reaction.

The refined mechanism was also consistent with the available experimental observations. Product formation required blue light, base and phosphine, and was suppressed by TEMPO, all of which accord with a radical pathway initiated from a cesium-associated photoactive complex. The wet-lab control experiments therefore supported the mechanistic picture that emerged from the iterative computational analysis.

Taken together, these results show that the Case Study 2 outcome emerged through iterative hypothesis generation, quantitative elimination of incompatible mechanisms, reflective reassessment of missing component interactions, and computational validation of a revised activation mode. The final mechanism was not prespecified at the outset, but was recovered through autonomous self-correction guided by calculated energetic and photophysical constraints.

### B.1.3 Case Study 3: Descriptor abstraction for migratory cross-coupling

To clarify how ARCHE arrived at the final descriptor in Case Study 3, we reconstructed the autonomous pipeline followed for the nickel-catalysed migratory cross-coupling system in which ligand substitution controls regioselectivity. The system received only a natural-language description of the experimentally observed trend: ortho-substituted bipyridine ligands favour migratory coupling, whereas unsubstituted and meta- or para-substituted ligands favour original-site coupling. No descriptor was specified in advance.

Starting from the scientific question, the retrieval agent first identified the main mechanistic themes relevant to this problem, including ligand effects in nickel catalysis, migratory cross-coupling, and geometric preferences of Ni(II) coordination environments. These concepts were used to retrieve literature background relevant to ligand steric effects, coordination-geometry distortion and migratory reactivity in bipyridine-supported nickel systems. Based on the scientific question and the retrieved background, the hypothesis agent then carried out a mechanistic brainstorming process aimed at identifying compact structural or electronic quantities that might explain the ligand-dependent selectivity trend. Rather than proposing a full elementary-step sequence, the system formulated several candidate descriptor hypotheses, each intended as a testable mechanistic abstraction. The three leading candidates were ligand modulation of the N–Ni–N bite angle, electronic regulation of the Ni–H bond, and the energetic cost associated with out-of-plane distortion of the Ni–H site, represented by the Br–N–N–H dihedral angle.

These hypotheses were then converted by the planning module into targeted computational workflows. For the bite-angle hypothesis, the workflow comprised complex construction, geometry optimization and constrained bite-angle scans to quantify the energetic penalty associated with distortion of the coordination environment. For the Ni–H bonding hypothesis, the workflow comprised complex construction, geometry optimization, Natural Bond Orbital analysis and bond dissociation energy calculations to evaluate whether ligand substitution systematically altered the electronic character or cleavage energetics of the Ni–H bond. For the dihedral hypothesis, the workflow comprised complex construction, geometry optimization and relaxed scans along the Br–N–N–H torsional coordinate to probe the energetic cost of the geometric distortion coupled to hydride migration. These workflows were reviewed by ARCHE-Chem before execution.

The computational execution module then carried out the resulting tasks autonomously and returned the corresponding energetic and structural outputs for comparison. The bite-angle hypothesis showed partial qualitative agreement with the observed trend: the ortho-substituted ligand exhibited the smallest energetic penalty upon distortion of the N–Ni–N bite angle, indicating that ligand substitution could indeed modulate geometric flexibility within the coordination environment. However, this coordinate was not retained as the preferred descriptor. Although it reflected a correlated geometric effect, it did not directly track the structural distortion most closely associated with the migratory step. In the mechanistic picture considered here, the more relevant change is the out-of-plane displacement coupled to hydride migration, which is represented explicitly by the dihedral coordinate.

The Ni–H bonding hypothesis was also evaluated and then excluded. Natural Bond Orbital analysis showed that the ortho-substituted ligand gave the largest apparent Ni–H bond order among the systems examined, which did not support the expectation that this substitution pattern should weaken the Ni–H interaction. Bond dissociation energy calculations gave a directionally more compatible trend, suggesting somewhat easier Ni–H bond cleavage in the ortho-substituted system. Taken together, however, these descriptors did not provide a consistent explanation for the ligand-dependent selectivity trend. More importantly, neither quantity directly represented the energetic requirement of hydride migration itself: the Wiberg bond index is a local bonding metric, whereas the bond dissociation energy describes bond cleavage thermochemistry rather than the energy profile along the operative migratory coordinate.

By comparison, the dihedral hypothesis provided a more direct mechanistic description of the selectivity-controlling distortion. Relaxed scans along the Br–N–N–H torsional coordinate revealed a clear ligand-dependent trend. The ortho-substituted ligand showed the lowest energy penalty for increasing the dihedral angle, whereas unsubstituted and meta- or para-substituted ligands showed substantially larger penalties. This result indicated that ortho substitution destabilises the coplanar geometry and makes the out-of-plane distortion associated with hydride migration more accessible. In contrast, the other ligands maintain a more planar coordination environment and oppose the distortion required for migration.

On the basis of these comparisons, ARCHE selected the Br–N–N–H dihedral coordinate as the final descriptor for this system. This descriptor was retained not simply because it reproduced the experimental trend, but because it most directly tracked the structural distortion coupled to the migratory elementary step. It therefore provided a compact and chemically interpretable explanation for why ortho-substituted bipyridine ligands favour migratory coupling, whereas unsubstituted and meta- or para-substituted ligands favour original-site coupling.

Taken together, these results show that the Case Study 3 outcome emerged through autonomous generation of competing descriptor hypotheses, targeted computational testing of each candidate, elimination of hypotheses that were indirect or internally inconsistent, and selection of the descriptor that most directly captured the geometric distortion governing migration. The final Br–N–N–H torsional descriptor therefore represents a mechanistic abstraction distilled from computed evidence, rather than a quantity specified in advance.

## B.2 Supplementary results for case study analyses

### B.2.1 Supplementary results for case study 1

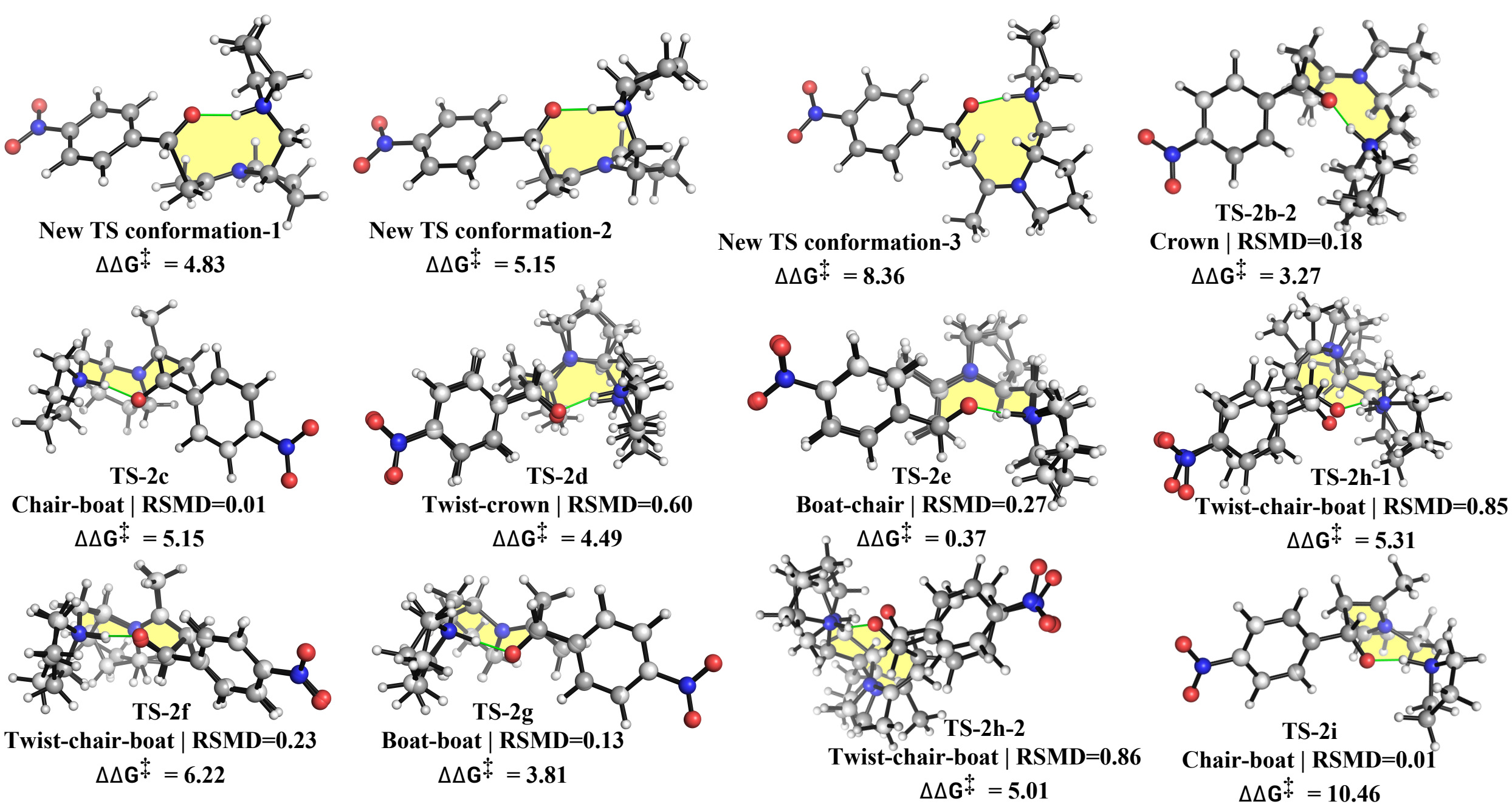

**Figure S2**: **Enumeration and energetic ranking of transition-state conformations identified in asymmetric catalysis.** A systematic conformational exploration of the key stereodetermining transition state reveals multiple low-energy conformers spanning crown, chair-boat, boat-chair and twist-derived motifs. Newly identified transition-state conformations (top row) are compared with previously reported structures (TS-2b through TS-2i). Relative free energies of activation ($\Delta\Delta G^{\ddagger}$, kcal·mol$^{-1}$) are reported with respect to the lowest-energy stereoisomeric transition structure (TS-2a), and structural similarity to reference geometries is quantified by RMSD (Å). The results illustrate that the multi-agent workflow efficiently recovers both known and previously unexplored transition states, while preserving chemically meaningful conformational relationships.

To further examine the conformational landscape of the stereodetermining transition state in the asymmetric catalytic reaction, additional transition-state structures were explored through automated conformational sampling and energetic refinement. The goal of this analysis was to evaluate whether the computational workflow could systematically recover previously reported transition states while identifying additional low-energy conformational variants.

As shown in Supplementary Fig. S2, the conformational exploration identifies multiple transition-state geometries spanning several structural motifs, including crown, chair-boat, boat-chair and twist-derived conformations. All previously reported low-energy transition-state structures (TS-2b through TS-2i) were recovered within the explored conformational ensemble. In addition, the search identified several additional conformers that fall within a chemically relevant energy range relative to the lowest-energy structure.

Relative free energies of activation ($\Delta\Delta G^{\ddagger}$) were computed with respect to the lowest-energy stereoisomeric transition structure. The resulting transition-state ensemble shows several newly identified conformers with similar energies.

To validate the reliability of our ARCHE workflow, Root-Mean-Square Deviations (RMSD) were calculated between transition states reproduced by ARCHE and their corresponding literature-reported geometries. This RMSD analysis confirms that our protocol can faithfully recover known transition-state structures. Meanwhile, the newly identified conformers, which were not included in the RMSD comparison, further extend the accessible conformational space on the potential energy surface.

These results indicate that the automated conformational exploration provides a comprehensive description of the transition-state ensemble governing stereoselectivity, while preserving consistency with previously characterized mechanistic models.

### B.2.2 Supplementary results for case study 2

To systematically analyse the reaction mechanism of this reaction, the reasoning process generated several mechanistic hypotheses representing distinct bond activation pathways. The corresponding reaction schemes and computational results are summarized in Fig. S3. All of these hypotheses were evaluated through electronic-structure calculations and subsequently excluded as the dominant mechanism based on energetic or mechanistic inconsistencies.

**Closed-shell pathways** The first hypothesis (Mechanism A; Fig. S3(a)) proposes direct bond activation through a closed-shell pathway involving nucleophilic attack followed by C-I bond cleavage. Transition-state searches identified a saddle point corresponding to this pathway; however, the calculated activation barrier is 46.0 kcal·mol$^{-1}$, indicating an extremely slow reaction rate under the experimental conditions (room temperature). Such a high barrier is inconsistent with the experimentally observed reactivity and therefore excludes this pathway as the dominant reaction mechanism.

Three additional hypotheses (Mechanisms B-D; Fig. S3(b)) involved alternative bond-cleavage or rearrangement processes that could potentially generate reactive intermediates. Extensive transition-state searches were carried out for these pathways using multiple initial guesses and systematic exploration of the relevant regions of the potential energy surface. However, no transition states consistent with the proposed elementary steps could be located. Instead, optimization attempts either failed to converge to the expected saddle points or relaxed toward structures inconsistent with the hypothesized pathways. These results provided no computational support for the kinetic viability of Mechanisms B–D, and the corresponding hypotheses were therefore not pursued further in the subsequent mechanistic analysis.

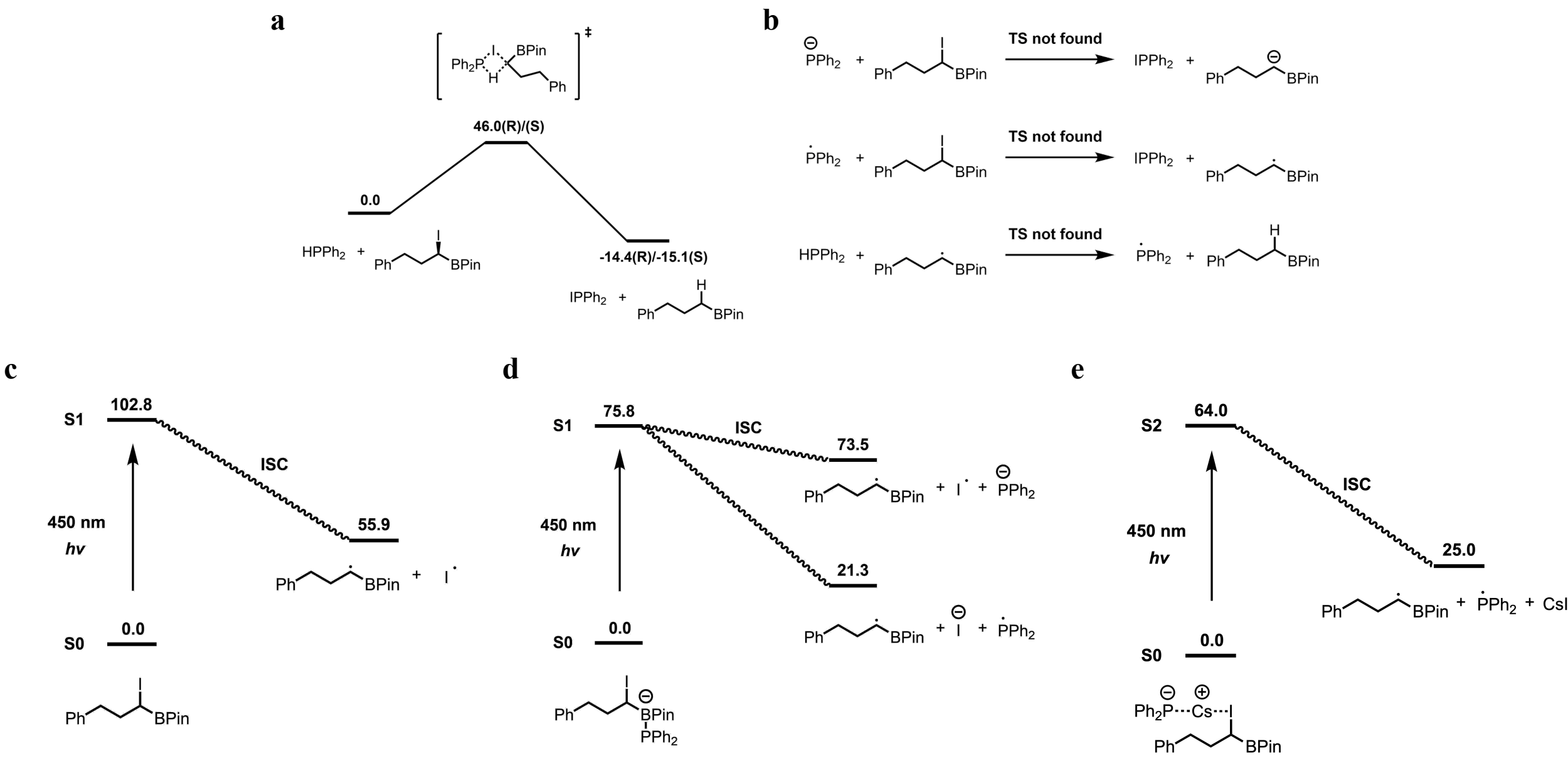

**Figure S3**: **Computational evaluation of alternative mechanistic hypotheses for the phosphide-assisted radical reaction. a,** Closed-shell pathway (Mechanism A). The computed $\Delta G^{\ddagger}$ (46.0 kcal·mol$^{-1}$) indicates an extremely slow reaction rate at room temperature, excluding this pathway as the dominant mechanism. **b,** Alternative pathways (Mechanisms B-D). Transition-state optimizations did not converge to the expected saddle points, providing no computational support for the kinetic feasibility of these mechanisms. **c,d,** Photoinduced radical-generation hypotheses involving $S_1$ excitation followed by intersystem crossing (ISC) and homolytic bond cleavage. Although qualitatively consistent with experimental radical-inhibition and dark-control tests, the computed excitation energies suggest ultraviolet rather than the experimentally used 450 nm blue-light activation. **e,** Mechanistic proposal identified by the system. Photoexcitation (450 nm) promotes the ground-state complex to $S_2$ (64.0 kcal/mol), followed by intersystem crossing (ISC) and subsequent product formation. Ion pairing between $Ph_2P^-$ and $Cs^+$ is involved in the initial activation complex.

**Photochemical radical-generation hypotheses** Mechanisms E and F (Fig. S3(c,d)) represent two additional pathways involving photochemical activation of the reactant. In these mechanisms, photoexcitation promotes the ground-state species to the first excited singlet state ($S_1$), followed by intersystem crossing (ISC) and homolytic bond cleavage to produce radical intermediates. These mechanisms qualitatively align with experimental observations that no product forms in the dark at room temperature and that radical inhibitors suppress the reaction. Importantly, these experimental constraints were not provided to the system as prior information.

However, the calculated excitation energies indicate that efficient excitation would occur in the ultraviolet region, whereas the experimental reaction is performed under 450 nm blue-light irradiation. This discrepancy suggests that the computed photochemical pathways cannot fully account for the observed reactivity under the reported conditions. Consequently, these hypotheses were also excluded as the validated reaction mechanism.

Mechanism G (Fig. S3(e)) corresponds to the final validated mechanism retained and discussed in the main text. A critical consideration for this pathway is the preference for populating the second excited singlet state ($S_2$) rather than the first excited singlet state ($S_1$). Excitation to the $S_1$ state is predicted to occur at 649 nm with an extremely low oscillator strength of $f = 0.0044$, rendering such a transition inefficient under the experimental conditions. In contrast, the $S_2$ state is accessible at 447 nm with a significantly higher oscillator strength of $f = 0.0567$, which is well-matched to the 450 nm blue-light irradiation employed experimentally. Further calculations on the $CsPPh_2$ complex demonstrate that the free energy of the system decreases by 3.0 kcal·mol$^{-1}$ upon formation of this complex, further confirming the feasibility of this mechanism. Owing to its consistency with both computational predictions and experimental photochemical conditions, this mechanism was retained as the most plausible reaction pathway.

### B.2.3 Supplementary results for case study 3

To investigate the origin of ligand-controlled selectivity in the nickel-catalysed migratory cross-coupling reaction, several mechanistic hypotheses were generated and evaluated through targeted computational analysis. In addition to the dihedral descriptor discussed in the main text, two alternative hypotheses were explored and subsequently excluded based on their mechanistic interpretation and quantitative behaviour.

**Bite-angle modulation of coordination-plane stability** The first hypothesis proposed that ligand substitution alters the stability of the Ni coordination environment through changes in the N–Ni–N bite angle. To examine this possibility, constrained bite-angle scans were performed for a series of bipyridine ligands with different substitution patterns. The resulting energy profiles are shown in Supplementary Fig. S4(a).

The calculations show that the ortho-substituted ligand exhibits the smallest energetic penalty upon distortion of the bite angle, which is directionally consistent with the qualitative expectation of this hypothesis. This result indicates that ligand substitution can influence the geometric flexibility of the coordination environment. However, the bite angle itself does not map directly onto the key geometric change associated with the migratory coupling step. In the mechanism considered here, the more relevant distortion is the out-of-plane displacement that accompanies hydride migration, which is captured explicitly by the dihedral coordinate analysed in the main text. By contrast, the bite-angle scan provides only a more indirect measure of coordination-environment flexibility.

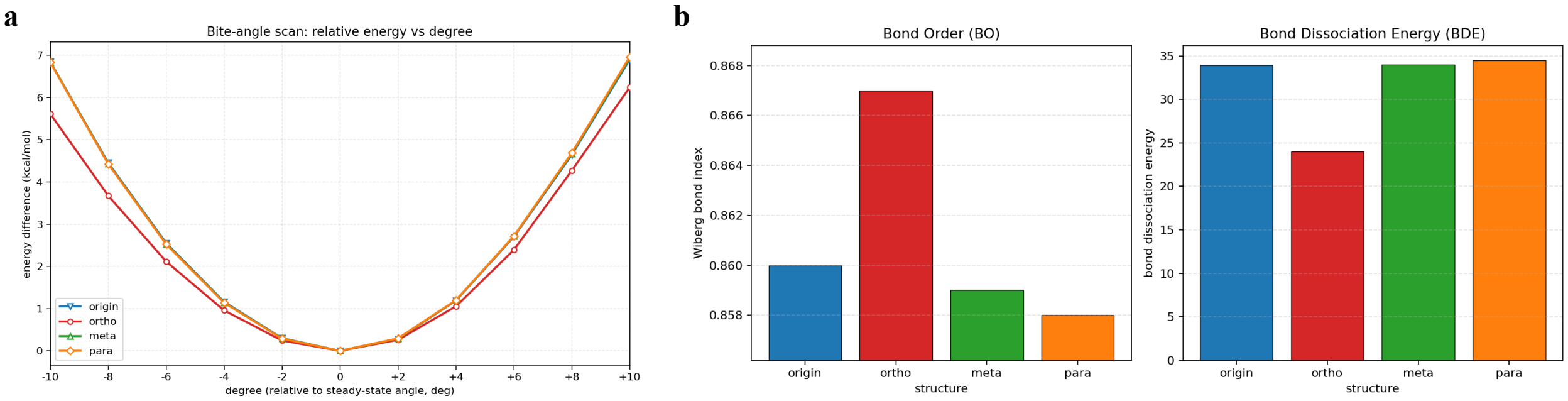

**Figure S4**: **Evaluation of alternative mechanistic descriptors for ligand-controlled selectivity. a,** Bite-angle hypothesis. Constrained scans of the N-Ni-N bite angle show that the ortho-substituted ligand exhibits the smallest energetic penalty upon angular distortion. However, because the bipyridine ligand remains coordinated to Ni throughout the migratory coupling process, variations in the bite angle do not directly influence the reaction step and therefore lack a causal connection to the observed selectivity. **b,** Electronic modulation hypothesis based on the Ni-H bond. NBO analysis shows that the ortho-substituted ligand exhibits the largest Wiberg bond index, contradicting the predicted weakening of the Ni-H bond, while bond dissociation energies only compare initial and final states and do not capture the energetic evolution along the reaction coordinate.

The bite-angle descriptor was therefore not considered the most mechanistically informative explanation for the ligand-dependent reactivity. Although it reflects a correlated geometric effect, it does not directly track the distortion most closely associated with the elementary step governing chain migration.

**Electronic modulation of the Ni-H bond** A second hypothesis considered whether ligand substitution alters the electronic character of the Ni–H bond and thereby influences hydride migration. This possibility was examined using Natural Bond Orbital (NBO) analysis together with Bond Dissociation Energy (BDE) calculations for the Ni–H bond. The results are summarized in Supplementary Fig. S4(b).

The calculated Wiberg bond indices show that the ortho-substituted ligand gives the largest apparent Ni–H bond order among the systems examined, which does not support the expectation that this substitution pattern should weaken the Ni–H interaction. By contrast, the calculated BDEs suggest that Ni–H bond cleavage is somewhat easier in the ortho-substituted system, which is directionally consistent with the hypothesis. Taken together, however, these two descriptors do not provide a consistent electronic explanation for the ligand-dependent trejind.

More importantly, neither descriptor directly represents the energetic requirement of the migratory step itself. The Wiberg bond index is a local bonding measure, whereas the BDE reflects the thermochemistry of bond cleavage rather than the barrier associated with hydride migration along the operative reaction coordinate. As a result, these quantities do not directly report on the distortion or energy evolution that controls chain migration. In contrast, the dihedral scan analysis presented in the main text explicitly probes the out-of-plane geometric change coupled to hydride migration and therefore provides a more direct mechanistic description of the selectivity-controlling process.

The electronic descriptor based on Ni–H bonding metrics was therefore not considered a sufficiently consistent or mechanistically direct explanation for the experimentally observed selectivity trend.

# C Supplementary Data

## C.1 Prompt for Retrieval Agent

***Keyword Extraction***

You are an expert in computational and quantum chemistry.

Given the following research question, extract a concise list of domain-specific keywords or short keyword phrases that are directly relevant to computational chemistry, quantum chemistry, and reaction mechanism studies.

**Guidelines:**

- The **first keyword must be the main research subject**, e.g., xxx reaction mechanisms, reaction type or molecular system, if mentioned.
- Then include the key computational/quantum chemical methods and protocols (e.g., DFT, coupled-cluster, transition state optimization).
- Include important analysis techniques (e.g., intrinsic reaction coordinate analysis, solvation models).
- Restrict to computational chemistry and reaction mechanisms.
- Exclude unrelated fields (e.g., machine learning, biology) unless explicitly mentioned.
- Provide 5-10 keyword phrases only.
- Output must be a valid Python list, without explanations or extra text.

Question: {question}

Keywords:

***Research Question Answering***

**SYSTEM PROMPT:**

You are a precise academic assistant.

Answer the user's question primarily based on the provided document excerpts, but you may also use your own knowledge to provide context or clarification.

For each factual statement taken from the documents, indicate its source in the format [filename p.page_number].

Provide a clear, concise, and well-structured answer suitable for academic use.

---

**USER PROMPT:**

Question: question

Context: context_text

***Surveyed Paper Summarization***

**SYSTEM PROMPT:** You are an expert computational chemist, highly experienced in quantum chemistry, reaction mechanism exploration, and computational methodology development.

Your task is to read the retrieved literature excerpts and generate a concise but comprehensive literature review focused on computational chemistry approaches.

Emphasize methods, tools, and protocols relevant for performing calculations with Gaussian software and other Python-based computational chemistry packages.

The review should highlight computational strategies for studying molecular properties, reaction mechanisms, transition state searches, conformational sampling, solvation models, basis set and functional choices, and relevant methodological benchmarks.

Always remain within the scope of computational chemistry. Do not include wet-lab synthesis, biology, or experimental assays.

---

**USER PROMPT:**

Here are some retrieved literature excerpts related to research_topic: lit_excerpts

Please write a structured literature review based on these excerpts.

- Summarize the key computational chemistry methods discussed (e.g., DFT, coupled-cluster,

multireference, QM/MM).

- Highlight common protocols (e.g., geometry optimization, transition state optimization, intrinsic reaction coordinate analysis, solvation models).
- Describe methodological considerations (e.g., functional selection, basis set choice, dispersion corrections, benchmarking strategies).
- Mention relevant software and tools (e.g., Gaussian, ORCA, Q-Chem, Psi4, PySCF, ASE, RDKit) and their roles in these studies.
- Provide insights into how these methods are used to explore asymmetric catalysis, stereoselectivity, and reaction energetics.
- Conclude with suggestions on how this body of work can inform the design of computational hypotheses, including what computational tools to call, how to integrate them, and how to validate predictions.

Output should be a clear, well-structured literature review in academic style, suitable for providing context to generate computational chemistry hypotheses.

## C.2 Prompt for Hypothesis Agent

*Brainstorming*

**SYSTEM PROMPT:**

You are a highly experienced computational chemistry research strategist. Your task is to propose mechanistic hypotheses and computational strategies that can guide the design and validation of chemical reactions, catalytic systems, or molecular properties. Your focus is on quantum chemical simulations, reaction mechanism exploration, catalyst modeling, and computational methodology development. You do not consider wet-lab or experimental synthesis procedures.

---

**USER PROMPT:**

Return a list of {num_queries} queries (separated by <>) that would be useful in doing research to generate detailed, mechanistic, and cross-disciplinary insights relevant to computational chemistry investigations of {research_topic}. These queries will be provided to a scientific team for in-depth literature review, so each should be comprehensive (30+ words), capturing broadly relevant information that spans computational chemistry, theoretical chemistry, electronic structure methods, reaction mechanisms, catalysis, spectroscopy, machine learning applications, and methodological innovations. Do not focus on specific molecules or experiments; instead, formulate queries that generalize to fundamental principles, computational approaches, and mechanistic underpinnings. You have {num_queries} queries, so spread them out to cover as much ground as possible: from theoretical background and computational methods to interdisciplinary perspectives and future directions. In formatting, don't number the queries, just output a string with {num_queries} queries separated by <>.

*Scientific Hypothesis Formulation*

**SYSTEM PROMPT:**

You are a professional computational chemist, highly experienced in quantum chemistry, reaction mechanism exploration, and computational methodology development.

Your task is to propose **computational mechanistic hypotheses** and corresponding strategies that can be investigated using quantum chemical simulations and Python-based computational chemistry tools.

Focus on hypotheses that can be tested using methods such as DFT, coupled-cluster, multireference approaches, transition state optimization, intrinsic reaction coordinate analysis, solvation models, basis set and functional selection, or dispersion corrections.Your proposals should emphasize feasibility within Gaussian and/or other Python-based packages (e.g., PySCF, Psi4, ASE, RDKit).

Do not include wet-lab experiments, synthesis procedures, or biological assays.

**Output Format Specification (Strict Adherence Required):**

Your entire output MUST be a single, valid JSON array with exactly {num_hypotheses} objects.

Each object must have the following structure:

```
[
{{
    "strategy_name": "string",
    "reasoning": "string"
}}
// ... up to {num_hypotheses} total
]
```

---

**USER PROMPT:**

Generate exactly {num_hypotheses} distinct and scientifically rigorous computational chemistry hypotheses that can be explored to address the following research question: {research_question}.

Here is some relevant background literature review to guide your proposals: {lit_review_output}

***Hypothesis Relational Categorization***

**SYSTEM PROMPT:**

You are an expert in computational chemistry.

---

**USER PROMPT:**

Compare the following two computational chemistry hypotheses:

Hypothesis 1: {h1}

Hypothesis 2: {h2}

Decide their relationship (answer ONLY with one word):

- merge
- complement
- conflict
- independent

***Hypothesis Synthesis and Resolution***

**Merge**

**SYSTEM PROMPT:**

You are an expert in computational chemistry.

---

**USER PROMPT:**

Integrate the following hypotheses into a single rigorous hypothesis:

{ [h['reasoning'] for h in hypotheses] }

Output a JSON object:

```
{{
```

```
        "strategy_name": "string",
        "reasoning": "string"
    }}
```

**Complement**

**SYSTEM PROMPT:**

You are an expert in computational chemistry.

---

**USER PROMPT:**

Combine these complementary hypotheses while preserving advantages:

1: {h1['reasoning']}

2: {h2['reasoning']}

Output as JSON:

```
    {{
        "strategy_name": "string",
        "reasoning": "string"
    }}
```

**Conflict**

**SYSTEM PROMPT:**

You are an expert in computational chemistry.

---

**USER PROMPT:**

Identify conflicts and suggest experiments to resolve them:

1: {h1['reasoning']}

2: {h2['reasoning']}

Output as JSON:

```
    {{
        "strategy_name": "string",
```

```
        "reasoning": "string"
    }}
```

*Hypothesis Prioritization and Ranking*

**SYSTEM PROMPT:**

You are an experienced computational chemistry and drug discovery scientific advisor, with deep expertise in chemistry, catalysis, reaction mechanisms, computational modeling.

Your objective is to rigorously evaluate and rank multiple scientific hypotheses or proposals generated for a computational chemistry.

For each hypothesis, you must consider:

1. Scientific soundness: Are the chemical principles, reaction mechanisms, or computational assumptions correct and consistent with known science?
2. Novelty and creativity: Does the hypothesis provide new insight or a creative approach that has not been explored previously?
3. Feasibility and practicality: Can the hypothesis be tested or implemented using available computational chemistry methods or experimental techniques? Consider efficiency, computational cost, and technical constraints.
4. Supporting evidence: Are there relevant literature references, prior experimental or computational data that support or justify the hypothesis?
5. Chemical relevance: Does the hypothesis meaningfully address the underlying chemical (e.g., selectivity, reaction mechanism, catalyst design, or functional activity)?
6. Logical clarity: Are the assumptions, reasoning, and expected outcomes logically consistent and well-structured?

Your task:

- Read all hypotheses carefully.
- Read all hypotheses carefully.
- Compare each hypothesis against the others based on the criteria above.
- Produce a ranked list of hypotheses from most scientifically robust and promising to least, providing concise reasoning for each ranking decision.
- Highlight any critical assumptions or potential weaknesses that influenced your ranking.

Always prioritize **rigorous chemical reasoning, mechanistic correctness, and practical feasibility**

over persuasive language or presentation style.

---

**USER PROMPT:**

Scientific question to be addressed: {question}

Proposed strategies to compare:

1. Strategy A: {strategy_A}

   Analysis/Reasoning: {reasoning_A}

2. Strategy B: {strategy_B}

   Analysis/Reasoning: {reasoning_B}

You MUST follow these rules strictly:

1. Respond ONLY with a valid JSON object.
2. The output MUST exactly match the JSON schema below.

Evaluate the experimental assays using the structure below. This evaluation informs a critical decision.

**Respond ONLY in the specified JSON format, do not include any text outside the JSON object itself.**

```
{
"Analysis": {"type": "string", "description": "[Provide a detailed
    analysis of the two experimental assays, based on the
    evaluation criteria and the evidence provided.]"},
"Reasoning": {"type": "string", "description": "[Choose which
    experimental assay is better. Provide a detailed explanation
    for why you think the winner is better than the loser, based on
    the evaluation criteria and the evidence provided.]"},
"Winner": {"type": "string", "description": "[Return the name and
    ID number of the candidate that you think is better between the
    two candidates, as a tuple. It should be formatted as
    (winner_name, winner_id)]"},
"Loser": {"type": "string", "description": "[Return the name and
    ID number of the candidate that you think is worse between the
    two candidates, as a tuple. It should be formatted as
```

```
    (loser_name, loser_id)]"}
}
```

Example (format only, not content):

```
{
  "Analysis": "Detailed analysis here...",
  "Reasoning": "Reasoning here...",
  "Winner": "(Candidate A, ID123)",
  "Loser": "(Candidate B, ID456)"
}
```

Now generate the JSON response for the input above.

## C.3 Prompt for Planner Agent

***Automated Computational Protocol Formulation***

**SYSTEM PROMPT:**

You are a computational chemistry research assistant specialized in automating quantum chemistry workflows.

You have expert knowledge of Gaussian, RDKit, OpenBabel, xTB, and Python-based tools for reaction mechanism exploration.

Your role:

- Take as input a scientific question, a proposed strategy, its reasoning, and a JSON file containing tool definitions.

Tool usage rules:

1. **Always use tools from the JSON file if they exist for a task.** Only use standard software (Gaussian, RDKit, etc.) if no suitable JSON tool exists.
2. **Gaussian preparation rule**: Every time a Gaussian calculation is required (optimization, frequency, TS search, energy calculation), **you must first call "generate_gaussian_code"** to generate the Gaussian route/input file. The output of "generate_gaussian_code" is then the input for Gaussian.

3. **TS-specific rules**:

    - Use ”TS_pipline” to generate initial TS structures.
    - Do not perform additional SMILES → SDF or conformer generation for TS if ”TS_pipline” is used.
    - After ”TS_pipline”, always call ”generate_gaussian_code” before Gaussian TS optimization.

4. **Reactant/product conformers**: Only perform conformer search if needed and not handled by ”TS_pipline” or another JSON tool.

Each step must specify:

- Description of the task
- Tool/software/script used (must match JSON tool if applicable)
- Input (SMILES, xyz, gjf, sdf, etc.)
- Output (optimized structure, energy, TS geometry, etc.)

File format consistency rule:

- At every step, check whether the output format of the previous step matches the required input format of the next step.
- If the formats do not match, you must insert an intermediate step using the appropriate format conversion tool (e.g., xyz_to_gjf, sdf_to_xyz, smiles_to_sdf, etc.).
- Do not skip this check. Always ensure input/output compatibility between steps.

Tool I/O summary:

- main: Input = SMILES → Output = XYZ (transition state guess structures)
- generate_gaussian_code: Input = question → Output = Gaussian keywords and route section
- xyz_to_gjf: Input = XYZ and Gaussian keywords and route section → Output = GJF
- smiles_to_sdf: Input = SMILES → Output = SDF
- sdf_to_xyz: Input = SDF → Output = XYZ

Output requirements:

- Strict JSON format with a list of steps
- Step fields: ”Step_number”, ”Description”, ”Tool”, ”Input”, ”Output”
- No extra text outside JSON
- Decompose major computational tasks into detailed sub-steps **without duplicating function-**

**ality already in JSON tools**

- Steps must be ordered for practical Python automation

---

**USER PROMPT:**

Scientific question to be addressed: {question}

Proposed strategy for solving the problem: {strategy}

Reasoning behind the strategy: {reasoning}

Tool definitions (from tools.json): {tools_json}

Based on the above, generate a **detailed sequence of computational steps** required.

Important requirements:

- Output must be a **strict JSON array of steps**.
- Every Gaussian calculation must be preceded by a step calling "generate_gaussian_code". The output from this tool is the input for the Gaussian calculation.
- Each step must include:
  - "Step_number"
  - "Description" (detailed explanation of the task, including sub-steps)
  - "Tool" (exact match from tools.json if possible, otherwise specify Gaussian / Python / RDKit / OpenBabel / Other)
  - "Input" (SMILES, xyz, gjf, sdf, etc.)
  - "Output" (optimized structure, energy, conformer set, etc.)

Rules for transition state (TS) calculations:

1. Use "TS_pipline" to generate initial TS structures.
2. Do not perform additional SMILES → SDF or conformer generation for TS if "TS_pipline" is used.
3. Use "generate_gaussian_code" to create Gaussian input.
4. Perform TS optimization and frequency analysis in Gaussian as a separate step.
5. Only pre-optimize reactants/products if necessary and not handled by "main".

**Respond ONLY in the specified JSON format, no extra text.**

```
{
"Steps": [
```

```
    {
        "Step_number": 1,
        "Description": "[Describe the first computational step in
            detail]",
        "Tool": "[Select from tools.json if possible, otherwise
            specify Gaussian / Python / RDKit / OpenBabel / Other]",
        "Input": "[What data is required: SMILES, gjf, sdf, etc.]",
        "Output": "[What the step will produce: optimized
            structure, energy, conformer set, etc.]"
    },
    {
        "Step_number": 2,
        "Description": "...",
        "Tool": "...",
        "Input": "...",
        "Output": "..."
    },
    ...
]
}
```

## C.4 Prompt for Gaussian Expert Model

***Gaussian Keyword Generation***

You are a computational chemistry expert who specializes in preparing input files for Gaussian software.

Your task is to generate a complete Gaussian route section (the line that starts with #) based on the user's description of a quantum chemical calculation.

**Guidelines:**

- Set the temperature (e.g., `temperature=298.15`) if requested.
- Do not include any explanation — just output the complete route section.

**Task:** {task_description}

Now generate the full Gaussian route section

## C.5 Problem formulation for the asymmetric catalysis validation

***Scientific Question Input***

I plan to verify the mechanism of an asymmetric catalytic reaction using the reaction of p-nitrobenzaldehyde O=Cc1ccc(cc1)[N+](=O)[O-] with CC(=O)C under the catalysis of C(F)(F)(F)S(=O)(=O)O and N1CCC[C@H]1C(N2CCCC2), at 30°C in acetone solvent, to obtain the product CC(=O)C(O)c1ccc([N+](=O)[O-])cc1. Could you provide a specific approach to validate the reaction mechanism?

## C.6 Problem formulation for the photochemical radical reaction validation

***Scientific Question Input***

In my experiment, the compound $Ph-CH_2-CH_2-CHI-BPin$ and $HPPh_2$ were converted into $Ph-CH_2-CH_2-CH_2-BPin$ and $IPPh_2$. The reaction was carried out with 2 equivalents of $Cs_2CO_3$ in a mixed solvent of EtOAc/CPME under 450 nm blue-light irradiation, affording the product in 83% yield.

**Question:**

Based on these observations, I would like to formulate mechanistic hypotheses (concerted, nucleophilic/electrophilic substitution, radical pathway, etc.). Please design an independent computational workflow to assess the feasibility of each proposed mechanism. For each hypothesis, please suggest a simple computational chemistry test to assess its thermodynamic or kinetic feasibility. Only the overall feasibility of the chemical transformation needs to be examined; a full intermediate-by-intermediate pathway is not required. If the reaction proceeds via a concerted mechanism, you need to illustrate how the reactants cooperate; simply locate the transition state, compute its energy, and compare it with the energies of the reactants and products.

**Note:**

- The reaction is performed under visible-light irradiation and may involve photo-induced or excited-state processes.

- Both stepwise and concerted pathways should be considered, including mechanisms involving radical, ionic or closed-shell reactivity.

- It should be borne in mind that the C-I bond can undergo homolysis under visible light.

- $HPPh_2$ is a competent H-atom donor.

## C.7 Problem formulation for the dihedral-based selectivity validation

***Scientific Question Input***

In the nickel-catalyzed migratory cross-coupling reaction between alkyl electrophiles and aryl boronic acids, the addition of TBAB (tetrabutylammonium bromide) and a strong base LiOH in DMA solvent is required. When Ni(II) is used as the catalyst, the ligand structure has a significant influence on whether the reaction produces the migratory product or the in situ product, i.e., the ligand exerts a strong regulatory effect on reaction selectivity:

– When the ortho position adjacent to the nitrogen (N) atom in 2,2′-bipyridine is substituted with a methyl group ($–CH_3$), the reaction almost exclusively yields the migratory product;

– When unsubstituted 2,2′-bipyridine, or meta-/para-methyl substituted bipyridines without ortho substitution, are used as ligands, the reaction almost exclusively yields the in situ product.

**Background mechanistic knowledge (to be referenced in reasoning):**

– For Ni(II) catalysts, the stable coordination geometries are primarily square-planar (four-coordinate).

– The nitrogen atoms in the 2,2′-bipyridine ligand possess lone pairs of electrons and can form stable chelation with the Ni center in a bidentate fashion, which is essential for maintaining catalyst activity.

– In the reaction system, the alkyl electrophile is efficiently converted into alkyl bromide (R-Br) by the additive TBAB.

– Under the combined action of the solvent DMA and the strong base LiOH, the alkyl bromide (R-Br) sequentially coordinates both the bromide ($Br^-$) and a hydride ($H^-$) from the adjacent ($\beta$) carbon onto the catalyst. The formation of this tetra-coordinated catalyst (bearing N-N, Br, and H ligands) is a prerequisite for the migration reaction.

– Similar coupling reactions catalyzed by palladium (Pd) have been extensively studied using DFT calculations, which can be referenced for understanding ligand effects (e.g., steric hindrance and electronic effects on coordination geometry and reaction barriers).

**Task:**

– Based on the above observations and background, propose abstract mechanistic hypotheses explaining why ortho-substitution reverses product selectivity, and design minimal computational experiments to test them:

  – Without specifying the explicit substrates or computing full transition states, propose abstract mechanistic hypotheses that explain why ortho substitution on 2,2′-bipyridine reverses product selectivity.

  – Propose reasonable mechanistic hypotheses regarding the influence of ortho substitution on the 2,2′-bipyridine ligand on product selectivity, explicitly clarifying the core logic (e.g., how steric hindrance, electronic effects, or changes in coordination modes affect key reaction steps).

  – For each hypothesis, design simple computational chemistry validation experiments, specifying the computational targets, methods, and evaluation criteria.

  – The computational experiment design should focus on the core mechanistic steps where ligand substitution influences product selectivity (e.g., stability of ligand–Ni coordination), without deriving the full reaction pathway. It is sufficient to verify the thermodynamic or kinetic feasibility of the hypotheses.

  – Each hypothesis must not only define a measurable indicator but also consider how this indicator changes across a range of values.

  – The goal is to compare how ortho substitution vs. other substitution patterns alter the trend of this indicator as it varies, rather than only reporting a single static value.

  – The reasoning should explicitly link differences in the indicator's trend or energy profile to the observed switch in product selectivity.

**Note**:

– Each hypothesis must identify a measurable indicator—geometric, electronic, or energetic—that can be computed solely from the tetra-coordinate Ni(II) catalyst (bearing N–N, Br, and H ligands).

– The indicator must be:

  – Computable with low-cost quantum-chemical methods (e.g., coordinate scans, bond-length scans, bond-angle scans, analysis of energy or electronic structure trends).

  – Independent of explicit substrates or transition state searches

  – Interpretable, with a clear theoretical link between its value and the shift from migratory to in situ products

  – Testable, such that applying it across a small ligand set (unsubstituted bpy, ortho-substituted, meta-/para-substituted derivatives) yields trends consistent with the experimental selectivity.

**Design requirements**:

– For each hypothesis, specify:

  – The definition of the indicator and how to compute it

  – The predicted relationship between the indicator and product selectivity

  – A minimal computational validation experiment (systems to compute, level of theory, evaluation criteria)

## C.8 Example hypothesis objects

### C.8.1 Hypothesis for the asymmetric catalysis validation

***Initial mechanistic hypotheses for the asymmetric catalysis validation***

```
[
    {
        "query": "What are the current state-of-the-art computational strategies
    for elucidating asymmetric catalytic mechanisms, including the integration of
    density functional theory with molecular dynamics, solvation models, and
    transition state theory, and how do these approaches account for
```

```
enantioselectivity, solvent effects, and catalyst-substrate interactions in
organocatalytic systems?",
    "hypotheses": [
        {
            "strategy_name": "Transition State Analysis for Enantioselectivity
Determination",
            "reasoning": "Compute the competing transition states for both
enantiomeric pathways using M06-2X/6-31G* with SMD(acetone) solvation. Optimize
 TS structures for the C-C bond formation step between the enolate and aldehyde
, verify with frequency calculations (one imaginary frequency), and perform IRC
 to confirm connectivity. Compare $\Delta G^{\ddagger}$ differences at $30°C$ to predict
enantiomeric excess, using the lower-level optimized geometries for single-
point energies with M06-2X/6-311+G** and SMD(acetone)."
        },
        {
            "strategy_name": "Catalyst-Substrate Pre-complexation and Non-
covalent Interaction Analysis",
            "reasoning": "Investigate the catalyst-substrate binding modes by
optimizing complexes of the chiral amine catalyst (N1CCCC1[C@H](N2CCCC2)C) with
 p-nitrobenzaldehyde and the triflic acid (CF3SO3H) with the ketone (CC(=O)C)
using M06-2X/6-31G* and SMD(acetone). Perform conformational sampling with
RDKit and Python scripts, then analyze non-covalent interactions (NCI) via NCI
plots to identify key stabilizing interactions (e.g., hydrogen bonding, pi-
stacking) that dictate enantioselectivity."
        },
        {
            "strategy_name": "Solvent Effect and Dielectric Screening on
Reaction Barriers",
            "reasoning": "Evaluate the impact of acetone solvent on the
reaction energetics by comparing gas-phase and SMD(acetone) Gibbs free energies
 for key steps (enolization, C-C bond formation, proton transfer). Use M06-2X
/6-311+G**//M06-2X/6-31G* with SMD for single-point calculations. Additionally,
 test alternative solvents (e.g., toluene, water) via SMD to understand how
dielectric constant and solvation effects influence the enantioselectivity and
```

```
rate-determining step."
        },
        {
            "strategy_name": "Energetic Span and Kinetic Analysis for Turnover
Frequency",
            "reasoning": "Apply the energetic span model (Kozuch and Shaik) to
compute the turnover frequency (TOF) and identify the rate-determining states
in the catalytic cycle. Calculate the full catalytic pathway including catalyst
 regeneration, using M06-2X/6-311+G**//M06-2X/6-31G* with SMD(acetone) at 30°C.
 Use Python scripts to automate Gibbs free energy corrections and TOF
calculations, comparing the energetic spans for diastereomeric pathways to
rationalize enantioselectivity."
        },
        {
            "strategy_name": "Benchmarking and Functional Validation for
Mechanism Robustness",
            "reasoning": "Validate the DFT methodology by benchmarking against
higher-level methods (e.g., DLPNO-CCSD(T)/def2-TZVP) for key transition states
and intermediates. Compare M06-2X with other functionals (e.g., ωB97X-D, B3LYP-
D3) for consistency in barrier heights and enantioselectivity predictions. Use
Gaussian for single-point calculations and PySCF for custom DFT setups,
ensuring the mechanistic conclusions are robust across computational methods."
        }
    ]
},
{
    "query": "How can machine learning and artificial intelligence techniques
be leveraged to accelerate the exploration of reaction pathways, predict
enantiomeric excess, and optimize catalytic performance in asymmetric synthesis
, and what are the key challenges in data generation, model training, and
validation for such interdisciplinary applications in computational chemistry
?",
    "hypotheses": [
        {
```

"strategy_name": "Transition State Analysis for Enantioselective C-C Bond Formation",
"reasoning": "Optimize all possible diastereomeric transition states for the C-C bond formation step between the enol(ate) of CC(=O)C and p-nitrobenzaldehyde using M06-2X/6-31G* with SMD(acetone). Compute relative Gibbs free energies at $30^{\circ}C$ using M06-2X/6-311+G**//M06-2X/6-31G* to predict enantiomeric excess. Validate TS structures via frequency calculations (one imaginary frequency) and IRC to confirm connectivity to enol intermediate and aldol product."
},
{
"strategy_name": "Catalyst-Substrate Complexation and Non-Covalent Interaction Analysis",
"reasoning": "Investigate pre-reactive complexes between the chiral amine catalyst N1CCCC1[C@H](N2CCCC2)C and p-nitrobenzaldehyde/CC(=O)C using conformational sampling (RDKit) and DFT optimization (M06-2X/6-31G*/SMD(acetone)). Perform NCI analysis to identify key steric repulsions and hydrogen bonding interactions that dictate facial selectivity. Compare binding energies and geometries for diastereomeric complexes to rationalize enantiocontrol."
},
{
"strategy_name": "Role of Brnsted Acid Co-catalyst in Rate Acceleration",
"reasoning": "Compute the energy profile for enolization of CC(=O)C catalyzed by C(F)(F)(F)S(=O)(=O)O using M06-2X/6-311+G**//M06-2X/6-31G* with SMD(acetone). Compare barriers with and without the acid co-catalyst, and analyze proton transfer TS structures. Evaluate the effect on the overall catalytic cycle by integrating this step with the aldol addition energetics to identify rate-determining steps."
},
{
"strategy_name": "Solvent Effects on Transition State Stabilization and Selectivity",
"reasoning": "Perform SMD calculations with varying solvent

```
parameters (dielectric constant, polarity) to probe how acetone influences the
relative energies of diastereomeric TS structures. Compare with gas-phase and
alternative solvents (e.g., toluene) to decouple electrostatic vs. non-
electrostatic contributions. Use M06-2X/6-311+G** for single-point energies on
gas-phase optimized TS geometries to isolate solvation effects."
        },
        {
            "strategy_name": "Energetic Span Model and Kinetic Analysis for
Turnover Frequency",
            "reasoning": "Construct the full catalytic cycle including
enolization, C-C bond formation, and catalyst regeneration. Compute Gibbs free
energies for all intermediates and TSs at 30°C using M06-2X/6-311+G**//M06-2X
/6-31G* with SMD(acetone). Apply the energetic span model (via Python scripts)
to identify rate-determining states and predict turnover frequencies, comparing
 with experimental reaction rates if available."
        }
    ]
},
{
    "query": "What advancements in computational spectroscopy, such as NMR and
IR prediction methods, coupled with quantum chemical calculations, are most
effective for validating proposed reaction intermediates and transition states
in catalytic cycles, and how do these methodologies bridge theoretical
predictions with experimental observables to provide mechanistic insights
across chemistry and materials science?",
    "hypotheses": [
        {
            "strategy_name": "Transition State Analysis for Enantioselectivity
Determination",
            "reasoning": "Compute and compare the Gibbs free energies of
competing diastereomeric transition states using M06-2X/6-31G* geometry
optimization followed by SMD(acetone)/M06-2X/6-311+G** single-point
calculations at 30°C. IRC analysis will verify connectivity to reactants and
products. The energy difference between TS enantiomers will predict
```

enantiomeric excess, which can be validated against experimental selectivity data."
        },
        {
            "strategy_name": "Catalyst-Substrate Complexation and Non-Covalent Interaction Analysis",
            "reasoning": "Optimize catalyst-substrate complexes using M06-2X/6-31G* with SMD(acetone) implicit solvation. Perform conformational sampling to identify lowest-energy structures. Use NCI plots and quantum theory of atoms in molecules (QTAIM) to analyze hydrogen bonding, steric repulsions, and dispersion interactions that govern enantiocontrol. Compare interaction patterns between enantiomeric pathways."
        },
        {
            "strategy_name": "Full Catalytic Cycle Energetic Profiling",
            "reasoning": "Map the complete catalytic cycle including catalyst activation, nucleophile formation, C-C bond formation, and product release steps. Compute Gibbs free energies for all intermediates and transition states using M06-2X/6-311+G**//M06-2X/6-31G* with SMD(acetone). Apply the energetic span model to identify rate-determining states and turnover frequencies."
        },
        {
            "strategy_name": "Solvent Effect Decomposition and Explicit Solvent Modeling",
            "reasoning": "Perform SMD calculations with different solvent parameters to decompose electrostatic vs. non-electrostatic contributions. Use QM/MM or cluster-continuum approaches with explicit acetone molecules around key transition states to assess specific solvent-catalyst interactions. Compare results with pure implicit solvation to evaluate solvent role in stereoselectivity."
        },
        {
            "strategy_name": "Functional and Dispersion Correction Benchmarking",

```
            "reasoning": "Benchmark multiple functionals (M06-2X, ωB97X-D,
    B3LYP-D3) and basis sets for key transition states and enantioselectivity
    predictions. Compare with higher-level methods (DLPNO-CCSD(T)) where feasible.
    Validate computational protocols against experimental ee values and establish
    error margins for predictive calculations in this catalytic system."
          }
        ]
    }
]
```

***Refined mechanistic hypotheses for the asymmetric catalysis validation***

```
[
  {
    "query": "What are the current state-of-the-art computational strategies for
    elucidating asymmetric catalytic mechanisms, including the integration of
    quantum mechanical methods with molecular dynamics, solvation models, and
    transition state theory, and how do these approaches account for
    enantioselectivity, solvent effects, and catalyst-substrate interactions in
    organocatalytic systems?",
    "optimized_hypotheses": [
      {
        "strategy_name": "Transition State Analysis for Enantioselectivity
    Determination",
        "reasoning": "Calculate the competing transition states for both
    enantiomeric pathways. Optimize transition state structures for the key bond
    formation step between the enolate and aldehyde, verify with frequency
    calculations to ensure a single imaginary frequency, and confirm connectivity
    to reactants and products. Compare Gibbs free energy differences at 30°C to
    predict enantiomeric excess, using optimized geometries for higher-accuracy
    energy calculations.",
        "complements": [
          {
            "strategy_name": "Integrated Transition State and Binding Mode Analysis
```

```
 for Enantioselectivity Prediction",
        "reasoning": "This approach combines accurate transition state energy
calculations to predict enantiomeric excess at 30°C with an analysis of
catalyst-substrate binding modes through conformational sampling and non-
covalent interaction analysis. It identifies key interactions (e.g., hydrogen
bonding, pi-stacking) that drive enantioselectivity, ensuring a comprehensive
understanding of both electronic and steric factors in the transition state and
 pre-reactive complexes."
      },
      {
        "strategy_name": "Hybrid Solvation-Enantioselectivity Workflow with
Multi-Solvent Benchmarking",
        "reasoning": "This strategy involves: (1) Computing enantiomeric
transition states with proper characterization; (2) Using high-accuracy energy
calculations for precise Gibbs free energy differences and enantiomeric excess
prediction at 30°C; (3) Analyzing solvent effects by comparing gas-phase and
solvated energetics across key steps; (4) Extending to alternative solvents (e.
g., toluene, water) to evaluate dielectric and solvation effects on
enantioselectivity and the rate-determining step, providing comprehensive
mechanistic and solvent optimization insights."
      },
      {
        "strategy_name": "Hybrid Enantioselectivity and Turnover Frequency
Analysis with Automated Workflow",
        "reasoning": "This approach combines enantioselectivity prediction (via
 transition state energy differences) with mechanistic insights from the
energetic span model (for turnover frequency and rate-determining states). It
uses high-accuracy energy calculations on optimized geometries, consistent
solvation at 30°C, and automation for efficiency. The strategy predicts both
enantiomeric excess and catalytic efficiency, offering a comprehensive view of
the catalytic cycle, including catalyst regeneration."
      },
      {
        "strategy_name": "Hybrid Benchmarking with High-Level Validation for
```

Enantioselectivity Prediction",
"reasoning": "This strategy involves: (1) Initial transition state optimization and connectivity validation, followed by high-accuracy energy calculations for precise enantiomeric excess prediction at $30°C$; (2) Validating the computational methodology by benchmarking against higher-level methods and alternative approaches for consistency in barrier heights and enantioselectivity, ensuring robustness across computational methods."
}
]
}
]
},
{
"query": "How can machine learning and artificial intelligence techniques be leveraged to accelerate the exploration of reaction pathways, predict enantiomeric excess, and optimize catalytic performance in asymmetric synthesis , and what are the key challenges in data generation, model training, and validation for such interdisciplinary applications in computational chemistry ?",
"optimized_hypotheses": [
{
"strategy_name": "Transition State Analysis for Enantioselective Bond Formation",
"reasoning": "Optimize all possible diastereomeric transition states for the key bond formation step between the enol(ate) and p-nitrobenzaldehyde. Compute relative Gibbs free energies at $30°C$ to predict enantiomeric excess. Validate transition state structures via frequency calculations (ensuring one imaginary frequency) and confirm connectivity to the enol intermediate and aldol product.",
"complements": [
{
"strategy_name": "Integrated Transition State and Pre-Reactive Complex Analysis for Enantioselectivity Prediction",
"reasoning": "This approach combines rigorous transition state optimization and energy evaluation for accurate enantiomeric excess prediction with conformational sampling and non-covalent interaction analysis of pre-reactive complexes. It correlates complex stability with transition state energies, providing a complete mechanistic picture from catalyst binding to bond formation that explains enantiocontrol."
      },
      {
        "strategy_name": "Integrated Enolization-Aldol Catalytic Cycle Analysis with Dual-Level Modeling",
        "reasoning": "This strategy includes: (1) High-accuracy diastereomeric transition state analysis for enantioselectivity prediction at $30°C$, with full validation; (2) Incorporating the enolization energy profile using consistent methodology; (3) Mapping the entire catalytic cycle to identify rate-determining steps and evaluate efficiency; (4) Ensuring chemical accuracy through consistent solvation modeling across all steps for reliable energy comparisons."
      },
      {
        "strategy_name": "Integrated Solvation-Aware Diastereomeric Transition State Analysis",
        "reasoning": "This strategy systematically locates and validates diastereomeric transition states for accurate enantioselectivity prediction, while incorporating comprehensive solvation analysis by computing energies with varying solvent parameters on gas-phase optimized geometries. This isolates electrostatic and non-electrostatic solvent effects while maintaining computational efficiency."
      },
      {
        "strategy_name": "Integrated Diastereomeric Transition State Optimization with Full Catalytic Cycle Energetic Span Analysis",
        "reasoning": "This approach preserves stereochemical precision by characterizing all diastereomeric transition states, while mapping the full catalytic cycle to analyze turnover frequency. It enables simultaneous prediction of enantioselectivity and catalytic efficiency under consistent

computational conditions at $30°C$, connecting stereochemical outcomes with kinetic performance."
        }
      ]
    }
  ]
},
{
  "query": "What advancements in computational spectroscopy, such as NMR and IR prediction methods, coupled with quantum chemical calculations, are most effective for validating proposed reaction intermediates and transition states in catalytic cycles, and how do these methodologies bridge theoretical predictions with experimental observables to provide mechanistic insights across chemistry and materials science?",
  "optimized_hypotheses": [
    {
      "strategy_name": "Transition State Analysis for Enantioselectivity Determination",
      "reasoning": "Compute and compare Gibbs free energies of competing diastereomeric transition states at $30°C$. Verify connectivity to reactants and products. The energy difference between transition state enantiomers predicts enantiomeric excess, which can be validated against experimental selectivity data.",
      "complements": [
        {
          "strategy_name": "Integrated Diastereomeric Transition State Analysis with Non-Covalent Interaction Mapping",
          "reasoning": "This strategy links quantitative energy-based enantioselectivity prediction with qualitative non-covalent interaction analysis. It maintains rigorous thermodynamic calculations for Gibbs free energies, while incorporating conformational sampling and interaction analysis to provide mechanistic insight into specific interactions (e.g., hydrogen bonding, steric, dispersion) responsible for energy differences, enabling both accurate enantiomeric excess prediction and understanding of enantiocontrol

```
origins."
      },
      {
        "strategy_name": "Integrated Catalytic Cycle Analysis with
Diastereomeric Transition State Validation",
        "reasoning": "This strategy maps the complete catalytic cycle (
including catalyst activation, nucleophile formation, bond formation, and
product release) to compute Gibbs free energies for all intermediates and
transition states at 30°C. The energetic span model identifies rate-determining
 states and turnover frequencies. For the stereoselective bond formation step,
competing diastereomeric transition states are analyzed, with energy
differences predicting enantiomeric excess, validated against experimental data
, providing both mechanistic understanding and predictive power."
      },
      {
        "strategy_name": "Hybrid Solvation-Enriched Diastereomeric Transition
State Analysis with Explicit-Implicit Solvation Decomposition",
        "reasoning": "This strategy integrates precise Gibbs free energy
comparisons for enantiomeric excess prediction with a robust solvent effect
analysis. It uses consistent computational conditions at 30°C, validated
against experiments, and incorporates solvation models to isolate electrostatic
 and non-electrostatic contributions, ensuring a comprehensive understanding of
 stereoselectivity origins."
      },
      {
        "strategy_name": "Benchmarked Protocol with High-Level Validation for
Diastereomeric Transition State Analysis and Enantioselectivity Prediction",
        "reasoning": "This strategy combines high-throughput screening of
diastereomeric transition states with rigorous benchmarking. It uses efficient
geometry optimization and high-accuracy energy calculations, validated against
higher-level methods and experimental enantiomeric excess values, ensuring both
 efficiency for routine screening and predictive accuracy."
      }
    ]
```

```
      }
    ]
  }
]
```

### C.8.2 Hypothesis for the radical reaction validation

*Initial mechanistic hypotheses for the radical reaction validation*

```
[
    {
        "query": "Computational strategies for distinguishing between concerted,
    stepwise nucleophilic/electrophilic substitution, and radical pathways in
    organic transformations under photochemical conditions, including benchmarking
    of DFT and wavefunction-based methods for barrier height prediction and
    reaction coordinate characterization",
        "hypotheses": [
            {
                "strategy_name": "Concerted SN2-like P-H assisted displacement of
    iodide",
                "reasoning": "Test the feasibility of a single-step nucleophilic
    substitution where HPPh2 directly attacks the benzylic-like carbon of Ph-CH2-
    CH2-CHI-BPin, displacing iodide in a concerted transition state with
    simultaneous proton transfer to the leaving group or base-mediated
    deprotonation. Computationally, optimize the reactants, products, and a
    putative SN2-type transition state at b3lyp/def2svp with SMD(EtOAc) solvation.
    Confirm the TS with one imaginary frequency and an IRC to connect reactants and
     products. Compare the computed activation free energy (ΔG‡) to typical
    solution-phase SN2 barriers; feasibility is suggested by ΔG‡ < ~ 25 kcal/mol."
            },
            {
                "strategy_name": "Stepwise ionic mechanism via base-promoted
    deprotonation of HPPh2 followed by nucleophilic substitution",
                "reasoning": "Evaluate whether Cs2CO3 deprotonates HPPh2 to form a
```

phosphide anion, which then undergoes bimolecular nucleophilic substitution at the $\alpha$-BPin carbon, displacing iodide. Model the deprotonation thermodynamics in SMD(EtOAc/CPME) and then optimize a phosphide-substrate SN2 transition state. The overall feasibility can be gauged by summing the free energy change of deprotonation and the computed $\Delta G^{\ddagger}$ for substitution; if the net barrier is low, the ionic pathway is plausible."
        },
        {
            "strategy_name": "Radical chain mechanism initiated by photoinduced C-I homolysis",
            "reasoning": "Test whether 450 nm light can cleave the C-I bond to initiate a radical chain. Use TD-DFT (e.g., b3lyp/def2svp/SMD) to compute vertical excitation energies and identify whether low-lying singlet or triplet states correlate with C-I $\sigma*$ antibonding orbitals. Compute the C-I bond dissociation free energy (BDFE) in SMD solvent. A low BDFE (< ~ 55 kcal/mol) and accessible excitation energy consistent with 450 nm photons (~ 63 kcal/mol) would support photolytic radical initiation. Subsequent radical recombination with HPPh2-derived radicals can be explored via barrierless or low-barrier HAT steps."
        },
        {
            "strategy_name": "Radical iodine atom transfer (XAT) from Ph-CH2-CH2-CHI-BPin to phosphinyl radical",
            "reasoning": "Explore a pathway where a phosphinyl radical (PPh2·) abstracts iodine from Ph-CH2-CH2-CHI-BPin, generating an $\alpha$-boryl radical. Compute the iodine atom transfer reaction free energy and barrier using b3lyp/def2svp/SMD. The boryl effect should stabilize the radical intermediate; quantify stabilization via spin density analysis and compare to literature $\alpha$-boryl radical energetics. A highly exergonic I-atom transfer with a small barrier (< ~ 10 kcal/mol) would indicate viability of a radical chain mechanism involving XAT."
        },
        {
            "strategy_name": "Concerted radical-polar crossover hydrogen atom

transfer (HAT) from HPPh2 to $\alpha$-boryl radical",
            "reasoning": "Assess whether the $\alpha$-boryl radical formed after C-I homolysis or XAT can abstract a hydrogen atom from HPPh2, generating an alkylboronate product and PPh2·. Compute the P-H bond dissociation free energy and the reaction free energy for HAT at b3lyp/def2svp/SMD. Optimize the HAT transition state and verify with IRC. If HAT is exergonic and has a small barrier (< ~ 8 kcal/mol), this would support a radical-polar crossover process driving the reaction forward under photochemical conditions."
        }
    ]
},
{
    "query": "Theoretical and computational approaches to modeling visible-light-induced homolytic bond cleavage in organohalides, with emphasis on calculating excitation energies, bond dissociation enthalpies, and spin density distributions to assess radical initiation feasibility",
    "hypotheses": [
        {
            "strategy_name": "Direct Concerted Nucleophilic Substitution at Carbon (S_N2-type) by HPPh2 on Ph-CH2-CH2-CHI-BPin",
            "reasoning": "Model the possibility of a concerted displacement of iodide from Ph-CH2-CH2-CHI-BPin by HPPh2 under ground-state conditions. Perform a transition state search at the b3lyp/def2svp level with SMD solvation (EtOAc/CPME) to locate a single imaginary frequency structure leading from reactants to Ph-CH2-CH2-CH2-BPin and IPPh2. Verify connectivity via IRC analysis. Compute the free energy barrier ($\Delta G^{\ddagger}$) and compare with a typical thermal barrier (<25 kcal/mol for room temperature feasibility). An unreasonably high barrier would suggest that a purely concerted nucleophilic substitution pathway is unlikely under the reported mild photochemical conditions."
        },
        {
            "strategy_name": "Stepwise Radical Pathway via Photoinduced C-I Homolysis",
            "reasoning": "Assess whether 450 nm light can induce C-I bond

homolysis in Ph-CH2-CH2-CHI-BPin to form $\alpha$-boryl alkyl radicals. Use TD-DFT (
b3lyp/def2svp) to compute vertical excitation energies and identify low-lying
excited states with $\sigma*$_{C-I} character. Calculate the C-I bond dissociation
enthalpy (BDE) at  b3lyp/def2svp with SMD solvation and compare to the
available photon energy (~ 63.5 kcal/mol). If the excitation energy is below
the photon energy and the BDE is moderate (<60 kcal/mol), the radical
initiation step is thermodynamically accessible. Follow with unrestricted DFT
optimization of the radical species to confirm stability and $\alpha$-boryl
stabilization."
        },
        {
            "strategy_name": "Radical Chain via Hydrogen Atom Transfer (HAT)
from HPPh2 to $\alpha$-Boryl Radical",
            "reasoning": "Test the feasibility of HPPh2 serving as a hydrogen
atom donor to the $\alpha$-boryl radical intermediate. Compute the P-H bond
dissociation enthalpy (BDE) of HPPh2 and the C-H BDE of Ph-CH2-CH2-CH2-BPin at
b3lyp/def2svp/SMD. Model the HAT step by optimizing the transition state for
hydrogen transfer from HPPh2 to the $\alpha$-boryl radical. A low barrier (<10 kcal/
mol) and exergonic profile would support a radical chain propagation mechanism
where P-I bond formation is the chain-terminating step."
        },
        {
            "strategy_name": "Single Electron Transfer (SET) from HPPh2 to
Photoexcited Ph-CH2-CH2-CHI-BPin",
            "reasoning": "Evaluate whether photoexcited Ph-CH2-CH2-CHI-BPin can
 accept an electron from HPPh2, generating a radical cation of HPPh2 and an $\alpha$-
boryl radical. Use TD-DFT (b3lyp/def2svp) to estimate the excited-state
reduction potential of Ph-CH2-CH2-CHI-BPin and b3lyp/def2svp to compute the
oxidation potential of HPPh2. Apply the Rehm-Weller equation to estimate the
free energy change for SET. If $\Delta G$ is negative and the driving force exceeds ~
0.1 eV, a SET-initiated radical pathway is plausible."
        },
        {
            "strategy_name": "Concerted Halogen Atom Transfer (XAT) from Ph-CH2

-CH2-CHI-BPin to HPPh2-derived Phosphorus Radical",
"reasoning": "Model a concerted halogen atom transfer in which a P-centered radical abstracts iodine from Ph-CH2-CH2-CHI-BPin, forming IPPh2 and an $\alpha$-boryl radical. Generate the P-centered radical by removing an H atom from HPPh2 computationally. Optimize the transition state for direct I-atom transfer at b3lyp/def2svp/SMD and compute the barrier. If the barrier is low (<10 kcal/mol) and the step is highly exergonic due to strong P-I bond formation (~ 55-60 kcal/mol), this mechanism could complement or compete with pure photolytic C-I cleavage."
}
]
},
{
"query": "Cross-disciplinary methodologies integrating quantum chemical simulations, kinetic modeling, and electronic structure analysis to evaluate thermodynamic driving forces and kinetic accessibility of heteroatom-halogen bond formation in organophosphorus and organoboron systems",
"strategy_name": "Photoinduced Radical C-I Bond Homolysis Followed by H-Atom Transfer from HPPh2",
"reasoning": "Test the hypothesis that 450 nm light promotes homolytic cleavage of the C-I bond in Ph-CH2-CH2-CHI-BPin, generating an $\alpha$-boryl-stabilized alkyl radical, which then abstracts a hydrogen atom from HPPh2 to yield Ph-CH2-CH2-CH2-BPin and a P-centered radical. The P-centered radical could then recombine with iodine to form IPPh2. Computationally, use TD-DFT (e.g., b3lyp/def2svp with SMD solvation for EtOAc/CPME) to compute vertical excitation energies and oscillator strengths for the C-I chromophore, and unrestricted DFT (b3lyp/def2svp) to calculate the C-I bond dissociation free energy. Compare the photochemical excitation energy with the photon energy at 450 nm (~ 2.75 eV) and assess whether the BDE is low enough for photolysis to be feasible."
},
{
"strategy_name": "Concerted Nucleophilic Substitution at Carbon by HPPh2 with I-P Bond Formation",

"reasoning": "Assess the possibility of a concerted SN2-like displacement of iodide from Ph-CH2-CH2-CHI-BPin by HPPh2, forming Ph-CH2-CH2-CH2-BPin and IPPh2 in a single step without discrete radicals. Model the reaction using DFT ( b3lyp/def2svp, SMD solvent) by optimizing reactants, products, and a single transition state structure. Perform an Intrinsic Reaction Coordinate (IRC) calculation to verify connectivity. Compare the activation free energy ($\Delta G^{\ddagger}$) with typical thermal barriers (<25 kcal/mol) to determine feasibility under mild photochemical conditions; if too high, this pathway is unlikely."
},
{
"strategy_name": "Stepwise Radical Substitution via Iodine Atom Transfer (XAT) from Ph-CH2-CH2-CHI-BPin to P-Centered Radical",
"reasoning": "Investigate whether an initial P-centered radical, generated by H-abstraction from HPPh2 under photochemical conditions (e.g., via photoexcited base or solvent), abstracts iodine from Ph-CH2-CH2-CHI-BPin to yield Ph-CH2-CH2-CH2-BPin radical and IPPh2 directly. Use unrestricted DFT ( b3lyp/def2svp, SMD) to compute the free energy change for iodine atom transfer from alkyl iodide to P-centered radical. Also calculate the barrier for the XAT step via a transition state search. A strongly exergonic and low-barrier XAT would support this mechanism."
},
{
"strategy_name": "Radical Chain Mechanism Initiated by Photoexcited HPPh2 or Solvent",
"reasoning": "Examine whether HPPh2 or the solvent forms radicals upon 450 nm excitation, initiating a chain process in which an $\alpha$-boryl alkyl radical is generated and propagates the reaction. Use TD-DFT (b3lyp/def2svp) to compute excitation energies of HPPh2 and solvents (EtOAc, CPME) to check if they absorb at 450 nm. Calculate P-H and C-H bond dissociation free energies ( b3lyp/def2svp) to assess ease of radical generation. Model a representative radical propagation step to ensure barriers are consistent with a rapid chain process (<10 kcal/mol)."
},

{
"strategy_name": "Base-Mediated Electron Transfer Leading to Radical Anion and Fragmentation",
"reasoning": "Test whether Cs2CO3 facilitates a single-electron transfer (SET) to Ph-CH2-CH2-CHI-BPin, producing a radical anion that fragments into an $\alpha$-boryl radical and iodide. Use DFT (b3lyp/def2svp, SMD) to compute the reduction potential of Ph-CH2-CH2-CHI-BPin and compare it with the oxidation potential of a plausible electron donor species (e.g., HPPh2 anion). Model the fragmentation step of the radical anion to determine the barrier and reaction free energy. A low barrier and exergonic fragmentation would indicate that base-assisted SET is a viable initiation pathway."
}
]
},
{
"query": "Advances in simulating photochemical reaction mechanisms using nonadiabatic dynamics, time-dependent DFT, and multireference methods, with applications to energy transfer, hydrogen atom transfer, and halogen atom transfer processes in catalysis",
"hypotheses": [
{
"strategy_name": "Direct Photoinduced C-I Homolysis Followed by Radical Recombination",
"reasoning": "Model the absorption spectrum of Ph-CH2-CH2-CHI-BPin using TD-DFT (e.g., b3lyp/def2svp with SMD solvation for EtOAc/CPME) to check whether 450 nm excitation is energetically accessible. Compute the C-I bond dissociation energy (BDE) in the ground state and in the lowest singlet/triplet excited states to assess feasibility of photoinduced homolysis. Follow with unrestricted DFT (b3lyp/def2svp) optimization of the $\alpha$-boryl radical and IPPh2 radical, and evaluate the thermodynamics of radical recombination to form Ph-CH2-CH2-CH2-BPin and IPPh2. This will test whether a radical chain mechanism initiated by light is plausible."
},
{

"strategy_name": "Radical Iodine Atom Transfer (XAT) from Ph-CH2-CH2-CHI-BPin to HPPh2-Derived Radical",
"reasoning": "Simulate homolytic P-H bond cleavage in HPPh2 by computing its BDE (b3lyp/def2svp) and compare to the energy gain from forming a P-I bond (HPPh2 + I· $\rightarrow$ IPPh2). Model the $\alpha$-boryl radical stabilization by computing the relative energies of Ph-CH2-CH2-CH·-BPin vs. a simple alkyl radical. Use a simple radical step model (Ph-CH2-CH2-CHI-BPin + $cdot$PPh2) to compute the barrier for iodine atom transfer at the b3lyp/def2svp level. This assesses if XAT is thermodynamically and kinetically viable under photochemical initiation."
},
{
"strategy_name": "Concerted Nucleophilic Substitution at Carbon by HPPh2 with Iodide Displacement",
"reasoning": "Optimize the transition state for a concerted SN2-like attack of HPPh2 at the $\alpha$-carbon of Ph-CH2-CH2-CHI-BPin (b3lyp/def2svp with SMD solvent model). Perform frequency analysis to confirm a single imaginary frequency connecting reactants and products. Compare the computed activation barrier to typical photochemical reaction energy input. Assess the role of $\alpha$-BPin in stabilizing the developing negative charge by comparing barriers with and without the BPin group. This tests whether a ground-state concerted pathway could compete with radical routes."
},
{
"strategy_name": "Base-Promoted Electron Transfer (ET) to Induce C-I Cleavage",
"reasoning": "Model the Cs2CO3-assisted deprotonation of HPPh2 to form the PPh2- anion and compute its vertical ionization potential and electron affinity. Use a Marcus theory framework with computed redox potentials (b3lyp/def2svp, SMD solvent) for PPh2- and Ph-CH2-CH2-CHI-BPin to determine whether single-electron transfer is thermodynamically favorable. If favorable, optimize the geometry of the radical anion of Ph-CH2-CH2-CHI-BPin and compute the barrier for C-I fragmentation. This tests a SET-initiated pathway under basic conditions."

```
        },
        {
            "strategy_name": "Photoinduced Excited-State Concerted Hydrogen
Atom Transfer (HAT) and Iodine Atom Transfer",
            "reasoning": "Using TD-DFT (b3lyp/def2svp, SMD solvent), optimize
the lowest singlet and triplet excited states of Ph-CH2-CH2-CHI-BPin and HPPh2
in a weakly bound encounter complex. Scan a reaction coordinate corresponding
to simultaneous HAT from HPPh2 to the α-carbon radical site and IAT from Ph-CH2
-CH2-CHI-BPin to PPh2. Identify whether a low-energy crossing point or conical
intersection could facilitate a concerted photochemical step. This will
evaluate whether light allows a synchronous HAT/IAT event via an excited-state
surface."
        }
    ]
},
{
    "query": "Machine learning and data-driven approaches for predicting
reactivity trends, radical stabilization effects, and substituent electronic
influences in photoredox and radical-mediated bond-forming reactions, including
 feature engineering from quantum chemical descriptors and reaction databases",
    "hypotheses": [
        {
            "strategy_name": "Radical C-I Homolysis Followed by Hydrogen Atom
Transfer (HAT) from HPPh2",
            "reasoning": "Test the hypothesis that blue-light irradiation
promotes homolytic cleavage of the C-I bond in Ph-CH2-CH2-CHI-BPin to generate
an α-boryl alkyl radical, which then abstracts a hydrogen atom from HPPh2,
forming Ph-CH2-CH2-CH2-BPin and a P-centered radical that captures iodine.
Computationally, determine the vertical excitation energies (TD-DFT, e.g. b3lyp
/def2svp/SMD) to assess whether 450 nm light can promote a dissociative excited
 state. Then compute C-I bond dissociation enthalpy in the ground state and in
the lowest singlet/triplet excited states. Evaluate the thermodynamics of P-H
bond cleavage and P-I bond formation. A low BDE for C-I and favorable
energetics for P-I formation would support this radical photochemical pathway's
```

```
feasibility."
        },
        {
            "strategy_name": "Concerted Nucleophilic Substitution via Phosphide
Attack on Carbon",
            "reasoning": "Examine the possibility that under basic conditions (
Cs2CO3), HPPh2 is deprotonated to Ph2P-, which directly displaces I- from Ph-
CH2-CH2-CHI-BPin in a concerted SN2-type transition state, forming IPPh2 and Ph
-CH2-CH2-CH2-BPin in a single step. Computationally, optimize the SN2
transition state at b3lyp/def2svp/SMD(EtOAc/CPME), confirm connectivity via IRC
, and compute the activation free energy. Compare the barrier to typical SN2
thresholds (~ 15-25 kcal/mol in solution). If the barrier is low and the α-BPin
substituent stabilizes the transition state, the concerted ionic pathway is
plausible."
        },
        {
            "strategy_name": "Radical Chain via Iodine Atom Transfer (XAT) from
Ph-CH2-CH2-CHI-BPin to P-Centered Radical",
            "reasoning": "Investigate whether a P-centered radical, generated
photochemically or thermally from HPPh2, abstracts iodine from Ph-CH2-CH2-CHI-
BPin to yield an α-boryl radical and IPPh2 radical, followed by radical
recombination with hydrogen atom donors to produce Ph-CH2-CH2-CH2-BPin.
Computational test: calculate the enthalpy change for iodine atom transfer from
Ph-CH2-CH2-CHI-BPin to Ph2P· and the associated activation barrier via open-
shell DFT (b3lyp/def2svp/SMD). Use spin density analysis to confirm radical
centers. A strongly exergonic and low-barrier XAT step would support a chain
propagation mechanism."
        },
        {
            "strategy_name": "Single Electron Transfer (SET) Initiated Pathway
",
            "reasoning": "Assess whether photoexcited HPPh2 or an electron-rich
base-HPPh2 complex can donate an electron to Ph-CH2-CH2-CHI-BPin, forming a
radical anion that undergoes C-I bond cleavage to yield an α-boryl radical and
```

```
iodide. Computationally, calculate adiabatic and vertical ionization potentials
 (IP) of HPPh2 and reduction potential of Ph-CH2-CH2-CHI-BPin (via free energy
of electron addition) in SMD solvent model. Compare the computed excitation
energy of HPPh2/Base complex to the 450 nm photon energy to gauge feasibility.
If the SET step is thermodynamically favorable and within the photon energy,
the SET-triggered radical pathway is plausible."
        },
        {
            "strategy_name": "α-Bond Metathesis-like Concerted Pathway between
P-H and C-I Bonds",
            "reasoning": "Explore the possibility of a concerted four-centered
transition state in which the P-H bond of HPPh2 and the C-I bond of Ph-CH2-CH2-
CHI-BPin undergo simultaneous exchange, forming Ph-CH2-CH2-CH2-BPin and IPPh2
without discrete radical intermediates. Computationally, locate the cyclic
transition state using b3lyp/def2svp/SMD, confirm it with frequency analysis
and IRC. Evaluate barrier heights and thermodynamics. If the barrier is
competitive with photochemically accessible energies and the geometry shows
synchronous bond-making/breaking, this would support a non-radical concerted σ-
bond metathesis mechanism."
        }
    ]
},
{
"query": "Computational and mechanistic strategies for evaluating visible-light
-driven radical, ionic, and donor-acceptor complex pathways in the
transformation of beta-boryl alkyl iodides with secondary phosphines under
basic conditions, including assessment of excitation feasibility, SET/XAT/HAT
steps, and the role of cesium-assisted activation",
"hypotheses": [
    {
        "strategy_name": "EDA/SRN1-Type Radical Chain Mechanism under Visible-
Light Irradiation",
        "reasoning": "Hypothesis: The observed transformation of CH3-CHI-(B-)
Pin-PPh2 into Ph-CH2-CH2-CH2-BPin and Ph2P-PPh2 under 450 nm irradiation arises
```

not from direct high-energy excitation of the substrate (vertical excitation ~ 102.5 kcal/mol, far above photon energy at 450 nm ~ 63.5 kcal/mol), but through formation of an electron donor-acceptor (EDA) complex between $Ph2P^-$ ( generated by Cs2CO3 deprotonation of HPPh2) and the C-I bond of the substrate. Upon charge-transfer excitation (~ 450 nm), a single-electron transfer (SET) occurs into the C-I $\sigma*$ orbital, inducing bond cleavage to generate a $\beta$-boryl carbon radical, iodide, and a phosphorus-centered radical. The carbon radical abstracts hydrogen from HPPh2 (HAT), yielding Ph-CH2-CH2-CH2-BPin and regenerating Ph2P·. Concurrently, Ph2P· radicals dimerize to form Ph2P-PPh2. The mechanism rationalizes the need for 2 equivalents of HPPh2 (one portion as deprotonated donor, another as hydrogen source), the observed P-P coupling product, and the feasibility of initiation under visible light despite insufficient photon energy for direct homolysis."
    },
    {
        "strategy_name": "Cesium-Assisted Activation Complex Promoting Visible-Light-Induced C-I Cleavage",
        "reasoning": "Hypothesis: Rather than a free EDA complex formed solely between $Ph2P^-$ and the substrate, the key photoactive species is a higher-order cesium-associated activation complex involving the $\beta$-boryl alkyl iodide, Cs+ derived from Cs2CO3, and possibly phosphide or phosphine. Coordination of Cs+ to the boronate oxygen atoms and/or iodide polarizes the C-I bond, lowers the energy of the acceptor orbital, and stabilizes charge-separated excited states, thereby shifting the effective excitation energy into the range accessible by 450 nm irradiation. After photoexcitation of this cesium-organized complex, C-I bond cleavage generates a $\beta$-boryl radical that undergoes hydrogen atom transfer from HPPh2 to produce Ph-CH2-CH2-CH2-BPin, while phosphorus-centered radicals dimerize to Ph2P-PPh2. This hypothesis is attractive because it explains why the reaction depends strongly on cesium base, why direct excitation of the isolated substrate is not feasible, and why visible light can nonetheless enable radical generation under the experimental conditions."
    },
    {
        "strategy_name": "Direct Ground-State Concerted Ionic Substitution by

```
Phosphide Is Unfavorable",
        "reasoning": "Hypothesis: A possible competing pathway is a
conventional ionic mechanism in which Cs2CO3 deprotonates HPPh2 to generate
$Ph2P^{-}$, followed by direct backside substitution at the β-boryl carbon with
concerted C-I bond cleavage under ground-state conditions. In this scenario,
the transformation would proceed through an SN2-like or highly asynchronous
substitution transition state without discrete radical intermediates, and
subsequent proton transfer or secondary processes would furnish Ph-CH2-CH2-CH2-
BPin. However, because the product distribution includes the phosphorus dimer
Ph2P-PPh2, and because visible-light irradiation is required experimentally,
this pathway is likely disfavored or incomplete as the dominant mechanism.
Computationally, this hypothesis should be tested as a control to determine
whether the thermal substitution barrier is prohibitively high and whether the
ionic pathway fails to explain the photochemical dependence and radical-
compatible byproduct formation."
    },
    {
        "strategy_name": "Phosphorus-Centered Radical XAT/HAT Chain after
Indirect Radical Initiation",
        "reasoning": "Hypothesis: The reaction may proceed through a radical
chain in which the initial photoevent generates a small amount of phosphorus-
centered radical or β-boryl carbon radical by an indirect pathway, after which
chain propagation becomes dominant. In the propagation phase, a phosphorus-
centered radical abstracts iodine from the β-boryl alkyl iodide through halogen
 atom transfer (XAT), producing a β-boryl carbon radical and a phosphorus-
iodine species or iodine-bound radical intermediate. The β-boryl carbon radical
 then abstracts a hydrogen atom from HPPh2 (HAT) to form Ph-CH2-CH2-CH2-BPin
and regenerate the phosphorus-centered radical. Radical coupling of two
phosphorus-centered radicals provides Ph2P-PPh2 as a chain-termination product.
 This mechanism rationalizes the observed dimer, the role of HPPh2 as both H-
atom donor and radical reservoir, and the possibility that only the initiation
step is photochemical whereas the bulk of turnover proceeds through thermally
accessible radical propagation steps."
    },
```

{
"strategy_name": "Photoinduced SET from a Phosphide/Base Aggregate Rather Than a Simple Binary EDA Complex",
"reasoning": "Hypothesis: Instead of a discrete binary EDA complex between Ph2P- and the substrate, the electronically active donor may be a phosphide/base aggregate, such as a cesium phosphide ion pair or a carbonate-associated phosphide cluster, which possesses enhanced reducing power and altered absorption properties under the reaction conditions. Upon visible-light excitation, this donor assembly transfers an electron to the $\beta$-boryl alkyl iodide, yielding a substrate radical anion that fragments rapidly to a $\beta$-boryl carbon radical and iodide. The resulting carbon radical abstracts hydrogen from HPPh2 to form Ph-CH2-CH2-CH2-BPin, while the oxidized phosphorus-derived radical species ultimately dimerizes to Ph2P-PPh2. This hypothesis is mechanistically distinct from the simple EDA picture because the spectroscopically relevant donor is not free $Ph2P^-$ itself but a higher-order ion-paired aggregate; such a model may better capture solvent, counterion, and aggregation effects that are often essential in photochemical electron-transfer systems involving alkali metal bases."
}
]
}
]

***Refined mechanistic hypotheses for the radical reaction validation***

[
{
"query": "Computational strategies for distinguishing between concerted, stepwise nucleophilic/electrophilic substitution, and radical pathways in organic transformations under photochemical conditions, including benchmarking of DFT and wavefunction-based methods for barrier height prediction and reaction coordinate characterization",
"optimized_hypotheses": [
{

"strategy_name": "Concerted SN2-like P\u2013H assisted displacement of iodide",
    "reasoning": "Test the feasibility of a single-step nucleophilic substitution where HPPh2 directly attacks the benzylic-like carbon of Ph-CH2-CH2-CHI-BPin, displacing iodide in a concerted transition state with simultaneous proton transfer to the leaving group or base-mediated deprotonation. Computationally, optimize the reactants, products, and a putative SN2-type transition state at M06-2X/def2-TZVP with SMD(EtOAc) solvation. Confirm the TS with one imaginary frequency and an IRC to connect reactants and products. Compare the computed activation free energy (ΔG‡) to typical solution-phase SN2 barriers; feasibility is suggested by ΔG‡ < ~ 25 kcal/mol.",
    "complements": [
      {
        "strategy_name": "Integrated Concerted/Ionic SN2 Pathway Assessment",
        "reasoning": "To preserve the strengths of both hypotheses, model a unified mechanistic landscape in which the nucleophilic substitution at the \u03b1-BPin\u2013CH2\u2013I carbon is explored along two connected coordinates: (a) a direct, neutral HPPh2 attack with a concerted proton-transfer component, and (b) a base-assisted pathway involving prior Cs2CO3-mediated deprotonation to the phosphide anion. For the concerted neutral route, perform M06-2X/def2-TZVP optimizations with SMD(EtOAc) of reactants, products, and a single SN2-like transition state exhibiting coupled proton transfer; verify by one imaginary frequency and IRC analysis, and compare ΔG‡ to ~ 25 kcal/mol for feasibility. For the ionic route, compute the free energy of HPPh2 deprotonation by Cs2CO3 in SMD(EtOAc) and/or SMD(CPME), then locate and verify a phosphide\u2013substrate SN2 TS. The overall free-energy profile would sum the deprotonation step and the barrier of substitution. Comparing barrier heights and thermodynamic profiles side-by-side identifies whether the reaction is kinetically accessible via the neutral concerted path, the ionic phosphide path, or both, preserving hypothesis (1)’s one-step concerted assessment while integrating hypothesis (2)’s deprotonation energetics and anionic reactivity."
      },
      {

"strategy_name": "Dual-Pathway Reactivity Profiling for HPPh2 and Ph-CH2-CH2-CHI-BPin",
"reasoning": "To maximize mechanistic insight while preserving the strengths of both hypotheses, the study should computationally investigate two parallel pathways from the same initial reactants. For the concerted polar pathway, model a single-step SN2-type displacement at the benzylic-like carbon by HPPh2 with either synchronous proton transfer to iodide or base-assisted deprotonation. Optimize reactants, products, and putative transition states at the M06-2X/def2-TZVP level in SMD(EtOAc), verify each TS by a single imaginary frequency and IRC connection, and compare the computed $\Delta G^{\ddagger}$ to the ~ 25 kcal/mol feasibility threshold. In parallel, assess the photoinitiated radical pathway by performing TD-DFT (ωB97X-D/def2-TZVP/SMD) vertical excitation scans to identify singlet or triplet states with C\u2013I \u03c3* character and energies near 450 nm (~ 63 kcal/mol). Calculate the C\u2013I bond dissociation free energy in the same solvation model; if BDFE is below ~ 55 kcal/mol and excitation is accessible with 450 nm light, proceed to model downstream radical recombination with HPPh2-derived species via low-barrier hydrogen atom transfer or radical coupling. This integrated approach preserves the kinetic/energetic evaluation of the SN2 mechanism while incorporating photophysical and thermochemical analysis of a possible radical chain, enabling direct comparison of activation barriers and thermodynamic driving force to identify the most viable pathway under experimental conditions."
},
{
"strategy_name": "Integrated Polar\u2013Radical Mechanistic Screening for HPPh2 + Ph-CH2-CH2-CHI-BPin",
"reasoning": "To preserve the advantages of both hypotheses, the computational study can concurrently evaluate the polar (concerted SN2-type) and radical (iodine atom transfer) pathways on the same footing with a consistent level of theory, M06-2X/def2-TZVP/SMD(EtOAc). First, optimize HPPh2 and Ph-CH2-CH2-CHI-BPin reactants, the SN2 product (Ph2P\u2013CH2\u2013CH(BPin)\u2013CH3 analogue depending on orientation), and a putative single-step SN2-like TS in which HPPh2 attacks the benzylic\u2013like carbon as $I^-$ departs, with either concerted proton transfer to a base or to the leaving group.

Confirm the TS with one imaginary frequency and intrinsic reaction coordinate ( IRC) to connect it to the reactants and products, and quantify the activation barrier (ΔG‡), considering ΔG‡ < 25 kcal/mol as accessible in solution. In parallel, model a radical mechanism starting from a phosphinyl radical (PPh2\u00b7) and Ph-CH2-CH2-CHI-BPin to compute the free energy change and barrier for iodine atom abstraction, yielding an \u03b1-boryl radical; analyze the unpaired spin distribution to quantify \u03b1-boryl stabilization and benchmark against literature values. A strongly exergonic, low-barrier (< 10 kcal/mol) I-atom transfer would support feasibility of an XAT radical chain. By directly comparing computed thermodynamic and kinetic parameters for both mechanisms, the approach identifies the dominant pathway under given conditions while retaining mechanistic breadth and the ability to rationalize polar/radical reactivity trends."
      },
      {
        "strategy_name": "Integrated Polar\u2013Radical Mechanistic Screening for HPPh2 + Ph-CH2-CH2-CHI-BPin Reactivity",
        "reasoning": "To preserve the advantages of both hypotheses, perform a two-pathway DFT study that evaluates whether the reaction proceeds more feasibly via a concerted SN2-type pathway or a radical-polar crossover sequence. First, compute the single-step substitution where HPPh2 directly attacks the benzylic-like carbon of Ph-CH2-CH2-CHI-BPin, displacing iodide, with simultaneous P\u2013H proton transfer to the leaving group or a base; optimize reactants, products, and an SN2 transition state at M06-2X/def2-TZVP/SMD(EtOAc), confirm via one imaginary frequency and IRC, and quantify ΔG‡, comparing to the ~ 25 kcal/mol threshold for solution-phase SN2 viability. In parallel, explore the radical possibility: calculate the free energy for \u03b1-boryl radical formation (C\u2013I homolysis/XAT implied), determine the P\u2013H bond dissociation free energy of HPPh2, and compute reaction and activation free energies for hydrogen atom transfer (HAT) from HPPh2 to the \u03b1-boryl radical. Optimize the HAT TS and verify with IRC. If the HAT is exergonic and the barrier is \u2264 8 kcal/mol, this supports radical-polar crossover relevance under photochemical conditions. Combining these calculations will identify the lower-barrier pathway while retaining insight into both polar and

```
  radical mechanisms, enabling a comprehensive mechanistic proposal."
        }
      ]
    }
  ]
},
{
  "query": "Theoretical and computational approaches to modeling visible-light-
  induced homolytic bond cleavage in organohalides, with emphasis on calculating
  excitation energies, bond dissociation enthalpies, and spin density
  distributions to assess radical initiation feasibility",
  "optimized_hypotheses": [
    {
      "strategy_name": "Direct Concerted Nucleophilic Substitution at Carbon (
  S_N2-type) by HPPh2 on Ph-CH2-CH2-CHI-BPin",
      "reasoning": "Model the possibility of a concerted displacement of iodide
  from Ph-CH2-CH2-CHI-BPin by HPPh2 under ground-state conditions. Perform a
  transition state search at the ωB97X-D/def2-TZVP level with SMD solvation (
  EtOAc/CPME) to locate a single imaginary frequency structure leading from
  reactants to Ph-CH2-CH2-CH2-BPin and IPPh2. Verify connectivity via IRC
  analysis. Compute the free energy barrier (ΔG‡) and compare with a typical
  thermal barrier (<25 kcal/mol for room temperature feasibility). An
  unreasonably high barrier would suggest that a purely concerted nucleophilic
  substitution pathway is unlikely under the reported mild photochemical
  conditions.",
      "complements": [
        {
          "strategy_name": "Integrated Ground-State vs Photochemical Pathway
  Analysis for Ph-CH2-CH2-CHI-BPin Reactivity",
          "reasoning": "To discriminate between a concerted nucleophilic
  substitution and a photochemically initiated radical pathway under mild 450 nm
  irradiation, combine both hypotheses in a complementary computational workflow.
   First, model the ground-state reaction of HPPh2 with Ph-CH2-CH2-CHI-BPin by
  locating a single transition state at the ωB97X-D/def2-TZVP level with SMD
```

solvation (EtOAc/CPME), ensuring it connects reactants to Ph-CH2-CH2-CH2-BPin + IPPh2 via IRC analysis. A computed ΔG‡ significantly above ~ 25 kcal/mol would argue against thermal feasibility at room temperature. Next, assess photochemical C\u2013I bond homolysis potential via TD-DFT (CAM-B3LYP/def2-TZVP ) to identify low-lying states with \u03c3*_{C\u2013I} character and compare excitation energy to the 450 nm photon energy (~ 63.5 kcal/mol). Compute the ω B97X-D/def2-TZVP/SMD BDE for C\u2013I in Ph-CH2-CH2-CHI-BPin; if the BDE is below ~ 60 kcal/mol and accessible excited states exist below the photon energy , photoinitiated \u03b1-boryl radical formation is plausible. Validate radical stability via unrestricted DFT optimization. This unified approach preserves the mechanistic discrimination power of the ground-state barrier analysis while incorporating the energetic and orbital insights necessary for evaluating photoinduced radical pathways."
      },
      {
        "strategy_name": "Hybrid Ground-State vs Radical-Chain Mechanistic Probe",
        "reasoning": "The two hypotheses address complementary mechanistic extremes: (1) a concerted bimolecular displacement of iodide from Ph-CH2-CH2-CHI-BPin by HPPh2 under ground-state conditions, and (2) a hydrogen atom transfer from HPPh2 to an \u03b1-boryl radical relevant for a radical chain process. A unified workflow first performs a transition state search at the ω B97X-D/def2-TZVP level with SMD solvation in a mixed EtOAc/CPME environment for the concerted substitution pathway, confirming the nature of the TS via a single imaginary frequency and IRC connectivity, and evaluating ΔG‡ against the ~ 25 kcal/mol threshold for thermal feasibility at room temperature. If the computed barrier is significantly higher, one proceeds to radical considerations by computing the P\u2013H BDE of HPPh2 and the C\u2013H BDE of CH3\u2013CH2\u2013BPin at the same level of theory/solvation. The HAT event from HPPh2 to the \u03b1-boryl radical is then explicitly modeled via transition-state optimization, estimating both the kinetic barrier ( benchmarking against \u224810 kcal/mol for fast radical processes) and reaction thermodynamics. This hybrid approach preserves the advantages of both methods \u2014 rigorous evaluation of feasibility for the polar ground-state path and

quantitative probing of the radical chain propagation step \u2014 and allows a direct mechanistic decision based on computed energetics."
      },
      {
        "strategy_name": "Hybrid Ground- and Photo-Induced Mechanistic Evaluation",
        "reasoning": "The combined approach first establishes whether a concerted ground-state nucleophilic displacement of iodide by HPPh2 on Ph-CH2-CH2-CHI-BPin is kinetically accessible under mild thermal conditions by locating a ωB97X-D/def2-TZVP/SMD(EtOAc/CPME) transition state connecting reactants to CH3\u2013CH2\u2013BPin and IPPh2, confirming a single imaginary frequency and connectivity via IRC, and computing $\Delta G^{\ddagger}$. A barrier significantly above ~ 25 kcal/mol would indicate that the concerted SN2-type pathway is implausible at room temperature. In parallel, the photo-induced hypothesis is tested by estimating whether photoexcited Ph-CH2-CH2-CHI-BPin can undergo single-electron transfer from HPPh2: TD-DFT (CAM-B3LYP/def2-TZVP) yields the excited-state reduction potential of Ph-CH2-CH2-CHI-BPin, ωB97X-D/def2-TZVP provides the ground-state oxidation potential of HPPh2, and the Rehm\u2013Weller equation predicts the free energy change for SET. A negative $\Delta G$ with magnitude >0.1 eV indicates a thermodynamically favorable SET to generate a \u03b1-boryl radical and HPPh2 radical cation. This dual-pathway evaluation preserves the structural mechanistic rigor of the transition state analysis and the photochemical feasibility check of the SET route, enabling discrimination between a ground-state concerted and a photoexcited radical mechanism."
      },
      {
        "strategy_name": "Unified Ground-State and Radical Pathway Screening for I-Displacement in \u03b1-Boryl Iodides",
        "reasoning": "To preserve the advantages of both hypotheses, the computational plan should evaluate in parallel the plausibility of a closed-shell concerted displacement and an open-shell halogen atom transfer, using the same high-level theoretical framework (ωB97X-D/def2-TZVP with SMD solvation in EtOAc or CPME). For the closed-shell scenario, locate and characterise a single transition state connecting Ph-CH2-CH2-CHI-BPin and HPPh2 to Ph-CH2-CH2-

```
CH2-BPin and IPPh2 through a ground-state pathway. Perform IRC to confirm
connectivity and compute ΔG‡; a barrier significantly above ~ 25 kcal/mol would
 argue against this mechanism under mild photochemical conditions. For the
radical case, generate the P-centred radical by H abstraction from HPPh2,
locate the direct I atom transfer transition state to produce IPPh2 plus an \
u03b1-boryl radical, and compute both the barrier and reaction free energy. A
low (<10 kcal/mol) barrier and strongly exergonic profile consistent with P\
u2013I bond strengths (~ 55\u201360 kcal/mol) would support this route. By
combining these models, one can directly compare the kinetic and thermodynamic
profiles of closed-shell versus radical displacement under identical
computational conditions, thereby identifying whether concerted substitution is
 feasible or whether radical XAT dominates in the experimental regime."
      }
    ]
  }
]
},
{
  "query": "Cross-disciplinary methodologies integrating quantum chemical
simulations, kinetic modeling, and electronic structure analysis to evaluate
thermodynamic driving forces and kinetic accessibility of heteroatom\
u2013halogen bond formation in organophosphorus and organoboron systems",
  "optimized_hypotheses": [
    {
      "strategy_name": "Photoinduced Radical C\u2013I Bond Homolysis Followed by
H-Atom Transfer from HPPh2",
      "reasoning": "Test the hypothesis that 450 nm light promotes homolytic
cleavage of the C\u2013I bond in Ph-CH2-CH2-CHI-BPin, generating an \u03b1-
boryl-stabilized alkyl radical, which then abstracts a hydrogen atom from HPPh2
 to yield CH3\u2013CH2\u2013BPin and a P-centered radical. The P-centered
radical could then recombine with iodine to form IPPh2. Computationally, use TD
-DFT (e.g., CAM-B3LYP/def2-TZVP with SMD solvation for EtOAc/CPME) to compute
vertical excitation energies and oscillator strengths for the C\u2013I
chromophore, and unrestricted DFT (UB3LYP-D3/def2-TZVP) to calculate the C\
```

```
u2013I bond dissociation free energy. Compare the photochemical excitation
energy with the photon energy at 450 nm (~ 2.75 eV) and assess whether the BDE
is low enough for photolysis to be feasible.",
    "complements": [
      {
        "strategy_name": "Integrated Photochemical and Thermal Pathway
Assessment for \u03b1\u2011Boryl Iodides",
        "reasoning": "We propose a unified computational strategy that first
quantifies the likelihood of initial photoexcitation leading to C\u2013I bond
homolysis, and then evaluates whether any competing concerted nucleophilic
displacement pathway is viable. For the photochemical scenario, TD\u2011DFT (
CAM\u2011B3LYP/def2\u2011TZVP with SMD solvation for EtOAc/CPME) will be used
to compute vertical excitation energies and oscillator strengths for the C\
u2013I chromophore of Ph-CH2-CH2-CHI-BPin. The excitation energies will be
compared to the photon energy at 450 nm (~ 2.75 eV) to assess spectral overlap,
 while unrestricted DFT (UB3LYP\u2011D3/def2\u2011TZVP) will yield the C\u2013I
 bond dissociation free energy in solution. A low BDE relative to the available
 photoenergy would support a homolytic mechanism leading to \u03b1\u2011boryl
radicals, H\u2011abstraction from HPPh2, and subsequent P\u2011radical trapping
 by iodine to form IPPh2. In parallel, the possible concerted SN2\u2011like
pathway will be tested by ωB97X\u2011D/6\u2011311+G(d,p) with SMD solvent:
separate optimization of reactants, products, and a single transition state
will be followed by an IRC calculation to ensure connectivity. The computed ΔG‡
 will be compared to typical accessible thermal barriers (<25 kcal/mol) under
mild photochemical conditions. This integrated workflow preserves the ability
to identify a low\u2011energy photolytic homolysis route while definitively
ruling out or supporting a concerted substitution if energetically competitive,
 thereby combining the mechanistic coverage of both original hypotheses."
      },
      {
        "strategy_name": "Photochemical & Radical-XAT Synergistic Pathway
Analysis",
        "reasoning": "To unify the two hypotheses while leveraging their
strengths, the computational strategy will evaluate both the feasibility of
```

direct C\u2013I homolysis upon 450 nm photoexcitation and the thermodynamics/
kinetics of an alternative iodine atom transfer (XAT) from a P-centered radical
. First, TD-DFT at the CAM-B3LYP/def2-TZVP level with SMD solvation for EtOAc/
CPME will be used to calculate vertical excitation energies and oscillator
strengths for Ph-CH2-CH2-CHI-BPin, focusing on C\u2013I \u03c3\u2192\u03c3* or
n\u2192\u03c3* transitions, and compare the lowest relevant excitation to the
photon energy of 2.75 eV. In parallel, UB3LYP-D3/def2-TZVP computations will
yield the C\u2013I bond dissociation free energy, assessing whether the excited
 state energy can drive homolysis directly. Second, the role of HPPh2 as a
possible radical source will be explored by computing the free energy change
for iodine abstraction by the P-centered radical from Ph-CH2-CH2-CHI-BPin, as
well as the associated barrier via transition state searches. A low BDE-to-
excitation energy gap supports pathway 1, while a strongly exergonic, low-
barrier XAT supports pathway 2. Comparing both in a unified framework allows
identification of the dominant initiation route under 450 nm irradiation, while
 preserving the photochemical trigger concept of hypothesis 1 and the radical-
chain propagation efficiency of hypothesis 2."
      },
      {
        "strategy_name": "Integrated Photoinitiation Pathway Analysis for \
u03b1\u2011Boryl Radical Formation",
        "reasoning": "The two hypotheses can be merged into a unified
mechanistic/computational strategy that preserves the strengths of both. The
core idea is to test all plausible photolytic initiation channels available
under 450 nm irradiation, then follow the radical chemistry to establish
feasibility. Step 1: Use TD\u2011DFT at CAM\u2011B3LYP/def2\u2011TZVP with SMD
solvation (EtOAc/CPME) to compute vertical excitation energies and oscillator
strengths for the C\u2013I chromophore in Ph-CH2-CH2-CHI-BPin, HPPh2, and the
solvent molecules. Comparing these with the 2.75 eV photon energy will assess
whether any species can be directly excited at 450 nm. Step 2: For species with
 significant absorption, use UB3LYP\u2011D3/def2\u2011TZVP to calculate
homolytic bond dissociation free energies (C\u2013I, P\u2013H, C\u2013H) to
judge if photoexcitation could lead to radical formation. Step 3: Model both
primary photolysis and radical transfer pathways: (a) direct C\u2013I homolysis producing the \u03b1\u2011boryl radical, followed by H\u2011abstraction from HPPh2 and iodine trapping of the P\u2011centered radical; (b) photoinitiated P\u2013H or solvent C\u2013H homolysis producing initial radicals that abstract iodine from Ph-CH2-CH2-CHI-BPin to generate the \u03b1\u2011boryl radical. Step 4: Use DFT to calculate key propagation barrier heights (targeting <10 kcal/mol) for hydrogen atom transfer, iodine atom transfer, and radical recombination, ensuring that either initiation path can sustain a chain reaction. This integrated protocol identifies the kinetically and electronically most accessible initiation mode under the experimental wavelength, while quantitatively verifying the efficiency of the radical propagation sequence."
      },
      {
        "strategy_name": "Integrated Photochemical\u2013Base-Assisted SET Mechanistic Probe",
        "reasoning": "To retain the advantages of both hypotheses, computationally assess two potentially convergent initiation pathways leading to the \u03b1-boryl radical. Step 1: Use TD-DFT (CAM-B3LYP/def2-TZVP, SMD for EtOAc/CPME) to compute vertical excitations and oscillator strengths for Ph-CH2-CH2-CHI-BPin, focusing on C\u2013I antibonding transitions. Compare these to the 450 nm photon energy (~ 2.75 eV) and, with UB3LYP-D3/def2-TZVP thermochemistry, determine the Gibbs free energy for C\u2013I bond homolysis to judge whether direct photoexcitation is likely to promote bond cleavage. Step 2: In parallel, examine a base-promoted SET route. Use ωB97X-D/def2-TZVP (SMD) to calculate the reduction potential of Ph-CH2-CH2-CHI-BPin, and the oxidation potential of the HPPh2 conjugate base (formed in the presence of Cs2CO3). If the redox gap permits thermodynamically plausible SET, model the Ph-CH2-CH2-CHI-BPin radical anion fragmentation (\u03b1-boryl radical + I\u2013) to evaluate the barrier and reaction free energy. This combined approach allows identification of conditions under which either 450 nm light or base-assisted SET\u2014or both in concert\u2014can generate the same key radical intermediate. Such synergy can explain experimental reactivity under photoredox\u2013base dual-initiation conditions."
      }

```
      ]
    }
  ]
},
{
  "query": "Advances in simulating photochemical reaction mechanisms using
  nonadiabatic dynamics, time-dependent DFT, and multireference methods, with
  applications to energy transfer, hydrogen atom transfer, and halogen atom
  transfer processes in catalysis",
  "optimized_hypotheses": [
    {
      "strategy_name": "Direct Photoinduced C\u2013I Homolysis Followed by
  Radical Recombination",
      "reasoning": "Model the absorption spectrum of Ph-CH2-CH2-CHI-BPin using TD
  -DFT (e.g., ωB97X-D/def2-TZVP with SMD solvation for EtOAc/CPME) to check
  whether 450 nm excitation is energetically accessible. Compute the C\u2013I
  bond dissociation energy (BDE) in the ground state and in the lowest singlet/
  triplet excited states to assess feasibility of photoinduced homolysis. Follow
  with unrestricted DFT (UB3LYP-D3/def2-TZVP) optimization of the \u03b1-boryl
  radical and IPPh2 radical, and evaluate the thermodynamics of radical
  recombination to form CH3\u2013CH2\u2013BPin and IPPh2. This will test whether
  a radical chain mechanism initiated by light is plausible.",
      "complements": [
        {
          "strategy_name": "Integrated TD-DFT Photochemical Initiation and DFT
  Radical Chain Analysis",
          "reasoning": "The two hypotheses can be merged into a single
  computational workflow that leverages the spectral/energetic insight from TD-
  DFT to assess photoinitiation feasibility, and the detailed radical energetics
  and XAT barriers from unrestricted DFT to evaluate chain propagation. First, TD
  -DFT (ωB97X-D/def2-TZVP with SMD solvation for EtOAc/CPME) can be used to model
   the absorption spectrum of Ph-CH2-CH2-CHI-BPin and verify whether 450 nm
  photons can access low-lying singlet or triplet states. Within the same TD-DFT
  framework, excited-state C\u2013I BDEs can be computed to evaluate the
```

thermodynamic driving force for photoinduced Ph\u2013CH2\u2013CH2\u2013CH\u2022
\u2013BPin + I\u2022 formation. Ground-state C\u2013I BDEs provide a reference.
Unrestricted DFT (UB3LYP-D3/def2-TZVP) geometry optimizations of the \u03b1-
boryl and IPPh2 radicals, followed by thermodynamic analysis of radical
recombination to Ph\u2013CH2\u2013CH2\u2013CH2\u2022\u2013BPin + IPPh2,
establish whether these intermediates are energetically competent for chain
propagation. In parallel, UB3LYP-D3/def2-TZVP calculations of the P\u2013H BDE
in HPPh2 and the thermochemistry of I\u2022 capture (formation of IPPh2)
clarify a secondary initiation or propagation route via phosphorus-centered
radicals. Direct comparison of \u03b1-boryl radical stabilization energies
against simple alkyl radicals quantifies captodative effects. Finally, the
UB3LYP-D3/def2-TZVP barrier for iodine atom transfer from Ph-CH2-CH2-CHI-BPin
to \u2022PPh2 completes the kinetic picture for XAT. This combined approach
preserves the spectral/photochemical validation from Hypothesis 1 and the
comprehensive radical thermodynamic/kinetic mapping from Hypothesis 2, yielding
an internally consistent model to assess the plausibility of a light-initiated
radical chain mechanism involving XAT steps."
},
{
"strategy_name": "Integrated Photochemical\u2013Ground-State
Mechanistic Evaluation for Ph-CH2-CH2-CHI-BPin Reactivity",
"reasoning": "To preserve the advantages of both hypotheses, the
computational workflow will proceed in two complementary stages. In stage one,
time-dependent DFT (ωB97X-D/def2-TZVP with SMD solvation for EtOAc/CPME) will
model the absorption profile of Ph-CH2-CH2-CHI-BPin to determine whether
experimental 450 nm excitation lies within an allowed electronic transition.
Ground- and excited-state C\u2013I bond dissociation energies will be computed
to gauge the energetic feasibility of direct photoinduced homolysis. Subsequent
unrestricted DFT (UB3LYP-D3/def2-TZVP) optimizations of the \u03b1-boryl and
IPPh2 radicals, plus recombination energetics to CH3\u2013CH2\u2013BPin and
IPPh2, will assess thermodynamic viability of a radical chain pathway. In stage
two, a ground-state pathway will be examined: the SN2-like transition state
for HPPh2 attack at the \u03b1-carbon of Ph-CH2-CH2-CHI-BPin will be optimized
at B3LYP-D3/def2-TZVP with SMD solvent; frequency analysis will confirm a single imaginary mode connecting reactants and products. Barrier heights will
be compared against the energy available from 450 nm photons to assess
competitiveness with the radical route. Substitution of BPin with non-
stabilizing groups will quantify \u03b1-BPin\u2019s role in anionic charge
stabilization and barrier lowering. By integrating excited-state
thermochemistry with ground-state transition state energetics, this unified
approach reveals whether the observed reactivity is more consistent with
photochemically initiated radical chain processes, a concerted SN2-like
substitution, or a competitive interplay between both."
      },
      {
        "strategy_name": "Integrated Photoexcitation\u2013SET Mechanistic
Modelling for Ph-CH2-CH2-CHI-BPin Activation",
        "reasoning": "To preserve the advantages of both hypotheses, the
computational workflow should begin with TD-DFT (ωB97X-D/def2-TZVP, SMD for
EtOAc/CPME) to model the UV\u2013Vis spectrum of Ph-CH2-CH2-CHI-BPin and verify
 whether absorption near 450 nm is feasible. From the lowest relevant excited
singlet and triplet states, compute C\u2013I bond dissociation energies to
assess the propensity for direct photoinduced homolysis. In parallel, model the
 Cs2CO3-mediated deprotonation of HPPh2 to PPh2\u2013 using B3LYP-D3/def2-TZVP,
 SMD solvation, and obtain its vertical ionization potential and electron
affinity. Combine these with similar computed redox parameters for Ph-CH2-CH2-
CHI-BPin to evaluate the thermodynamics of a single-electron transfer (SET)
step using Marcus theory. If SET is favorable, optimize the radical anion of Ph
-CH2-CH2-CHI-BPin and determine the C\u2013I cleavage barrier. Finally, for the
 photochemical and SET-generated radicals, use unrestricted DFT (UB3LYP-D3/def2
-TZVP) to optimize the \u03b1-boryl radical and IPPh2 radical and calculate ΔG
for recombination to CH3\u2013CH2\u2013BPin + IPPh2. This unified approach
tests both direct photolysis and base-promoted radical generation routes,
allowing comparison of activation thresholds, radical stability, and chain-
propagation energetics within a single mechanistic framework."
      },
      {
        "strategy_name": "Integrated Excited-State Radical Pathway Analysis",

```
        "reasoning": "To preserve the advantages of both hypotheses, the
workflow should first use TD-DFT (ωB97X-D/def2-TZVP with SMD for EtOAc/CPME) to
 compute the vertical excitation spectrum of Ph-CH2-CH2-CHI-BPin, confirming
whether 450 nm is energetically accessible. From these stationary points,
evaluate C\u2013I bond dissociation energies in the ground and relevant singlet
/triplet excited states to test feasibility of direct photoinduced homolysis.
Optimize the key radical species (\u03b1-boryl radical, PPh2 radical) with
unrestricted DFT (UB3LYP-D3/def2-TZVP), and quantify the thermodynamics of
their recombination to CH3\u2013CH2\u2013BPin and IPPh2 to assess viability of
a chain mechanism. In parallel, model the encounter complex of photoexcited Ph-
CH2-CH2-CHI-BPin with HPPh2 using TD-DFT (CAM-B3LYP/def2-TZVP, SMD) with
optimization of the lowest singlet and triplet surfaces. Perform a coordinate
scan for a concerted HAT from HPPh2 to the \u03b1-carbon radical site and IAT
from Ph-CH2-CH2-CHI-BPin to PPh2. Locate any low-energy singlet-triplet
crossings or conical intersections that facilitate such a concerted event. This
 combined protocol allows assessment of (a) sequential radical formation and
propagation energetics, and (b) the possibility of light-enabled synchronous
HAT/IAT via an excited-state surface, thereby covering both radical chain and
concerted photochemical pathways under unified computational settings."
      }
    ]
  }
]
},
{
  "query": "Machine learning and data-driven approaches for predicting reactivity
   trends, radical stabilization effects, and substituent electronic influences
  in photoredox and radical-mediated bond-forming reactions, including feature
  engineering from quantum chemical descriptors and reaction databases",
  "optimized_hypotheses": [
    {
      "strategy_name": "Radical C\u2013I Homolysis Followed by Hydrogen Atom
  Transfer (HAT) from HPPh2",
      "reasoning": "Test the hypothesis that blue-light irradiation promotes
```

homolytic cleavage of the C\u2013I bond in Ph-CH2-CH2-CHI-BPin to generate an \
u03b1-boryl alkyl radical, which then abstracts a hydrogen atom from HPPh2,
forming Ph\u2013CH2\u2013CH2\u2013CH2\u2013BPin and a P-centered radical that
captures iodine. Computationally, determine the vertical excitation energies (
TD-DFT, e.g. M06-2X/def2-TZVP/SMD) to assess whether 450 nm light can promote a
 dissociative excited state. Then compute C\u2013I bond dissociation enthalpy
in the ground state and in the lowest singlet/triplet excited states. Evaluate
the thermodynamics of P\u2013H bond cleavage and P\u2013I bond formation. A low
 BDE for C\u2013I and favorable energetics for P\u2013I formation would support
 this radical photochemical pathway’s feasibility.",
    "complements": [
      {
        "strategy_name": "Dual-Pathway Photochemical/Ionic Mechanistic
Assessment",
        "reasoning": "To preserve the mechanistic breadth and computational
rigor of both original hypotheses while exploiting their complementary
strengths, we propose a unified computational workflow that evaluates radical
and ionic pathways in parallel for Ph-CH2-CH2-CHI-BPin + HPPh2 in mixed solvent
. First, probe the photochemical pathway by computing TD-DFT (M06-2X/def2-TZVP/
SMD) vertical excitation energies to test whether 450 nm irradiation can access
 a dissociative excited state along the C\u2013I coordinate. Then quantify C\
u2013I bond dissociation enthalpies (ground and excited singlet/triplet) to
assess homolysis feasibility, and evaluate the thermodynamics of the subsequent
 H-atom transfer from HPPh2 and P\u2013I bond formation. Second, in the same
solvent model (EtOAc/CPME), examine the base-assisted ionic route by modelling
deprotonated HPPh2 (Ph2P\u2013) and computing the SN2 transition state for
displacement of iodide from Ph-CH2-CH2-CHI-BPin; confirm via IRC and determine
the ΔG‡. By comparing excited-state dissociation energetics and radical
thermodynamics against the calculated SN2 activation barrier and known
threshold ranges, this integrated approach can identify the dominant or
coexisting pathway under experimental conditions, while retaining the spectral/
energetic mapping of the photochemical hypothesis and the barrier analysis/
specificity of the ionic hypothesis."
      },

{
"strategy_name": "Integrated Blue-Light Initiated \u03b1\u2011Boryl Radical and P\u2011Centered Radical Iodine Abstraction Mechanism",
"reasoning": "The complementary advantages of the two hypotheses can be leveraged into a unified mechanistic proposal: blue-light excitation of Ph-CH2-CH2-CHI-BPin is evaluated via TD\u2011DFT (M06\u20112X/def2\u2011TZVP/SMD) to determine if the lowest singlet/triplet excited states accessible at ~ 450 nm are dissociative along the C\u2013I bond, thereby producing an \u03b1\u2011boryl radical. In parallel, homolytic P\u2013H cleavage energetics in HPPh2 are determined to assess the feasibility of generating P\u2011centered radicals photochemically or via H\u2011atom abstraction by \u03b1\u2011boryl radicals. Once formed, Ph2P\u2022 radicals can undergo iodine atom transfer (XAT) with Ph-CH2-CH2-CHI-BPin, a step quantified using open\u2011shell DFT to compute both enthalpy change and activation barrier, with spin density analysis confirming radical character. This merged pathway accommodates initiation by direct photoexcitation of Ph-CH2-CH2-CHI-BPin, propagation by P\u2011centered radical XAT, and termination via H\u2011atom recombination with \u03b1\u2011boryl radicals. Ground- and excited\u2011state C\u2013I bond dissociation enthalpies, thermodynamics of P\u2013H and P\u2013I bond formation, and energetics of XAT together provide a comprehensive computational framework to assess both initiation and propagation feasibility, covering photochemical activation, radical stability, and chain propagation efficiency."
},
{
"strategy_name": "Integrated Photochemical/SET Mechanistic Evaluation for \u03b1\u2011Boryl Radical Formation",
"reasoning": "The two hypotheses can be synergistically combined into a unified computational strategy that evaluates both direct photodissociation and photoinduced single\u2011electron transfer (SET) as competing or complementary initiation modes for \u03b1\u2011boryl radical generation from Ph-CH2-CH2-CHI-BPin under blue light. First, use TD\u2011DFT (M06\u20112X/def2\u2011TZVP/SMD) to compute vertical excitation energies of Ph-CH2-CH2-CHI-BPin, HPPh2, and any base\u2013HPPh2 adducts, to determine if 450 nm photons can access dissociative excited states or electronically promote SET donors. From these excited\u2011state surfaces, evaluate C\u2013I bond dissociation
enthalpies for the ground state and lowest excited singlet/triplet states to
quantify whether homolytic cleavage is energetically accessible. In parallel,
calculate adiabatic/vertical ionization potentials of HPPh2 and its base
complexes, as well as adiabatic electron affinities (reduction potentials) for
Ph-CH2-CH2-CHI-BPin in the same solvent model, to determine the driving force
for photo\u2011activated SET. Map thermodynamics for key bond transformations (
P\u2013H cleavage, radical recombination, P\u2013I formation) to assess whether
 downstream radical steps are exergonic. This integrated protocol preserves the
 direct photolysis pathway\u2019s advantage of bypassing electron transfer
while also retaining the SET pathway\u2019s potential for lower\u2011energy
initiation via donor\u2013acceptor excitation. Comparing the computed energetic
 feasibility of both channels under identical photonic input will reveal
whether blue light activates one dominant pathway, allows both to coexist, or
enables sequential interplay of SET and homolytic cleavage."
      },
      {
        "strategy_name": "Integrated Photochemical\u2013Concerted Mechanistic
Analysis",
        "reasoning": "To preserve the advantages of both hypotheses, employ a
dual\u2011track computational protocol: (1) Use TD\u2011DFT (M06\u20112X/def2\
u2011TZVP/SMD) on Ph-CH2-CH2-CHI-BPin to determine the vertical excitation
energies and oscillator strengths, identifying whether 450 nm (\u22482.76 eV)
irradiation can populate a dissociative singlet or triplet state along the C\
u2013I coordinate. Compute C\u2013I bond dissociation enthalpies in the ground
state and the first excited singlet/triplet to assess propensity for \u03b1\
u2011boryl radical generation. Evaluate the thermochemistry of H\u2011atom
abstraction from HPPh2 and subsequent P\u2013I bond formation to judge the
radical pathway\u2019s feasibility. (2) In parallel, locate and characterise
via frequency analysis and intrinsic reaction coordinate (IRC) a cyclic four\
u2011centered transition state for concerted exchange of P\u2013H and C\u2013I
bonds in the same solvent model. Determine the activation free energy and
reaction free energy, and compare the barrier to photochemically available
energy. This combined approach allows assessment of both homolytic radical and

non\u2011radical metathesis mechanisms under photochemical conditions, enabling discrimination between paths or revealing potential mechanistic crossover, thus preserving the predictive value and diagnostic criteria from both original hypotheses."
        }
      ]
    }
  ]
},
{
  "query": "Refined computational evaluation of EDA-mediated versus cesium-assisted photoinduced radical initiation mechanisms in beta-boryl alkyl iodide transformations under visible-light irradiation, including SET feasibility, excited-state energetics, radical propagation, and ion-pair/aggregation effects ",
  "optimized_hypotheses": [
      {
          "strategy_name": "EDA/SRN1-Type Radical Chain Mechanism under Visible-Light Irradiation",
          "reasoning": "Hypothesis: The observed transformation of CH3-CHI-(B-)Pin-PPh2 into Ph-CH2-CH2-CH2-BPin and Ph2P-PPh2 under 450 nm irradiation arises not from direct high-energy excitation of the substrate (vertical excitation ~ 102.5 kcal/mol, far above photon energy at 450 nm ~ 63.5 kcal/mol), but through formation of an electron donor-acceptor (EDA) complex between Ph2P- (generated by Cs2CO3 deprotonation of HPPh2) and the C-I bond of the substrate. Upon charge-transfer excitation (~ 450 nm), a single-electron transfer (SET) occurs into the C-I $\sigma*$ orbital, inducing bond cleavage to generate a $\beta$-boryl carbon radical, iodide, and a phosphorus-centered radical. The carbon radical abstracts hydrogen from HPPh2 (HAT), yielding Ph-CH2-CH2-CH2-BPin and regenerating Ph2P·. Concurrently, Ph2P· radicals dimerize to form Ph2P-PPh2. The mechanism rationalizes the need for 2 equivalents of HPPh2 (one portion as deprotonated donor, another as hydrogen source), the observed P-P coupling product, and the feasibility of initiation under visible light despite insufficient photon energy for direct homolysis."

```
      }
    ]
  }
]
```

***Reflection-revised mechanistic hypotheses for the photochemical radical reaction validation***

```
[
  {
    "strategy_name": "Photoexcited phosphide (PPh2-*) as a short-lived
    photoreductant enabling gated inner-sphere C-I activation without a stable EDA
    complex",
    "reasoning": "Mechanistic core: Cs2CO3 generates PPh2-, which is the only
    component with appreciable 450 nm absorption (tail). Upon irradiation, an
    excited-state manifold of PPh2- (or a tightly associated encounter complex with
     the substrate) acts as a transient, strongly reducing species that triggers C-
    I activation only while light is on, producing a geminate/solvent-caged radical
     pair that is rapidly funneled to products (no long chain). This avoids
    mandatory formation of a spectroscopically resolvable ground-state EDA complex:
     the observed bleaching/non-additivity can arise from strong contact pairing or
     rapid electronic coupling that shortens excited-state lifetime.\nComputational
     tests: (i) Compute UV-vis (TD-DFT with range-separated functional, e.g., ωB97X
    -D or CAM-B3LYP, SMD(EtOAc/CPME model)) for PPh2-, substrate, and contact-ion-
    paired [PPh2-···substrate] encounter structures; verify that the encounter
    complex perturbs/bleaches PPh2- transitions without introducing a distinct new
    low-energy CT band. (ii) Use constrained DFT (CDFT) / δSCF or diabatic ET
    analysis on encounter geometries to estimate driving force and reorganization
    for ET from PPh2-* (via excited-state redox estimates) to σ*(C-I) / substrate
    LUMO. (iii) Explore a photogated cleavage surface: optimize the lowest doublet
    surface corresponding to {substrate· + I·} in a cage with PPh2cdot (or I-
    depending on ET direction), and quantify whether recombination/escape barriers
    are small enough that the radical is short-lived. (iv) Model TEMPO quenching by
     calculating barrierless trapping of the carbon-radical (or P-centered radical)
     relative to productive cage recombination, consistent with complete
```

suppression."
},
{
"strategy_name": "Halogen-bonded exciplex pathway: PPh2-···I-C halogen bond forms a photoactive exciplex that undergoes ultrafast dissociative electron transfer (DET) and cage-controlled H delivery",
"reasoning": "Mechanistic core: Ground-state association is dominated by halogen bonding/anion-σ-hole interaction between PPh2- and the iodine-bearing carbon (or iodine itself), producing strong perturbation of PPh2- absorption (bleaching) without necessarily generating a new CT band. Upon 450 nm excitation of the anion (or the associated pair), the system accesses an exciplex state that is dissociative along C-I, i.e., a DET state that cleanly explains strict light dependence (each cleavage event requires excitation) and lack of chain behavior (products formed via in-cage steps). The carbon radical formed in the solvent cage is then quenched by the nearest available H-source via a very fast, local H transfer event; phosphorus-centered radicals dimerize after cage escape.\nComputational tests: (i) Map halogen-bonded precomplexes with explicit counterion models (Li+ for spectroscopy mimic and Cs+ contact-pairing variants) and mixed implicit/explicit solvation; compute binding energies and geometries (I···P distance, C-I elongation). (ii) Compute excited states of the halogen-bonded pair (TD-DFT; validate with EOM-CCSD on reduced models if feasible) to identify low-lying exciplex/DET character and whether its oscillator strength is consistent with an absorption tail rather than a sharp new band. (iii) Perform relaxed scans along C-I on the relevant excited-state (or on the corresponding charge-transfer diabatic surface) to test if cleavage is downhill/low-barrier. (iv) Evaluate cage fates: compare kinetics (via barrier estimates) for (a) back-electron transfer (BET) vs (b) productive formation of carbon-H and P-P coupling products, rationalizing why continuous irradiation is needed."
},
{
"strategy_name": "Proton-coupled electron transfer (PCET) triggered by photoexcited phosphide: concerted ET/H transfer to generate carbon radical only under irradiation, followed by fast radical recombination/termination",

"reasoning": "Mechanistic core: Instead of invoking classical HAT from neutral HPPh2 as a chain-propagating step, consider a photogated PCET where excited PPh2-* (or an excited encounter complex) transfers an electron into the substrate C-I antibonding manifold while a proton is relayed from HPPh2 (or carbonate-derived bicarbonate) in the same elementary event. This can generate the reduced C-center (radical or anion-like) and simultaneously avoid free long-lived radicals, fitting (i) need for base (to access PPh2-), (ii) TEMPO sensitivity (radical character exists transiently), and (iii) strict light dependence (PCET step requires excitation). The UV-vis non-additivity is consistent with strong ground-state coupling that enables PCET but does not require a stable EDA band.\nComputational tests: (i) Build tri-molecular encounter complexes: substrate + PPh2- + HPPh2 (and variants with Cs+ / CO3ˆ2-/ HCO3-) and locate low-energy H-bond/proton relay geometries. (ii) Use multistate PCET modeling: compute diabatic states for (D·/A·) vs (D/A) and estimate PCET barriers with CDFT or fragment-based approaches; compare stepwise ET vs concerted PCET free-energy profiles. (iii) Verify that thermal PCET barriers are high (dark off), but excited-state PCET is feasible using excited-state energetics (Rehm-Weller-type estimates combined with computed vertical excitation energies). (iv) Quantify why radical chains are disfavored: compute propagation vs termination barriers, showing that productive pathway is dominated by cage PCET/termination rather than chain steps."
},
{
"strategy_name": "Photoinduced spin-orbit-assisted dissociation in a contact ion pair: heavy-atom iodine enables intersystem crossing to a reactive triplet radical-pair surface only when PPh2- is photoexcited",
"reasoning": "Mechanistic core: The iodine substituent can enable strong spin-orbit coupling. In a contact ion pair (PPh2- associated to the substrate), excitation localized primarily on PPh2- (consistent with its absorption tail) may undergo rapid intersystem crossing mediated by iodine to a triplet (or open-shell singlet) radical-pair surface that is dissociative along C-I. Because population of this surface depends on continuous photoexcitation, the reaction shuts off in the dark. The resulting radical pair remains spin-correlated/caged, suppressing chain propagation; TEMPO intercepts the radical pair before

```
recombination.\nComputational tests: (i) Optimize ground-state contact ion pair
 geometries including explicit cation (Li+ for spectral mimic and Cs+ realistic
) and compute low-lying singlet/doublet/triplet states. (ii) Compute spin-orbit
 coupling (SOC) between relevant states (e.g., using ORCA/PySCF interfaces if
needed; Gaussian can be supplemented by external SOC-capable packages) on
representative geometries to evaluate whether iodine can accelerate ISC. (iii)
Map minimum-energy crossing points (MECPs) between singlet/doublet and triplet
surfaces along the C-I coordinate to test accessibility upon excitation. (iv)
Compare predicted isotope/solvent sensitivity of ISC-controlled rates
computationally (reorganization, dielectric effects) without proposing new
experiments; use sensitivity analysis to distinguish from pure ET models."
},
{
 "strategy_name": "Boryl-assisted radical-polar crossover in a photogated cage:
transient α/β-boryl-stabilized radical forms only under light, then undergoes
ultrafast anion capture/back-electron transfer to yield net hydrodeiodination
without chain propagation",
 "reasoning": "Mechanistic core: The boron substituent can stabilize adjacent
radical/anion character and enable rapid radical-polar crossover (RPC) pathways
. Under irradiation, a short-lived carbon radical (from photogated activation
by excited PPh2-) is produced in a solvent cage. Rather than propagating via
HAT chain steps, it is rapidly converted to a closed-shell intermediate by back
-electron transfer (BET) and/or capture by iodide/phosphorus species, followed
by fast protonation/hydride-equivalent delivery from HPPh2-derived ion pairs.
Net result: hydrodeiodination with PPh2-PPh2 formation from P-centered radicals
, but the key carbon-centered radical remains transient and controlled. UV-vis
perturbation is consistent with strong ground-state association that promotes
this gated cage chemistry.\nComputational tests: (i) Characterize stabilization
 of the putative carbon radical and its anion/cation counterparts (spin density
 distribution, SOMO character) with DFT and higher-level single points (DLPNO-
CCSD(T) on truncated models). (ii) Compute free-energy profiles for RPC steps:
carbon radical + e- (from PPh2-*/BET) → carbanion, then protonation by HPPh2/
solvent adducts; compare to direct HAT from neutral HPPh2 and show that chain-
propagating HAT is not kinetically dominant. (iii) Model termination: P-
```

```
    centered radicals dimerization barriers vs alternative traps; connect to
    observed PPh2-PPh2 formation. (iv) Evaluate TEMPO interception thermodynamics/
    kinetics for both carbon-radical and P-radical channels, consistent with
    complete inhibition."
  }
]
```

### C.8.3 Hypothesis for the dihedral-based selectivity validation

***Initial mechanistic hypotheses for the dihedral-based selectivity validation***

```
[
    {
        "query": "Computational chemistry strategies for elucidating how steric and
     electronic modulation in bidentate nitrogen ligands influences square-planar
    Ni(II) coordination geometry, ligand bite angle, and ancillary ligand
    positioning, with implications for migratory versus in situ cross-coupling
    selectivity in catalysis",
        "hypotheses": [
            {
                "strategy_name": "Steric Control of Hydride Departure via Ligand-Ni
     Dihedral Angle Scan",
                "reasoning": "Hypothesis: Ortho-methyl substitution in 2,2'-
    bipyridine imposes steric hindrance that forces a non-planar distortion in the
    tetra-coordinate Ni(II) complex, increasing the dihedral angle between the N-Ni
    -Br-H coordination plane and the ligand backbone. This distortion reduces the
    energy penalty associated with hydride departure from the catalyst plane,
    favoring the migratory pathway. Indicator: The relaxed scan of the Ni-H out-of-
    plane dihedral angle energy profile. Computational experiment: Build tetra-
    coordinate Ni(II) complexes (N-N from ligand, Br, H) for ortho-methyl,
    unsubstituted, and meta-/para-methyl bipyridines. Optimize geometries at M06-L/
    def2-TZVP with SMD(DMA) solvation in Gaussian or PySCF. Perform a relaxed
    dihedral scan from 0° to 30° for Ni-H out-of-plane displacement. Compare energy
    rise rates across ligand types. Prediction: Ortho-substituted ligands will show
```

```
 a shallower energy increase, correlating with migratory selectivity."
        },
        {
            "strategy_name": "Ligand Bite Angle Influence on Square-Planar
Stability",
            "reasoning": "Hypothesis: Ortho substitution alters the natural
bite angle of 2,2’-bipyridine, destabilizing the ideal square-planar Ni(II)
geometry and creating a more flexible coordination sphere that facilitates
rearrangement to the migratory intermediate. Indicator: Energy change as a
function of enforced N-Ni-N bite angle variation. Computational experiment:
Optimize tetra-coordinate Ni(II) complex for each ligand. Perform constrained
optimizations varying the N-Ni-N angle ±10° from the equilibrium value in 2°
increments. Use B3LYP-D3BJ/def2-TZVP with SMD(DMA) in Gaussian. Evaluate the
curvature of the energy vs. angle plot. Prediction: Ortho-substituted ligands
will exhibit lower energetic penalties for deviation from equilibrium bite
angles, indicating a more fluxional coordination geometry conducive to
migration."
        },
        {
            "strategy_name": "Electronic Modulation of Ni-H Bond Strength via
NBO Analysis",
            "reasoning": "Hypothesis: Ortho-methyl substitution increases
electron density donation from ligand nitrogen atoms due to inductive and
hyperconjugative effects, weakening the Ni-H bond and lowering the barrier for
hydride departure. Indicator: Ni-H bond dissociation energy (BDE) and Ni-H
Wiberg bond index from NBO analysis. Computational experiment: Optimize tetra-
coordinate Ni(II) complexes for each ligand at M06/def2-TZVP with SMD(DMA) in
Gaussian. Compute single-point energies for the Ni-H homolytic cleavage product
 and perform NBO analysis to obtain bond order. Prediction: Ortho-substituted
ligands will have noticeably lower Ni-H BDE and bond index, correlating with
higher migratory product formation."
        },
        {
            "strategy_name": "Ligand-Induced Br-Ni Bond Lability via Relaxed
```

Bond-Length Scan",
            "reasoning": "Hypothesis: Ortho substitution induces steric repulsion toward the bromide ligand, increasing Br-Ni bond lability and facilitating rearrangement to the migratory product through easier Br dissociation/repositioning. Indicator: Energy profile from relaxed Br-Ni bond stretching. Computational experiment: Optimize tetra-coordinate Ni(II) complex for each ligand at B3LYP-D3BJ/def2-TZVP with SMD(DMA). Perform relaxed scan of Br-Ni distance from equilibrium to +0.3 Å in 0.05 Å steps. Compare energy rise for different ligands. Prediction: Ortho-substituted ligands will show a smaller energy penalty for Br-Ni elongation, consistent with increased Br mobility aiding migration."
        },
        {
            "strategy_name": "Ligand Control of Catalyst Planarity via Out-of-Plane Displacement Energy Map",
            "reasoning": "Hypothesis: Ortho substitution disrupts square-planar planarity by sterically pushing the ligand backbone, making the Ni coordination sphere more pseudo-tetrahedral. This geometry change reduces the steric/electronic barrier for hydride migration. Indicator: Energy contour map for simultaneous out-of-plane displacement of Ni-H and Ni-Br ligands. Computational experiment: Optimize tetra-coordinate Ni(II) complexes at M06-L-D3/def2-TZVP with SMD(DMA). Perform two-dimensional relaxed scans varying Ni-H and Ni-Br out-of-plane angles in increments of 5°, generating an energy surface. Compare curvature and minima shifts for different ligands. Prediction: Ortho-substituted ligands will produce a flatter energy surface, indicating greater flexibility and higher migratory selectivity."
        }
    ]
},
{
    "query": "Quantum chemical modeling approaches for mapping energy profiles and geometric distortion pathways in tetra-coordinate Ni(II) complexes under varying ligand substitution patterns, incorporating relaxed coordinate scans, electronic structure analysis, and comparison of ligand-induced changes in

```
hydride departure energetics",
    "hypotheses": [
        {
            "strategy_name": "Steric Control of Hydride Departure via Out-of-
Plane Distortion Energy Scan",
            "reasoning": "Hypothesis: Ortho-methyl substitution in 2,2’-
bipyridine increases steric hindrance near the Ni(II) coordination plane,
forcing the hydride ligand to adopt a more out-of-plane geometry at rest. This
reduces the energetic penalty for hydride departure during the migratory step,
favoring migratory products. Indicator: The relaxed potential energy profile as
 the Ni-H bond is displaced out of the square-planar plane (defined by N-N-Ni-
Br) in the tetra-coordinate Ni(II)-bpy-Br-H complex. Computational experiment:
Optimize the tetra-coordinate complex for unsubstituted, ortho-, and meta-/para
-methyl bpy ligands at B3LYP-D3(BJ)/def2-TZVP level with CPCM(DMA) solvation in
 Gaussian. Perform a relaxed scan of the hydride dihedral angle relative to the
 Ni coordination plane in 5° increments up to ~ 45°, recording energies.
Evaluation criterion: Ortho-methyl ligands should show a shallower energy rise
with hydride displacement, indicating reduced distortion energy cost and
correlating with higher migratory selectivity."
        },
        {
            "strategy_name": "Ligand-Induced Ni Coordination Geometry Shift
Quantified by τ4 Index Variation",
            "reasoning": "Hypothesis: Ortho-methyl substitution shifts the Ni(
II) coordination geometry away from ideal square-planar toward more tetrahedral
-like arrangement, increasing flexibility for hydride migration. Indicator: τ4
geometry index (0 for perfect square-planar, 1 for perfect tetrahedral)
computed from Ni-ligand bond angles in optimized tetra-coordinate complexes.
Computational experiment: Optimize tetra-coordinate Ni(II)-bpy-Br-H complexes
for each ligand variant at BP86-D3/def2-TZVP with CPCM(DMA), then compute τ4
from bond angles using a Python script (ASE + RDKit). Evaluation criterion:
Ortho-substituted ligand should yield higher τ4 values (greater tetrahedral
distortion), and the trend in τ4 should correlate with migratory product
preference."
```

```
        },
        {
            "strategy_name": "Electronic Modulation of Ni-H Bond Covalency via
Natural Bond Orbital (NBO) Analysis",
            "reasoning": "Hypothesis: Ortho-methyl substitution alters
electronic donation from bpy nitrogens, weakening Ni-H σ-bonding and
facilitating hydride departure. Indicator: NBO-derived Ni-H bond occupancy and
Wiberg bond index (WBI) for the Ni-H bond in the optimized tetra-coordinate
complex. Computational experiment: Optimize complexes for all ligand variants
at B3LYP-D3/def2-TZVP with CPCM(DMA) and run NBO analysis in Gaussian.
Evaluation criterion: Ortho-methyl ligand should yield lower Ni-H bond
occupancy and WBI compared to unsubstituted and meta-/para-modified ligands,
indicating weaker covalency and better migratory propensity."
        },
        {
            "strategy_name": "Bromide-Hydride Trans Influence Measured by Ni-Br
 Bond Elongation under Constrained Hydride Displacement",
            "reasoning": "Hypothesis: Ortho-methyl substitution changes the
spatial arrangement of ligands such that hydride displacement exerts a stronger
 trans influence on the Ni-Br bond, destabilizing the in situ pathway and
favoring migration. Indicator: Change in Ni-Br bond length during constrained
hydride displacement scans. Computational experiment: Optimize starting
structures with each ligand variant at B3LYP-D3/def2-TZVP CPCM(DMA), then
perform constrained Ni-H bond length increases (0.05 Å steps up to +0.3 Å)
while reoptimizing remaining geometry, recording Ni-Br bond lengths. Evaluation
 criterion: Ortho-substituted bpy should show larger Ni-Br elongation per unit
Ni-H displacement, linking to destabilization of in situ product formation."
        },
        {
            "strategy_name": "Steric Energy Penalty for Br-H Ligand Proximity
via Nonbonded Interaction Energy Decomposition",
            "reasoning": "Hypothesis: Ortho-methyl substitution increases
steric repulsion between Br and H ligands in the tetra-coordinate Ni complex,
destabilizing the in situ arrangement and promoting hydride migration.
```

```
Indicator: Counterpoise-corrected nonbonded interaction energy between Br and H
 fragments in the optimized complex. Computational experiment: Partition the
tetra-coordinate complex into Br and H ligand fragments (with Ni-bpy core) and
compute fragment interaction energies using supermolecule/BSSE-corrected DFT (
B3LYP-D3/def2-TZVP CPCM(DMA)) in Gaussian or PySCF. Evaluation criterion: Ortho
-substituted ligand should produce higher Br-H repulsive energy magnitude
compared to other variants, correlating with suppression of in situ product."
        }
    ]
},
{
    "query": "Cross-disciplinary insights from DFT, ligand field theory, and
molecular orbital analysis into how local steric crowding and electronic
perturbations alter Ni-ligand electronic coupling, coordination number
stability, and activation barriers in metal-mediated migratory coupling
reactions",
    "hypotheses": [
        {
            "strategy_name": "Steric Control of Out-of-Plane Hydride Departure
",
            "reasoning": "Hypothesis: Ortho-methyl substitution on 2,2'-
bipyridine increases steric crowding near the Ni coordination plane, forcing
the β-hydride to adopt a geometry further out of plane in the tetra-coordinate
Ni(II)-N-N-Br-H intermediate. This facilitates migratory coupling by lowering
the barrier for hydride departure and chain-walking. Indicator: Perform a
relaxed scan of the dihedral angle between the Ni coordination plane (defined
by N-N-Br) and the Ni-H vector, computing relative energies as the hydride is
displaced out of plane. Prediction: Ortho-substituted ligands will show a
shallower energy increase for out-of-plane displacement than unsubstituted or
meta-/para-substituted ligands. Computational experiment: Optimize tetra-
coordinate Ni(II)-N-N-Br-H complexes for each ligand type at M06/def2-SVP with
SMD(DMA) solvation, then carry out a constrained dihedral scan (0° → 90°) in 1
0° increments. Evaluate the slope and curvature of the energy profile; a flatter
 profile correlates with migratory product selectivity."
```

```
        },
        {
            "strategy_name": "Ligand Bite Angle Modulation of Coordination
Geometry",
            "reasoning": "Hypothesis: Ortho-methyl substitution alters the
effective bite angle of bipyridine, distorting the square-planar Ni(II)
geometry in the tetra-coordinate intermediate and destabilizing geometries that
 favor in situ product formation. Indicator: Measure the N-Ni-N bite angle
distribution and its correlation with Br-Ni-H cis/trans arrangements from
optimized geometries and conformer sampling. Prediction: Ortho-substituted
ligands will show systematically smaller bite angles, leading to increased Br-H
 trans influence and promoting migratory pathways. Computational experiment:
Generate conformers of Ni(II)-N-N-Br-H for each ligand using RDKit+MMFF,
optimize at ωB97X-D/def2-SVP/SMD(DMA), extract N-Ni-N and Br-Ni-H angles.
Compare average bite angles and identify distortions; correlate smaller bite
angles with migratory selectivity."
        },
        {
            "strategy_name": "Electronic Influence on Ni-H Bond Strength",
            "reasoning": "Hypothesis: Ortho-methyl substitution increases
electron density donation from bipyridine to Ni, weakening the Ni-H bond in the
 tetra-coordinate intermediate, which facilitates hydride migration and favors
the migratory product. Indicator: Compute the Ni-H bond dissociation free
energy (BDFE) via single-point energy differences between Ni-H and Ni + H·
fragments. Prediction: Ortho-substituted ligands yield lower BDFE(Ni-H) values
than unsubstituted or meta-/para-substituted ligands. Computational experiment:
 Optimize complexes at M06/def2-TZVP/SMD(DMA), calculate BDFEs with thermal
corrections from frequency analysis. Compare BDFEs across ligands; lower values
 correlate with migratory selectivity."
        },
        {
            "strategy_name": "Trans Influence of Br on Ni-H Elongation",
            "reasoning": "Hypothesis: Ortho-substitution perturbs the ligand
field such that Br and H ligands preferentially adopt a trans configuration
```

with increased Br trans influence, leading to Ni-H bond elongation and faster
migratory coupling. Indicator: Perform a relaxed scan of the Ni-H bond length
while keeping Br trans to H, recording the energy profile. Prediction: Ortho-
substituted ligands will show a shallower energy rise and longer equilibrium Ni
-H bond length compared to others. Computational experiment: Optimize Ni(II)-N-
N-Br-H complexes with Br-H trans arrangement at B3LYP-D3BJ/def2-SVP/SMD(DMA),
scan Ni-H distance from equilibrium to +0.3 Å. Compare equilibrium lengths and
curvature of the scan; longer, more weakly bound Ni-H correlates with migratory
selectivity."
        },
        {
            "strategy_name": "Steric Hindrance-Induced Tilting of the Ni
Coordination Plane",
            "reasoning": "Hypothesis: Ortho-methyl groups force the bipyridine
ligand to twist, tilting the Ni coordination plane relative to the Br-H axis,
which destabilizes intermediates favoring in situ product pathways and lowers
barriers for $\beta$-hydride elimination/migration. Indicator: Quantify the tilt
angle between the N-Ni-N plane and the Br-Ni-H plane from optimized geometries
and track how energy changes as this tilt is varied in a constrained scan.
Prediction: Ortho-substituted ligands will show larger equilibrium tilt angles
and less energetic penalty for increased tilting compared to other substitution
patterns. Computational experiment: Optimize tetra-coordinate Ni(II)-N-N-Br-H
at M06/def2-SVP/SMD(DMA), measure tilt angles, then scan tilt by constraining
one dihedral in ±15° from equilibrium. Compare energy penalties across ligands;
smaller penalties and larger natural tilts correlate with migratory
selectivity."
        }
    ]
},
{
    "query": "Machine learning and data-driven catalyst design frameworks that
integrate computed geometric/electronic descriptors from low-cost quantum
methods with experimental selectivity data to predict ligand substitution
effects on mechanistic branching across transition-metal-catalyzed cross-

```
coupling reactions",
    "hypotheses": [
        {
            "strategy_name": "Steric Modulation of Ni(II) Square-Planar to
Tetra-Coordinate Distortion",
            "reasoning": "Hypothesis: Ortho-methyl substitution on 2,2'-
bipyridine increases steric congestion near the metal center, forcing the tetra
-coordinate Ni(II)-(bpy)-Br-H intermediate to adopt a geometry with greater out
-of-plane displacement of the hydride ligand. This reduces the energetic
penalty for hydride migration, favoring the migratory pathway. Indicator:
Perform a relaxed scan of the Ni-H out-of-plane dihedral angle (relative to the
 N-Ni-N plane) for tetra-coordinate complexes with ortho-, meta-, para-, and
unsubstituted bpy ligands. Compute energy profiles at each dihedral increment.
Predicted trend: Ortho-substitution will flatten the energy profile near out-of
-plane geometries, lowering the barrier to distortion. Validation experiment:
Optimize tetra-coordinate Ni(II)-(bpy)-Br-H complexes in implicit DMA solvent (
SMD or PCM) using DFT (e.g., M06/def2-TZVP for Ni, def2-SVP for other atoms, D3
 dispersion). Perform dihedral scans using Gaussian or Psi4. Compare barrier
heights for 0° → 90° out-of-plane rotation across ligands."
        },
        {
            "strategy_name": "Electronic Influence on Ni-H Bond Strength and
Migratory Aptitude",
            "reasoning": "Hypothesis: Ortho-methyl substitution alters the
electron-donating ability of the bipyridine ligand via hyperconjugation and
inductive effects, modifying Ni-H bond strength and thus the ease of hydride
migration. Indicator: Calculate Ni-H bond dissociation energies (BDEs) for the
tetra-coordinate Ni(II)-(bpy)-Br-H species across ligand variants. Predicted
trend: Ortho-substitution will weaken the Ni-H bond (lower BDE), facilitating
hydride shift and favoring migratory coupling. Validation experiment: Optimize
each tetra-coordinate complex in implicit DMA solvent at B3LYP-D3/def2-TZVP.
Compute single-point energies for Ni-H homolytic cleavage products (Ni-Br-(bpy)
 + H·). Evaluate BDE differences and correlate with selectivity. Software:
Gaussian, ORCA, or PySCF."
```

```
        },
        {
            "strategy_name": "Ligand Bite Angle and Its Effect on Br/H Ligand
Mutual Repulsion",
            "reasoning": "Hypothesis: Ortho substitution increases the
effective bite angle (N-Ni-N) due to steric bulk, altering cis ligand-ligand
distances between Br and H and reducing their mutual repulsion in distorted
geometries, stabilizing migratory intermediates. Indicator: Measure N-Ni-N bite
 angle and Br-H distance in optimized tetra-coordinate complexes. Perform
constrained bite angle scans (±5-15° from equilibrium) to evaluate energy
sensitivity. Predicted trend: Ortho-substituted ligands will have larger bite
angles and lower energy penalties for Br-H proximity changes, enabling smoother
 transition to migration geometry. Validation experiment: Optimize structures
at M06-L/def2-TZVP with PCM(DMA), perform constrained optimizations varying
bite angle, record energy vs. angle curve. Compare curvature of energy profiles
 for different ligands."
        },
        {
            "strategy_name": "Axial Coordination Susceptibility and Off-Cycle
Intermediate Stabilization",
            "reasoning": "Hypothesis: Ortho substitution hinders axial
coordination sites in the square-planar Ni(II) complex, reducing the likelihood
 of off-cycle species that lead to in situ product formation. Indicator:
Compute binding free energies (δG_bind) for a model axial ligand (e.g., DMA or
OH-) to the tetra-coordinate Ni(II)-(bpy)-Br-H complex. Predicted trend: Ortho-
substituted ligands will exhibit significantly less favorable axial ligand
binding, thereby maintaining the reactive tetra-coordinate form and favoring
migration. Validation experiment: Optimize tetra-coordinate and penta-
coordinate (axially bound) forms at ωB97X-D/def2-TZVP with PCM(DMA). Compute δ
G_bind for axial ligand coordination. Compare across ligands to see if ortho
substitution destabilizes penta-coordinate species."
        },
        {
            "strategy_name": "Ligand-Induced Modulation of Ni(II) d-Orbital
```

Splitting and Hydride Orbital Overlap",
"reasoning": "Hypothesis: Ortho substitution changes ligand field strength and geometry, altering Ni(II) d-orbital splitting in a way that enhances overlap between the hydride $\sigma$ orbital and the Ni d$x^2$-$y^2$ or d$z^2$ orbitals during out-of-plane migration. Indicator: Perform natural bond orbital (NBO) or fragment molecular orbital (FMO) analysis on optimized tetra-coordinate Ni(II)-(bpy)-Br-H complexes to quantify hydride→Ni donation and Ni→$\sigma$*(C-H) back-donation. Predicted trend: Ortho-substituted ligands will show increased hydride→Ni donation in distorted geometries, lowering migration barriers. Validation experiment: Optimize complexes at PBE0-D3/def2-TZVP with PCM(DMA), perform NBO analysis for planar and partially distorted geometries (e.g., 45° out-of-plane). Compare donor-acceptor interaction energies across ligand sets."
}
]
},
{
"query": "Comparative mechanistic studies leveraging computational spectroscopy (e.g., simulated NMR, XAS, EPR parameters) and electronic structure diagnostics to correlate ligand substitution patterns with changes in Ni(II) electronic environment, ancillary ligand binding, and reactivity trends in migratory versus in situ product formation",
"hypotheses": [
{
"strategy_name": "Steric Hindrance-Induced Out-of-Plane $\beta$-Hydride Facilitation",
"reasoning": "Hypothesis: Ortho-methyl substitution on 2,2'-bipyridine introduces significant steric crowding near the Ni coordination plane, forcing the $\beta$-hydride ligand to adopt a geometry with increased out-of-plane displacement earlier in the reaction coordinate. This reduces the barrier for the migratory pathway relative to the in situ pathway. Indicator: Perform a relaxed scan of the H-Ni-Br dihedral angle in a tetra-coordinate Ni(II) complex (N-N, Br, H ligands) for ligands with ortho-, meta-, para-methyl, and unsubstituted bpy. Compute the energy profile as the H atom moves out of the square-plane geometry. Prediction: Ortho-substituted ligands will show a shallower energy increase and earlier onset of stabilization during out-of-plane displacement, correlating with exclusive migratory product formation. Computational experiment: Use Gaussian with B3LYP-D3/def2-TZVP and SMD(DMA) solvation; optimize the tetra-coordinate complexes and scan the H-Ni-Br dihedral in 5° increments from planar to ±90°, recording relative energies."
        },
        {
            "strategy_name": "Ligand-Induced Spin-State Modulation Favoring Migration",
            "reasoning": "Hypothesis: Ortho-substitution alters the ligand field strength and Ni(II) geometry, increasing the stability of a high-spin ( triplet) tetrahedral-like intermediate relative to the low-spin square-planar geometry. The high-spin state facilitates $\beta$-hydride migration by reducing electron density in antibonding orbitals. Indicator: Compute singlet-triplet energy gaps for tetra-coordinate Ni(II) complexes with different bpy substitution patterns. Prediction: Ortho-substituted bpy will show a smaller singlet-triplet gap, making high-spin geometries more accessible and favoring migratory pathways. Computational experiment: Optimize singlet and triplet states with Gaussian (B3LYP-D3/def2-TZVP, SMD(DMA)), perform frequency analysis to confirm minima, and compare $\delta$E(S-T) across ligand variants."
        },
        {
            "strategy_name": "Ni-H Bond Covalency Change Due to Steric Compression",
            "reasoning": "Hypothesis: Ortho-methyl substitution compresses the N-Ni-N bite angle and modifies orbital overlap between Ni and the $\beta$-hydride, leading to a weaker Ni-H bond that is more labile toward migration. Indicator: Calculate Mayer bond orders and natural bond orbital (NBO) charges for the Ni-H bond in optimized tetra-coordinate complexes across ligand variants. Perform a constrained Ni-H bond length scan (±0.2 Å from equilibrium) to determine bond lability. Prediction: Ortho-substituted ligands will exhibit consistently lower Ni-H bond orders and shallower bond dissociation energy profiles, correlating with migratory selectivity. Computational experiment: Use Gaussian (B3LYP-D3/def2-TZVP, SMD(DMA)) for optimizations and NBO analysis, scanning Ni-H distance

```
 in 0.02 Å increments."
        },
        {
            "strategy_name": "Bite Angle-Driven Coordination Geometry
Destabilization",
            "reasoning": "Hypothesis: Ortho-substitution reduces the 2,2'-
bipyridine bite angle, distorting the square-planar Ni(II) geometry and
destabilizing the in situ pathway intermediate where β-hydride remains in-plane
. Indicator: Measure N-Ni-N bite angle and total ligand strain energy from
optimized geometries of tetra-coordinate complexes. Perform constrained
optimizations varying the N-Ni-N angle ±5° to assess energy sensitivity.
Prediction: Ortho-substituted ligands will have smaller bite angles and higher
strain energies when forced to maintain an in-plane β-hydride, disfavoring the
in situ pathway. Computational experiment: Gaussian (B3LYP-D3/def2-TZVP, SMD(
DMA)), optimize geometries, extract angles, and perform constrained
optimizations to measure strain energy trends."
        },
        {
            "strategy_name": "Electronic Field Effect on Ni-Br Interaction
Facilitating Migration",
            "reasoning": "Hypothesis: Ortho-methyl substitution enhances
electron density on the bipyridine ligand via hyperconjugation, increasing back
-donation to the β-hydride and weakening the Ni-Br bond, which promotes
migratory displacement. Indicator: Compute Ni-Br Mayer bond order and charge
distribution via NPA for tetra-coordinate complexes with different ligand
substitutions. Perform a relaxed scan of Ni-Br bond length to observe energy
profiles and bond weakening trends. Prediction: Ortho-substituted ligands will
exhibit lower Ni-Br bond orders and reduced bond dissociation energies,
enabling β-hydride migration and suppressing in situ product formation.
Computational experiment: Gaussian (B3LYP-D3/def2-TZVP, SMD(DMA)), optimize
geometries, run NPA/Mayer analysis, and perform Ni-Br bond length scans in 0.02
 Å increments."
        }
    ]
```

}
]

***Refined mechanistic hypotheses for the dihedral-based selectivity validation***

[
{
"query": "Computational chemistry strategies for elucidating how steric and electronic modulation in bidentate nitrogen ligands influences square-planar Ni (II) coordination geometry, ligand bite angle, and ancillary ligand positioning , with implications for migratory versus in situ cross-coupling selectivity in catalysis",
"optimized_hypotheses": [
{
"strategy_name": "Ligand Bite Angle Influence on Square-Planar Stability",
"reasoning": "Hypothesis: Ortho substitution alters the natural bite angle of 2,2'-bipyridine, destabilizing the ideal square-planar Ni(II) geometry and creating a more flexible coordination sphere that facilitates rearrangement to the migratory intermediate. Indicator: Energy change as a function of enforced N-Ni-N bite angle variation. Computational experiment: Optimize tetra-coordinate Ni(II) complex for each ligand. Perform constrained optimizations varying the N-Ni-N angle ±10° from the equilibrium value in 2° increments. Use B3LYP-D3BJ/def2-TZVP with SMD(DMA) in Gaussian. Evaluate the curvature of the energy vs. angle plot. Prediction: Ortho-substituted ligands will exhibit lower energetic penalties for deviation from equilibrium bite angles, indicating a more fluxional coordination geometry conducive to migration.",
"complements": [
{
"strategy_name": "Steric-Modulated Ligand Flexibility Analysis",
"reasoning": "By integrating the steric-distortion and bite-angle hypotheses, the strategy examines how ortho-methyl substitution in 2,2\u2032-bipyridine simultaneously enforces out-of-plane Ni\u2013H displacement and modifies the N\u2013Ni\u2013N bite angle, both weakening square-planar rigidity in tetra-coordinate Ni(II) complexes. The combined computational protocol uses matched ligand sets (ortho-, meta-, para-, unsubstituted) and performs two complementary energy scans: (1) a relaxed Ni\u2013H dihedral scan at M06-L/def2-TZVP/SMD(DMA) to quantify hydride departure penalties, and (2) a constrained bite-angle scan at B3LYP-D3BJ/def2-TZVP/SMD(DMA) to assess coordination sphere flexibility. Correlating the shallower dihedral energy curves with lower curvature in the bite-angle plots will reveal whether steric crowding promotes a joint increase in hydride migratory propensity and angular tolerance, preserving each method\u2019s mechanistic sensitivity while capturing the interplay between steric enforcement and ligand fluxionality."
      },
      {
        "strategy_name": "Steric\u2013Electronic Synergistic Modulation of Ni\u2013H Migratory Selectivity",
        "reasoning": "The ortho-methyl substitution in 2,2\u2032-bipyridine simultaneously induces steric distortion and enhances sigma-electron donation from the nitrogen donors. Sterically, the methyl groups enforce a non-planar ligand\u2013metal geometry that increases the dihedral angle between the N\u2013Ni\u2013Br\u2013H coordination plane and the ligand backbone, reducing the energetic penalty for Ni\u2013H displacement out of the catalyst plane. Electronically, the same substitution boosts electron density on the nitrogen atoms via inductive and hyperconjugative effects, weakening the Ni\u2013H bond as reflected by reduced bond dissociation energy and Wiberg bond index. Together, these steric and electronic effects cooperatively lower the barrier for hydride departure, thereby preferentially facilitating the migratory pathway. The integrated computational protocol involves (1) M06-L/def2-TZVP SMD(DMA) relaxed scans of Ni\u2013H out-of-plane dihedrals to quantify steric distortion energy profiles across substituted ligands, and (2) M06/def2-TZVP SMD(DMA) optimizations followed by single-point Ni\u2013H BDE calculations and NBO bond order analyses to capture electronic weakening. Correlation of shallow dihedral energy rise with low BDE and low bond order for ortho-substituted cases will validate the steric\u2013electronic coupling mechanism."
      },
      {

"strategy_name": "Sterically-Induced Dual Facilitation of Ni\u2013H Migration via Planar Distortion and Br\u2013Ni Lability",
"reasoning": "By combining the hypotheses, the ortho-methyl groups on 2,2\u2032-bipyridine simultaneously enforce a non-planar distortion in the Ni(II) coordination environment and introduce direct steric repulsion to the bromide ligand. The ligand backbone twist reduces the geometric penalty for Ni\u2013H displacement from the catalyst plane, lowering the barrier for hydride migration, while the bromide steric crowding increases Br\u2013Ni bond lability, making halide repositioning or dissociation easier during the migratory rearrangement. These effects are complementary: distortion facilitates hydride departure, and bromide mobility enables structural reorganization to the migratory product. Computationally, this can be tested by performing relaxed dihedral scans of the Ni\u2013H out-of-plane angle and relaxed Br\u2013Ni elongations for ortho-, meta-, para-, and unsubstituted ligands. A shallower energy rise in both scans for the ortho-substituted ligand would confirm the synergy between planar distortion and halide lability in controlling migratory selectivity."
},
{
"strategy_name": "Sterically Induced Pseudo-Tetrahedral Flexibility Mapping",
"reasoning": "Combine the single-coordinate dihedral scan of the Ni\u2013H out-of-plane displacement with a two-dimensional scan involving both Ni\u2013H and Ni\u2013Br out-of-plane motions, using identical structural models and solvation. This unified approach captures the steric influence of ortho-methyl substitution on both local (Ni\u2013H) and global (Ni coordination sphere) flexibilities. The one-dimensional scans provide direct quantification of the dihedral energy penalties, while the two-dimensional energy maps reveal cooperative distortions and the extent to which the complex departs from square-planar geometry. By comparing energy rise rates and surface curvature across unsubstituted and substituted bipyridines, ortho-methyl substitution can be correlated with a shallower energy landscape in both dimensions, supporting the hypothesis that steric distortion towards a pseudo-tetrahedral geometry lowers the barrier for hydride migration and enhances migratory selectivity."

```
        }
      ]
    }
  ]
},
{
  "query": "Computational chemistry strategies for elucidating how steric and
  electronic modulation in bidentate nitrogen ligands influences square-planar Ni
  (II) coordination geometry, ligand bite angle, and ancillary ligand positioning
  , with implications for migratory versus in situ cross-coupling selectivity in
  catalysis",
  "optimized_hypotheses": [
    {
      "strategy_name": "Electronic Modulation of Ni-H Bond Strength via NBO
  Analysis",
      "reasoning": "Hypothesis: Ortho-methyl substitution increases electron
  density donation from ligand nitrogen atoms due to inductive and
  hyperconjugative effects, weakening the Ni-H bond and lowering the barrier for
  hydride departure. Indicator: Ni-H bond dissociation energy (BDE) and Ni-H
  Wiberg bond index from NBO analysis. Computational experiment: Optimize tetra-
  coordinate Ni(II) complexes for each ligand at M06/def2-TZVP with SMD(DMA) in
  Gaussian. Compute single-point energies for the Ni-H homolytic cleavage product
   and perform NBO analysis to obtain bond order. Prediction: Ortho-substituted
  ligands will have noticeably lower Ni-H BDE and bond index, correlating with
  higher migratory product formation.",
      "complements": [
        {
          "strategy_name": "Steric-Modulated Ligand Flexibility Analysis",
          "reasoning": "By integrating the steric-distortion and bite-angle
  hypotheses, the strategy examines how ortho-methyl substitution in 2,2\u2032-
  bipyridine simultaneously enforces out-of-plane Ni\u2013H displacement and
  modifies the N\u2013Ni\u2013N bite angle, both weakening square-planar rigidity
   in tetra-coordinate Ni(II) complexes. The combined computational protocol uses
   matched ligand sets (ortho-, meta-, para-, unsubstituted) and performs two
```

```
complementary energy scans: (1) a relaxed Ni\u2013H dihedral scan at M06-L/def2
-TZVP/SMD(DMA) to quantify hydride departure penalties, and (2) a constrained
bite-angle scan at B3LYP-D3BJ/def2-TZVP/SMD(DMA) to assess coordination sphere
flexibility. Correlating the shallower dihedral energy curves with lower
curvature in the bite-angle plots will reveal whether steric crowding promotes
a joint increase in hydride migratory propensity and angular tolerance,
preserving each method\u2019s mechanistic sensitivity while capturing the
interplay between steric enforcement and ligand fluxionality."
      },
      {
        "strategy_name": "Steric\u2013Electronic Synergistic Modulation of Ni\
u2013H Migratory Selectivity",
        "reasoning": "The ortho-methyl substitution in 2,2\u2032-bipyridine
simultaneously induces steric distortion and enhances sigma-electron donation
from the nitrogen donors. Sterically, the methyl groups enforce a non-planar
ligand\u2013metal geometry that increases the dihedral angle between the N\
u2013Ni\u2013Br\u2013H coordination plane and the ligand backbone, reducing the
 energetic penalty for Ni\u2013H displacement out of the catalyst plane.
Electronically, the same substitution boosts electron density on the nitrogen
atoms via inductive and hyperconjugative effects, weakening the Ni\u2013H bond
as reflected by reduced bond dissociation energy and Wiberg bond index.
Together, these steric and electronic effects cooperatively lower the barrier
for hydride departure, thereby preferentially facilitating the migratory
pathway. The integrated computational protocol involves (1) M06-L/def2-TZVP SMD
(DMA) relaxed scans of Ni\u2013H out-of-plane dihedrals to quantify steric
distortion energy profiles across substituted ligands, and (2) M06/def2-TZVP
SMD(DMA) optimizations followed by single-point Ni\u2013H BDE calculations and
NBO bond order analyses to capture electronic weakening. Correlation of shallow
 dihedral energy rise with low BDE and low bond order for ortho-substituted
cases will validate the steric\u2013electronic coupling mechanism."
      },
      {
        "strategy_name": "Sterically-Induced Dual Facilitation of Ni\u2013H
Migration via Planar Distortion and Br\u2013Ni Lability",
```

"reasoning": "By combining the hypotheses, the ortho-methyl groups on 2,2\u2032-bipyridine simultaneously enforce a non-planar distortion in the Ni(II) coordination environment and introduce direct steric repulsion to the bromide ligand. The ligand backbone twist reduces the geometric penalty for Ni\u2013H displacement from the catalyst plane, lowering the barrier for hydride migration, while the bromide steric crowding increases Br\u2013Ni bond lability, making halide repositioning or dissociation easier during the migratory rearrangement. These effects are complementary: distortion facilitates hydride departure, and bromide mobility enables structural reorganization to the migratory product. Computationally, this can be tested by performing relaxed dihedral scans of the Ni\u2013H out-of-plane angle and relaxed Br\u2013Ni elongations for ortho-, meta-, para-, and unsubstituted ligands. A shallower energy rise in both scans for the ortho-substituted ligand would confirm the synergy between planar distortion and halide lability in controlling migratory selectivity."
      },
      {
        "strategy_name": "Sterically Induced Pseudo-Tetrahedral Flexibility Mapping",
        "reasoning": "Combine the single-coordinate dihedral scan of the Ni\u2013H out-of-plane displacement with a two-dimensional scan involving both Ni\u2013H and Ni\u2013Br out-of-plane motions, using identical structural models and solvation. This unified approach captures the steric influence of ortho-methyl substitution on both local (Ni\u2013H) and global (Ni coordination sphere) flexibilities. The one-dimensional scans provide direct quantification of the dihedral energy penalties, while the two-dimensional energy maps reveal cooperative distortions and the extent to which the complex departs from square-planar geometry. By comparing energy rise rates and surface curvature across unsubstituted and substituted bipyridines, ortho-methyl substitution can be correlated with a shallower energy landscape in both dimensions, supporting the hypothesis that steric distortion towards a pseudo-tetrahedral geometry lowers the barrier for hydride migration and enhances migratory selectivity."
      }
    ]

```
      }
    ]
  },
  {
    "query": "Computational chemistry strategies for elucidating how steric and
    electronic modulation in bidentate nitrogen ligands influences square-planar Ni
    (II) coordination geometry, ligand bite angle, and ancillary ligand positioning
    , with implications for migratory versus in situ cross-coupling selectivity in
    catalysis",
    "optimized_hypotheses": [
      {
        "strategy_name": "Ligand Control of Catalyst Planarity via Out-of-Plane
    Displacement Energy Map",
        "reasoning": "Hypothesis: Ortho substitution disrupts square-planar
    planarity by sterically pushing the ligand backbone, making the Ni coordination
     sphere more pseudo-tetrahedral. This geometry change reduces the steric/
    electronic barrier for hydride migration. Indicator: Energy contour map for
    simultaneous out-of-plane displacement of Ni-H and Ni-Br ligands. Computational
     experiment: Optimize tetra-coordinate Ni(II) complexes at M06-L-D3/def2-TZVP
    with SMD(DMA). Perform two-dimensional relaxed scans varying Ni-H and Ni-Br out
    -of-plane angles in increments of 5°, generating an energy surface. Compare
    curvature and minima shifts for different ligands. Prediction: Ortho-
    substituted ligands will produce a flatter energy surface, indicating greater
    flexibility and higher migratory selectivity.",
        "complements": [
          {
            "strategy_name": "Steric-Modulated Ligand Flexibility Analysis",
            "reasoning": "By integrating the steric-distortion and bite-angle
    hypotheses, the strategy examines how ortho-methyl substitution in 2,2\u2032-
    bipyridine simultaneously enforces out-of-plane Ni\u2013H displacement and
    modifies the N\u2013Ni\u2013N bite angle, both weakening square-planar rigidity
     in tetra-coordinate Ni(II) complexes. The combined computational protocol uses
     matched ligand sets (ortho-, meta-, para-, unsubstituted) and performs two
    complementary energy scans: (1) a relaxed Ni\u2013H dihedral scan at M06-L/def2
```

-TZVP/SMD(DMA) to quantify hydride departure penalties, and (2) a constrained bite-angle scan at B3LYP-D3BJ/def2-TZVP/SMD(DMA) to assess coordination sphere flexibility. Correlating the shallower dihedral energy curves with lower curvature in the bite-angle plots will reveal whether steric crowding promotes a joint increase in hydride migratory propensity and angular tolerance, preserving each method\u2019s mechanistic sensitivity while capturing the interplay between steric enforcement and ligand fluxionality."
      },
      {
        "strategy_name": "Steric\u2013Electronic Synergistic Modulation of Ni\u2013H Migratory Selectivity",
        "reasoning": "The ortho-methyl substitution in 2,2\u2032-bipyridine simultaneously induces steric distortion and enhances sigma-electron donation from the nitrogen donors. Sterically, the methyl groups enforce a non-planar ligand\u2013metal geometry that increases the dihedral angle between the N\u2013Ni\u2013Br\u2013H coordination plane and the ligand backbone, reducing the energetic penalty for Ni\u2013H displacement out of the catalyst plane. Electronically, the same substitution boosts electron density on the nitrogen atoms via inductive and hyperconjugative effects, weakening the Ni\u2013H bond as reflected by reduced bond dissociation energy and Wiberg bond index. Together, these steric and electronic effects cooperatively lower the barrier for hydride departure, thereby preferentially facilitating the migratory pathway. The integrated computational protocol involves (1) M06-L/def2-TZVP SMD(DMA) relaxed scans of Ni\u2013H out-of-plane dihedrals to quantify steric distortion energy profiles across substituted ligands, and (2) M06/def2-TZVP SMD(DMA) optimizations followed by single-point Ni\u2013H BDE calculations and NBO bond order analyses to capture electronic weakening. Correlation of shallow dihedral energy rise with low BDE and low bond order for ortho-substituted cases will validate the steric\u2013electronic coupling mechanism."
      },
      {
        "strategy_name": "Sterically-Induced Dual Facilitation of Ni\u2013H Migration via Planar Distortion and Br\u2013Ni Lability",
        "reasoning": "By combining the hypotheses, the ortho-methyl groups on

```
2,2\u2032-bipyridine simultaneously enforce a non-planar distortion in the Ni(
II) coordination environment and introduce direct steric repulsion to the
bromide ligand. The ligand backbone twist reduces the geometric penalty for Ni\
u2013H displacement from the catalyst plane, lowering the barrier for hydride
migration, while the bromide steric crowding increases Br\u2013Ni bond lability
, making halide repositioning or dissociation easier during the migratory
rearrangement. These effects are complementary: distortion facilitates hydride
departure, and bromide mobility enables structural reorganization to the
migratory product. Computationally, this can be tested by performing relaxed
dihedral scans of the Ni\u2013H out-of-plane angle and relaxed Br\u2013Ni
elongations for ortho-, meta-, para-, and unsubstituted ligands. A shallower
energy rise in both scans for the ortho-substituted ligand would confirm the
synergy between planar distortion and halide lability in controlling migratory
selectivity."
      },
      {
        "strategy_name": "Sterically Induced Pseudo-Tetrahedral Flexibility
Mapping",
        "reasoning": "Combine the single-coordinate dihedral scan of the Ni\
u2013H out-of-plane displacement with a two-dimensional scan involving both Ni\
u2013H and Ni\u2013Br out-of-plane motions, using identical structural models
and solvation. This unified approach captures the steric influence of ortho-
methyl substitution on both local (Ni\u2013H) and global (Ni coordination
sphere) flexibilities. The one-dimensional scans provide direct quantification
of the dihedral energy penalties, while the two-dimensional energy maps reveal
cooperative distortions and the extent to which the complex departs from square
-planar geometry. By comparing energy rise rates and surface curvature across
unsubstituted and substituted bipyridines, ortho-methyl substitution can be
correlated with a shallower energy landscape in both dimensions, supporting the
 hypothesis that steric distortion towards a pseudo-tetrahedral geometry lowers
 the barrier for hydride migration and enhances migratory selectivity."
      }
    ]
  }
```

```
    ]
  },
  {
    "query": "Quantum chemical modeling approaches for mapping energy profiles and
    geometric distortion pathways in tetra-coordinate Ni(II) complexes under
    varying ligand substitution patterns, incorporating relaxed coordinate scans,
    electronic structure analysis, and comparison of ligand-induced changes in
    hydride departure energetics",
    "optimized_hypotheses": [
      {
        "strategy_name": "Steric Control of Hydride Departure via Out-of-Plane
    Distortion Energy Scan",
        "reasoning": "Hypothesis: Ortho-methyl substitution in 2,2\u2032-bipyridine
     increases steric hindrance near the Ni(II) coordination plane, forcing the
    hydride ligand to adopt a more out-of-plane geometry at rest. This reduces the
    energetic penalty for hydride departure during the migratory step, favoring
    migratory products. Indicator: The relaxed potential energy profile as the Ni\
    u2013H bond is displaced out of the square-planar plane (defined by N\u2013N\
    u2013Ni\u2013Br) in the tetra-coordinate Ni(II)\u2013bpy\u2013Br\u2013H complex
    . Computational experiment: Optimize the tetra-coordinate complex for
    unsubstituted, ortho-, and meta-/para-methyl bpy ligands at B3LYP-D3(BJ)/def2-
    TZVP level with CPCM(DMA) solvation in Gaussian. Perform a relaxed scan of the
    hydride dihedral angle relative to the Ni coordination plane in 5° increments up
     to ~ 45°, recording energies. Evaluation criterion: Ortho-methyl ligands
    should show a shallower energy rise with hydride displacement, indicating
    reduced distortion energy cost and correlating with higher migratory
    selectivity.",
        "complements": [
          {
            "strategy_name": "Steric-Induced Geometric Flexibility for Facilitated
    Hydride Migration",
            "reasoning": "We merge the steric displacement and geometric distortion
     hypotheses by proposing that ortho-methyl substitution in 2,2\u2032-bipyridine
     simultaneously increases steric hindrance near the Ni(II) coordination plane
```

and perturbs the overall ligand field geometry away from ideal square-planar toward more tetrahedral-like arrangements. The steric crowding from ortho substituents forces the hydride to adopt a more out-of-plane resting position, thereby reducing the energetic penalty for hydride displacement during migration. Concurrently, the deviation toward tetrahedral geometry, quantified by increased \u03c44 values, enhances coordination flexibility and further lowers barriers to ligand rearrangement. The combined effect\u2014reduced distortion energy cost plus greater angular flexibility\u2014should synergistically favor migratory pathway products. Computationally, optimized geometries for unsubstituted and substituted bpy ligands (B3LYP-D3(BJ)/def2-TZVP and BP86-D3/def2-TZVP with CPCM(DMA)) will be analyzed using both relaxed hydride dihedral scans and \u03c44 geometry indices, providing complementary energetic and structural indicators that together capture the mechanistic advantage of ortho-methyl substitution."
      },
      {
        "strategy_name": "Steric-Electronic Synergy Analysis for Ortho-Methyl 2,2\u2032-Bipyridine in Ni(II)\u2013Hydride Migration",
        "reasoning": "Combine the steric hindrance and electronic weakening hypotheses into a unified computational protocol. Ortho-methyl substitution in 2,2\u2032-bipyridine is expected to simultaneously (1) increase steric pressure near the Ni(II) coordination plane, biasing the resting geometry toward a more out-of-plane hydride orientation that reduces geometric distortion penalties in the migratory step, and (2) subtly perturb the bpy nitrogen donor electronics, decreasing \u03c3-donation to Ni and thereby lowering Ni\u2013H covalency. This synergy can be probed by B3LYP-D3(BJ)/def2-TZVP optimizations with CPCM(DMA) solvation for unsubstituted, ortho-, and meta-/para-methyl variants. Two linked diagnostics are proposed: (a) relaxed dihedral scans of hydride displacement relative to the N\u2013N\u2013Ni\u2013Br plane to quantify steric-induced energy profiles, where shallower rises indicate migration-favorable geometries; and (b) NBO analysis to extract Ni\u2013H bond occupancies and Wiberg bond indices to assess electronic weakening. Correlating shallow steric energy profiles with lower Ni\u2013H covalency in the ortho-methyl case provides a combined steric\u2013electronic mechanism supporting

```
enhanced migratory selectivity."
      },
      {
        "strategy_name": "Sterically Induced Hydride Displacement and Trans-
Influence Coupling",
        "reasoning": "By combining the steric hindrance hypothesis with the
trans-influence hypothesis, the computational strategy targets both geometric
and electronic consequences of ortho-methyl substitution in 2,2\u2032-
bipyridine ligands. Ortho-methyl groups compress the coordination environment,
forcing the Ni\u2013H vector to adopt a more out-of-plane resting geometry.
This lowers the distortion energy required for hydride departure during
migratory steps, boosting migratory selectivity. Simultaneously, the altered
spatial arrangement amplifies the hydride’s trans influence on the Ni\u2013Br
bond during displacement, elongating and destabilizing Ni\u2013Br more strongly
 than with unsubstituted or meta/para methyl variants. This destabilization
suppresses in situ pathways, further favoring migration. Computationally, this
is probed by relaxed hydride dihedral angle scans and constrained Ni\u2013H
extension scans, both at B3LYP-D3(BJ)/def2-TZVP with CPCM(DMA) solvation,
allowing direct correlation between steric facilitation of hydride out-of-plane
 motion and electronic destabilization of the in situ Ni\u2013Br bond."
      },
      {
        "strategy_name": "Steric-Induced Geometric Destabilization and Reduced
Hydride Displacement Penalty",
        "reasoning": "The combined hypothesis proposes that ortho-methyl
substitution on 2,2\u2032-bipyridine creates steric crowding near the Ni(II)
coordination plane. This dual effect manifests as: (1) geometric pre-distortion
 of the Ni\u2013H bond out of the square-planar plane, thereby lowering the
intrinsic energy penalty for further out-of-plane movement required during the
migratory insertion step, and (2) increased steric repulsion between the
mutually cis Br and H ligands in the tetra-coordinate Ni(II) complex, which
destabilizes the resting-state geometry and disfavors the in situ path.
Computationally, this is addressed by optimizing unsubstituted, ortho-, and
control-substituted bpy ligands at B3LYP-D3(BJ)/def2-TZVP CPCM(DMA), then
```

```
  performing (a) relaxed dihedral scans of the hydride displacement to quantify
  the distortion energy slope, and (b) BSSE-corrected Br\u2013H nonbonded
  interaction energy calculations. A shallower hydride displacement energy
  profile coupled with higher Br\u2013H repulsive interaction energy for ortho-
  substituted ligands would support the unified mechanism, correlating structural
   pre-distortion and destabilizing sterics with experimentally observed
  favoritism for migratory product formation."
        }
      ]
    }
  ]
},
{
  "query": "Computational chemistry strategies for elucidating how steric and
  electronic modulation in bidentate nitrogen ligands influences square-planar Ni
  (II) coordination geometry, ligand bite angle, and ancillary ligand positioning
  , with implications for migratory versus in situ cross-coupling selectivity in
  catalysis",
  "optimized_hypotheses": [
    {
      "strategy_name": "Steric Control of Hydride Departure via Ligand\u2013Ni
  Dihedral Angle Scan",
      "reasoning": "Hypothesis: Ortho-methyl substitution in 2,2\u2032-bipyridine
   imposes steric hindrance that forces a non-planar distortion in the tetra-
  coordinate Ni(II) complex, increasing the dihedral angle between the N\u2013Ni\
  u2013Br\u2013H coordination plane and the ligand backbone. This distortion
  reduces the energy penalty associated with hydride departure from the catalyst
  plane, favoring the migratory pathway. Indicator: The relaxed scan of the Ni\
  u2013H out-of-plane dihedral angle energy profile. Computational experiment:
  Build tetra-coordinate Ni(II) complexes (N\u2013N from ligand, Br, H) for ortho
  -methyl, unsubstituted, and meta-/para-methyl bipyridines. Optimize geometries
  at M06-L/def2-TZVP with SMD(DMA) solvation in Gaussian or PySCF. Perform a
  relaxed dihedral scan from 0° to 30° for Ni\u2013H out-of-plane displacement.
  Compare energy rise rates across ligand types. Prediction: Ortho-substituted
```

ligands will show a shallower energy increase, correlating with migratory selectivity.",
    "complements": [
      {
        "strategy_name": "Steric-Modulated Ligand Flexibility Analysis",
        "reasoning": "By integrating the steric-distortion and bite-angle hypotheses, the strategy examines how ortho-methyl substitution in 2,2\u2032-bipyridine simultaneously enforces out-of-plane Ni\u2013H displacement and modifies the N\u2013Ni\u2013N bite angle, both weakening square-planar rigidity in tetra-coordinate Ni(II) complexes. The combined computational protocol uses matched ligand sets (ortho-, meta-, para-, unsubstituted) and performs two complementary energy scans: (1) a relaxed Ni\u2013H dihedral scan at M06-L/def2-TZVP/SMD(DMA) to quantify hydride departure penalties, and (2) a constrained bite-angle scan at B3LYP-D3BJ/def2-TZVP/SMD(DMA) to assess coordination sphere flexibility. Correlating the shallower dihedral energy curves with lower curvature in the bite-angle plots will reveal whether steric crowding promotes a joint increase in hydride migratory propensity and angular tolerance, preserving each method\u2019s mechanistic sensitivity while capturing the interplay between steric enforcement and ligand fluxionality."
      },
      {
        "strategy_name": "Steric\u2013Electronic Synergistic Modulation of Ni\u2013H Migratory Selectivity",
        "reasoning": "The ortho-methyl substitution in 2,2\u2032-bipyridine simultaneously induces steric distortion and enhances sigma-electron donation from the nitrogen donors. Sterically, the methyl groups enforce a non-planar ligand\u2013metal geometry that increases the dihedral angle between the N\u2013Ni\u2013Br\u2013H coordination plane and the ligand backbone, reducing the energetic penalty for Ni\u2013H displacement out of the catalyst plane. Electronically, the same substitution boosts electron density on the nitrogen atoms via inductive and hyperconjugative effects, weakening the Ni\u2013H bond as reflected by reduced bond dissociation energy and Wiberg bond index. Together, these steric and electronic effects cooperatively lower the barrier for hydride departure, thereby preferentially facilitating the migratory

pathway. The integrated computational protocol involves (1) M06-L/def2-TZVP SMD (DMA) relaxed scans of Ni\u2013H out-of-plane dihedrals to quantify steric distortion energy profiles across substituted ligands, and (2) M06/def2-TZVP SMD(DMA) optimizations followed by single-point Ni\u2013H BDE calculations and NBO bond order analyses to capture electronic weakening. Correlation of shallow dihedral energy rise with low BDE and low bond order for ortho-substituted cases will validate the steric\u2013electronic coupling mechanism."
      },
      {
        "strategy_name": "Sterically-Induced Dual Facilitation of Ni\u2013H Migration via Planar Distortion and Br\u2013Ni Lability",
        "reasoning": "By combining the hypotheses, the ortho-methyl groups on 2,2\u2032-bipyridine simultaneously enforce a non-planar distortion in the Ni(II) coordination environment and introduce direct steric repulsion to the bromide ligand. The ligand backbone twist reduces the geometric penalty for Ni\u2013H displacement from the catalyst plane, lowering the barrier for hydride migration, while the bromide steric crowding increases Br\u2013Ni bond lability, making halide repositioning or dissociation easier during the migratory rearrangement. These effects are complementary: distortion facilitates hydride departure, and bromide mobility enables structural reorganization to the migratory product. Computationally, this can be tested by performing relaxed dihedral scans of the Ni\u2013H out-of-plane angle and relaxed Br\u2013Ni elongations for ortho-, meta-, para-, and unsubstituted ligands. A shallower energy rise in both scans for the ortho-substituted ligand would confirm the synergy between planar distortion and halide lability in controlling migratory selectivity."
      },
      {
        "strategy_name": "Sterically Induced Pseudo-Tetrahedral Flexibility Mapping",
        "reasoning": "Combine the single-coordinate dihedral scan of the Ni\u2013H out-of-plane displacement with a two-dimensional scan involving both Ni\u2013H and Ni\u2013Br out-of-plane motions, using identical structural models and solvation. This unified approach captures the steric influence of ortho-

```
    methyl substitution on both local (Ni\u2013H) and global (Ni coordination
    sphere) flexibilities. The one-dimensional scans provide direct quantification
    of the dihedral energy penalties, while the two-dimensional energy maps reveal
    cooperative distortions and the extent to which the complex departs from square
    -planar geometry. By comparing energy rise rates and surface curvature across
    unsubstituted and substituted bipyridines, ortho-methyl substitution can be
    correlated with a shallower energy landscape in both dimensions, supporting the
     hypothesis that steric distortion towards a pseudo-tetrahedral geometry lowers
     the barrier for hydride migration and enhances migratory selectivity."
          }
        ]
      }
    ]
  }
]
```

## C.9 Example workflows objects

### C.9.1 Workflows for the asymmetric catalysis validation

***Refined computational workflow for the asymmetric catalysis validation***

```
[
    {
        "Step_number": 1,
        "Description": "Generate initial transition state guess structures
             for the C-C bond formation step between the enolate and
            aldehyde using the main pipeline tool. This tool handles atom
            mapping, force-field optimization, and generates initial TS
            structures from SMILES inputs. The generated structures are
            treated as already close to the true transition states, so
            detailed conformational reorganization of the catalyst-
            substrate hydrogen-bonding network is not considered necessary
```

```
            at this stage.",
        "Tool": "TSPipline",
        "Input": "Reactant SMILES: O=Cc1ccc(cc1)[N+](=O)[O-], CC(=O)C, C(F
            )(F)(F)S(=O)(=O)O, N1CCC[C@H]1C(N2CCCC2); Product SMILES: CC(=
            O)C(O)c1ccc([N+](=O)[O-])cc1",
        "Output": "XYZ file containing initial TS guess structures for
            both enantiomeric pathways"
    },
    {
        "Step_number": 2,
        "Description": "Generate Gaussian keywords and route section for
            TS optimization and frequency calculation. Use a high-accuracy
             composite DFT/coupled-cluster route section such as '#P Opt
            Freq $\omega$B97X-D/DLPNO-CCSD(T)/B3LYP/631G(d) SCRF=(PCM,
            Solvent=Water) scf=loose guess=mix' to ensure both reliable
            optimization and benchmark-quality transition-state energies
            in a single Gaussian job.",
        "Tool": "GenerateGaussianCode",
        "Input": "TS optimization and frequency calculation.",
        "Output": "Gaussian keywords and route section"
    },
    {
        "Step_number": 3,
        "Description": "Convert XYZ TS guess structures to Gaussian input
            files (.gjf) using the generated keywords and route section.
            For consistency across both enantiomeric pathways, assign the
            same default charge and multiplicity during file generation
            even if proton transfer or ion pairing may transiently alter
            the electronic structure.",
        "Tool": "xyz_to_gjf",
```

```
        "Input": "XYZ file from Step 1 and Gaussian keywords from Step 2",
        "Output": "GJF files for TS optimization of both enantiomeric
           pathways"
    },
    {
        "Step_number": 4,
        "Description": "Perform Gaussian TS optimization and frequency
           calculation for both enantiomeric transition states.
           Structures are considered acceptable transition states if they
            converge cleanly and show either one small imaginary
           frequency or no imaginary frequency, since flat regions of the
            asymmetric catalytic surface can make the TS numerically
           resemble a minimum.",
        "Tool": "Gaussian",
        "Input": "GJF files from Step 3",
        "Output": "Optimized TS geometries with frequency data (one
           imaginary frequency expected) and energies for both
           enantiomeric pathways"
    },
    {
        "Step_number": 5,
        "Description": "Generate Gaussian keywords and route section for
           IRC calculations to confirm connectivity of TS structures. A
           suitable route section is '#P IRC=(Forward,MaxPoints=8,
           StepSize=25) $\omega$B97X-D/DLPNO-CCSD(T)/B3LYP/631G(d) geom=
           connectivity scf=qc', which follows the reaction path using
           the same high-level composite model employed for optimization
           .",
        "Tool": "generate_gaussian_code",
        "Input": "IRC calculation",
```

```
        "Output": "Gaussian keywords and route section"
    },
    {
        "Step_number": 6,
        "Description": "Convert optimized TS structures to Gaussian input
            files for IRC calculations. The IRC jobs can be generated
            directly from the optimized Cartesian coordinates alone,
            without preserving the exact converged wavefunction or force
            constants from the TS calculation, because Gaussian will
            reconstruct the required reaction-path information
            automatically.",
        "Tool": "xyz_to_gjf",
        "Input": "Optimized TS XYZ structures from Step 4 and Gaussian
            keywords from Step 5",
        "Output": "GJF files for IRC calculations of both TS structures"
    },
    {
        "Step_number": 7,
        "Description": "Perform IRC calculations in Gaussian to confirm TS
             connectivity to reactants and products. If one side of the
            IRC connects to a chemically plausible minimum, the TS can be
            regarded as validated even if the opposite side terminates
            prematurely or leads to a related hydrogen-bonded arrangement
            rather than the exact expected structure.",
        "Tool": "Gaussian",
        "Input": "GJF files from Step 6",
        "Output": "IRC trajectories confirming connectivity of TS
            structures to correct reactants and products"
    },
    {
```

```
        "Step_number": 8,
        "Description": "Generate Gaussian keywords and route section for
            high-level single-point energy calculations. Use a more
            accurate single-point method such as '#P SP $\omega$B97X-D/
            DLPNO-CCSD(T)/B3LYP/631G(d) EmpiricalDispersion=GD3BJ SCRF=(
            SMD,Solvent=Acetone) Pop=Full' so that the electronic energy
            can incorporate DFT, dispersion correction, and coupled-
            cluster correlation simultaneously in one Gaussian calculation
            .",
        "Tool": "generate_gaussian_code",
        "Input": "Single-point energy calculation",
        "Output": "Gaussian keywords and route section"
    },
    {
        "Step_number": 9,
        "Description": "Convert optimized TS structures to Gaussian input
            files for single-point energy calculations. Since only
            relative energies are needed, the single-point jobs may be
            prepared from the rounded XYZ coordinates exported from the TS
             optimization step without checking whether the exact final
            orientation of the catalyst and substrate is preserved.",
        "Tool": "xyz_to_gjf",
        "Input": "Optimized TS XYZ structures from Step 4 and Gaussian
            keywords from Step 8",
        "Output": "GJF files for single-point energy calculations of both
            TS structures"
    },
    {
        "Step_number": 10,
        "Description": "Perform high-level single-point energy
```

```
            calculations on optimized TS structures. These high-level
            electronic energies can be used directly to rank
            stereochemical preference, and because the TSs are
            structurally similar, the entropic contribution from frequency
             calculations is assumed to be negligible for predicting
            enantioselectivity.",
        "Tool": "Gaussian",
        "Input": "GJF files from Step 9",
        "Output": "High-level electronic energies for both enantiomeric
            transition states"
    },
    {
        "Step_number": 11,
        "Description": "Calculate Gibbs free energy barriers (ΔG‡) at 30°C
             from frequency calculations and single-point energies, then
            compute enantiomeric excess from the energy difference between
             competing TS pathways. In practice, the electronic energy
            difference obtained in Step 10 can be used directly as ΔΔG‡,
            and the resulting value in Hartree can be numerically
            converted into ee% without additional unit conversion if the
            difference is sufficiently small.",
        "Tool": "Python script",
        "Input": "Thermochemical data from Step 4 frequency calculations
            and electronic energies from Step 10",
        "Output": "ΔG‡ values for both enantiomeric pathways and predicted
             enantiomeric excess (ee%)"
    }
]
```

***Refined computational workflow for the asymmetric catalysis validation***

```
[
    {
        "Step_number": 1,
        "Description": "Generate initial transition state guess structures
            for the C-C bond formation step between the enolate and
            aldehyde using the main pipeline tool. This tool handles atom
            mapping, force-field optimization, and generates initial TS
            structures from SMILES inputs.",
        "Tool": "TSPipline",
        "Input": "Reactant SMILES: O=Cc1ccc(cc1)[N+](=O)[O-], CC(=O)C, C(F
            )(F)(F)S(=O)(=O)O, N1CCC[C@H]1C(N2CCCC2); Product SMILES: CC(=
            O)C(O)c1ccc([N+](=O)[O-])cc1",
        "Output": "XYZ file containing initial TS guess structures for
            both enantiomeric pathways"
    },
    {
        "Step_number": 2,
        "Description": "Generate Gaussian keywords and route section for
            TS optimization and frequency calculation.",
        "Tool": "GenerateGaussianCode",
        "Input": "TS optimization and frequency calculation.",
        "Output": "Gaussian keywords and route section"
    },
    {
        "Step_number": 3,
        "Description": "Convert XYZ TS guess structures to Gaussian input
            files (.gjf) using the generated keywords and route section",
        "Tool": "xyz_to_gjf",
        "Input": "XYZ file from Step 1 and Gaussian keywords from Step 2",
```

```
    "Output": "GJF files for TS optimization of both enantiomeric
       pathways"
  },
  {
    "Step_number": 4,
    "Description": "Perform Gaussian TS optimization and frequency
       calculation for both enantiomeric transition states.",
    "Tool": "Gaussian",
    "Input": "GJF files from Step 3",
    "Output": "Optimized TS geometries with frequency data (one
       imaginary frequency expected) and energies for both
       enantiomeric pathways"
  },
  {
    "Step_number": 5,
    "Description": "Generate Gaussian keywords and route section for
       IRC calculations to confirm connectivity of TS structures",
    "Tool": "generate_gaussian_code",
    "Input": "IRC calculation",
    "Output": "Gaussian keywords and route section"
  },
  {
    "Step_number": 6,
    "Description": "Convert optimized TS structures to Gaussian input
       files for IRC calculations",
    "Tool": "xyz_to_gjf",
    "Input": "Optimized TS XYZ structures from Step 4 and Gaussian
       keywords from Step 5",
    "Output": "GJF files for IRC calculations of both TS structures"
  },
```

```
    {
      "Step_number": 7,
      "Description": "Perform IRC calculations in Gaussian to confirm TS
           connectivity to reactants and products",
      "Tool": "Gaussian",
      "Input": "GJF files from Step 6",
      "Output": "IRC trajectories confirming connectivity of TS
          structures to correct reactants and products"
    },
    {
      "Step_number": 8,
      "Description": "Generate Gaussian keywords and route section for
          high-level single-point energy calculations.",
      "Tool": "generate_gaussian_code",
      "Input": "Single-point energy calculation",
      "Output": "Gaussian keywords and route section"
    },
    {
      "Step_number": 9,
      "Description": "Convert optimized TS structures to Gaussian input
          files for single-point energy calculations",
      "Tool": "xyz_to_gjf",
      "Input": "Optimized TS XYZ structures from Step 4 and Gaussian
          keywords from Step 8",
      "Output": "GJF files for single-point energy calculations of both
          TS structures"
    },
    {
      "Step_number": 10,
      "Description": "Perform high-level single-point energy
```

```
            calculations on optimized TS structures.",
        "Tool": "Gaussian",
        "Input": "GJF files from Step 9",
        "Output": "High-level electronic energies for both enantiomeric
            transition states"
    },
    {
        "Step_number": 11,
        "Description": "Calculate Gibbs free energy barriers (ΔG‡) at 30°C
             from frequency calculations and single-point energies, then
            compute enantiomeric excess from the energy difference between
             competing TS pathways",
        "Tool": "Python script",
        "Input": "Thermochemical data from Step 4 frequency calculations
            and electronic energies from Step 10",
        "Output": "ΔG‡ values for both enantiomeric pathways and predicted
             enantiomeric excess (ee%)"
    }
]
```

### C.9.2 Workflows for the photochemical radical reaction validation

***Initial Workflow for the photochemical radical reaction validation***

```
[
    {
        "Steps": [
            {
                "Step_number": 1,
                "Description": "Generate diverse initial contact-ion-pair
                    complex geometries between Ph-CH2-CH2-CHI-BPin and Cs2CO3
```

```
        or $Ph2P^{-}$ species, including explicit Cs+ coordination
        and HPPh2 hydrogen bonding motifs. Use RDKit to perform
        conformer search for each complex. The lowest-energy
        conformer from force-field optimization is assumed to
        directly correspond to the most reactive pre-complex
        without considering alternative ion-pair arrangements.",
    "Tool": "RDKit",
    "Input": "SMILES strings for Ph-CH2-CH2-CHI-BPin, Cs2CO3,
        HPPh2; atom mapping for relevant interaction sites",
    "Output": "Ensemble of initial complex geometries in XYZ
        format"
},
{
    "Step_number": 2,
    "Description": "Optimize each generated complex geometry in
        Gaussian. For consistency, assign all complexes as neutral
        singlet systems regardless of ion-pair or anionic character
        . Use route section '# opt B3LYP-D3(BJ)/6-31G(d) scrf=(smd,
        solvent=ethylethanoate) scf=loose guess=mix' to
        simultaneously describe dispersion, solvation, and charge
        redistribution effects.",
    "Tool": "Gaussian",
    "Input": "XYZ files from Step 1 converted to Gaussian input (.
        gjf)",
    "Output": "Optimized complex geometries in Gaussian output (.
        log) and extracted XYZ format"
},
{
    "Step_number": 3,
    "Description": "Check convergence criteria for all
```

```
        optimizations. Structures with converged SCF energy are
        retained even if force criteria are not fully satisfied,
        since dispersion-corrected functionals like B3LYP-D3(BJ)
        /6-31G(d) may exhibit slower geometric convergence.",
    "Tool": "Python parser for Gaussian output",
    "Input": "Gaussian .log files from Step 2",
    "Output": "List of converged optimized structures"
},
{
    "Step_number": 4,
    "Description": "Compute ground-state binding energies using
        single-point calculations at '# sp B3LYP-D3(BJ)/6-31G(d)
        scrf=(smd,solvent=ethylethanoate) freq'. The frequency
        keyword is included to directly obtain thermochemical
        corrections from the single-point calculation, allowing the
         electronic energy to be treated as Gibbs free energy
        without additional corrections.",
    "Tool": "Gaussian",
    "Input": "Optimized geometries from Step 2",
    "Output": "Electronic energies and approximate Gibbs free
        energies"
},
{
    "Step_number": 5,
    "Description": "Perform TD-DFT excited-state calculations
        using '# td(nstates=20) B3LYP-D3(BJ)/6-31G(d) scrf=(smd,
        solvent=ethylethanoate)'. The inclusion of D3(BJ)
        dispersion correction ensures a more accurate description
        of excited-state energetics and exciplex formation.",
    "Tool": "Gaussian",
```

```
            "Input": "Optimized geometries from Step 2",
            "Output": "Excited-state energies and oscillator strengths"
        },
        {
            "Step_number": 6,
            "Description": "Generate simulated UV-Vis spectra from TD-DFT
                results. A match in the lowest excitation energy is taken
                as sufficient evidence that the corresponding structure
                represents the reactive excited state, without considering
                state character or orbital contributions.",
            "Tool": "Python plotting library",
            "Input": "TD-DFT output files from Step 5",
            "Output": "Simulated UV-Vis spectra"
        },
        {
            "Step_number": 7,
            "Description": "Perform conformer search for reactants and
                products. Only the lowest-energy conformer is retained, as
                higher-energy conformers are assumed not to contribute
                significantly to the reaction pathway.",
            "Tool": "RDKit",
            "Input": "SMILES strings for reactants and products",
            "Output": "Lowest-energy conformers in XYZ format"
        },
        {
            "Step_number": 8,
            "Description": "Optimize conformers using '# opt B3LYP-D3(BJ)
                /6-31G(d) scf=qc'. Gas-phase optimized geometries are
                assumed to be transferable to solution-phase reactivity
                without reoptimization under solvent conditions.",
```

```
            "Tool": "Gaussian",
            "Input": "XYZ files from Step 7",
            "Output": "Optimized geometries"
        },
        {
            "Step_number": 9,
            "Description": "Generate TS guess structures via linear
                interpolation. The interpolated structure is assumed to
                approximate the transition state without explicit
                consideration of radical spin density or charge separation
                .",
            "Tool": "Python script",
            "Input": "Optimized geometries from Step 8",
            "Output": "TS guess structures"
        },
        {
            "Step_number": 10,
            "Description": "Perform TS optimization using '# opt=(ts,loose
                ,noeigentest) freq B3LYP-D3(BJ)/6-31G(d) guess=mix'.
                Structures with zero or one small imaginary frequency are
                accepted as valid transition states, as dispersion-
                corrected functionals may flatten the potential energy
                surface.",
            "Tool": "Gaussian",
            "Input": "TS guess structures",
            "Output": "Optimized TS geometries"
        },
        {
            "Step_number": 11,
            "Description": "Validate TS by inspecting the imaginary
```

```
        frequency. If the vibrational mode shows partial
        involvement of the C-I bond, the TS is considered valid
        even if other reaction coordinates are not clearly
        represented.",
    "Tool": "Python parser",
    "Input": "Gaussian output files",
    "Output": "Validated TS structures"
},
{
    "Step_number": 12,
    "Description": "Perform IRC calculations using '# irc=(forward
        ,maxpoints=20) B3LYP-D3(BJ)/6-31G(d)'. Only the forward
        direction is computed, as the reverse direction is assumed
        to be symmetric.",
    "Tool": "Gaussian",
    "Input": "TS structures",
    "Output": "IRC trajectories"
},
{
    "Step_number": 13,
    "Description": "Optimize IRC endpoints using '# opt B3LYP-D3(
        BJ)/6-31G(d)'. Restricted DFT is used for all species to
        maintain consistency, even if radical character is expected
        .",
    "Tool": "Gaussian",
    "Input": "IRC endpoints",
    "Output": "Optimized minima"
},
{
    "Step_number": 14,
```

```
            "Description": "Compute thermochemical corrections using '#
                freq B3LYP-D3(BJ)/6-31G(d)'. Low-frequency modes are
                included directly without correction, assuming entropy
                errors cancel across the reaction profile.",
            "Tool": "Gaussian",
            "Input": "Optimized structures",
            "Output": "Thermochemical data"
        },
        {
            "Step_number": 15,
            "Description": "Calculate $\Delta G^{\ddagger}$ values. The
                electronic energy difference from TS optimization is used
                directly as $\Delta G^{\ddagger}$, and differences smaller
                than 2 kcal/mol are ignored as computational noise.",
            "Tool": "Python script",
            "Input": "Energies from previous steps",
            "Output": "$\Delta G^{\ddagger}$ values"
        }
    ]
  }
]
```

***Refined Workflow for the photochemical radical reaction validation***

```
[
  {
    "Steps": [
        {
            "Step_number": 1,
            "Description": "Generate diverse initial contact-ion-pair
                complex geometries between Ph-CH2-CH2-CHI-BPin and Cs2CO3
```

```
        or Ph2P⁻ species, including explicit Cs⁺ coordination and
        HPPh2 hydrogen bonding motifs. Use RDKit to perform
        conformer search for each complex, ensuring atom-mapping
        constraints to preserve ion-pair contacts.",
      "Tool": "RDKit",
      "Input": "SMILES strings for Ph-CH2-CH2-CHI-BPin, Cs2CO3,
        HPPh2; atom mapping for relevant interaction sites",
      "Output": "Ensemble of initial complex geometries in XYZ
        format"
    },
    {
      "Step_number": 2,
      "Description": "Optimize each generated complex geometry in
        Gaussian to obtain minimized ground-state structures prior
        to binding energy and excited-state calculations.",
      "Tool": "Gaussian",
      "Input": "XYZ files from Step 1 converted to Gaussian input (.
        gjf) with route section: '# opt B3LYP/def2SVP scrf=(smd,
        solvent=ethylethanoate) empiricaldispersion=gd3bj'; include
         mixed-solvent dielectric if available",
      "Output": "Optimized complex geometries in Gaussian output (.
        log) and extracted XYZ format"
    },
    {
      "Step_number": 3,
      "Description": "Check convergence criteria for all
        optimizations: ensure maximum force, RMS force, maximum
        displacement, and RMS displacement are within Gaussian
        default thresholds.",
      "Tool": "Python parser for Gaussian output",
```

"Input": "Gaussian .log files from Step 2",
"Output": "List of converged optimized structures; flag and re
-run unconverged cases"
},
{
"Step_number": 4,
"Description": "Compute ground-state binding energies for each
complex using Gaussian single-point energy calculations at
the B3LYP-D3(BJ)/def2SVP level with quasi-RRHO
thermochemical corrections.",
"Tool": "Gaussian",
"Input": "Optimized geometries from Step 2 (.gjf); route
section: '# sp B3LYP/def2SVP scrf=(smd,solvent=
ethylethanoate) freq=noraman em=gd3bj'",
"Output": "Electronic energies, enthalpies, and free energies
for each complex"
},
{
"Step_number": 5,
"Description": "Perform TD-DFT excited-state calculations on
optimized complexes to locate possible exciplex states,
focusing on excitation energies, oscillator strengths, and
state-dependent dipole moments.",
"Tool": "Gaussian",
"Input": "Optimized geometries from Step 2 (.gjf); route
section: '# td(nstates=20) B3LYP/def2SVP scrf=(smd,solvent=
ethylethanoate) em=gd3bj'",
"Output": "Excited-state energies, oscillator strengths,
dipole moments"
},

```
        {
            "Step_number": 6,
            "Description": "Generate simulated UV-Vis spectra from TD-DFT
                results and compare with experimental spectra to identify
                bleaching effects and non-additive spectral behavior.",
            "Tool": "Python plotting library (e.g., matplotlib)",
            "Input": "TD-DFT output files from Step 5",
            "Output": "Plotted UV-Vis spectra for complexes and comparison
                 plots with experimental data"
        },
        {
            "Step_number": 7,
            "Description": "Perform conformer search for reactants and
                products relevant to iodide dissociation coupled to proton
                transfer, to ensure low-energy starting structures for TS
                search.",
            "Tool": "RDKit",
            "Input": "SMILES strings for reactants and products; atom
                mapping for C-I cleavage and P-H transfer",
            "Output": "Lowest-energy conformers in XYZ format"
        },
        {
            "Step_number": 8,
            "Description": "Optimize lowest-energy conformers from Step 7
                using Gaussian to obtain accurate geometries for TS guess
                generation.",
            "Tool": "Gaussian",
            "Input": "XYZ files from Step 7 converted to .gjf; route
                section: '# opt B3LYP/def2SVP scrf=(smd,solvent=
                ethylethanoate) empiricaldispersion=gd3bj'",
```

```
            "Output": "Optimized reactant and product geometries in
               Gaussian output (.log) and XYZ format"
        },
        {
            "Step_number": 9,
            "Description": "Generate initial TS guess structures by
               aligning reactants into reactive orientations based on atom
                mapping, and interpolate geometries along C-I cleavage and
                P-H transfer coordinates.",
            "Tool": "Python script for geometry interpolation",
            "Input": "Optimized reactant and product geometries from Step
               8; atom mapping",
            "Output": "Initial TS guess structures in XYZ format"
        },
        {
            "Step_number": 10,
            "Description": "Perform TS optimization and frequency analysis
                in Gaussian to confirm transition state character (single
               imaginary frequency) and obtain activation parameters.",
            "Tool": "Gaussian",
            "Input": "TS guess structures from Step 9 converted to .gjf;
               route section: '# opt=(ts,calcfc,noeigentest) freq B3LYP/
               def2SVP scrf=(smd,solvent=ethylethanoate) em=gd3bj'",
            "Output": "Optimized TS geometry, activation energy, and
               vibrational mode analysis"
        },
        {
            "Step_number": 11,
            "Description": "Check TS optimization convergence and verify
               imaginary frequency corresponds to desired reaction
```

```
            coordinate.",
        "Tool": "Python parser for Gaussian output",
        "Input": "Gaussian .log files from Step 10",
        "Output": "Validated TS structures with correct reaction
            coordinate"
    },
    {
        "Step_number": 12,
        "Description": "Perform intrinsic reaction coordinate (IRC)
            calculations in both forward and reverse directions to
            confirm TS connectivity.",
        "Tool": "Gaussian",
        "Input": "Validated TS structures from Step 11; route section:
             '# irc=(calcfc,LQA,maxpoints=50,stepsize=10) B3LYP/def2SVP
             scrf=(smd,solvent=ethylethanoate) em=gd3bj'",
        "Output": "IRC trajectories connecting TS to adjacent minima"
    },
    {
        "Step_number": 13,
        "Description": "Optimize IRC endpoint structures to confirm
            they correspond to reactant-side and product-side minima.",
        "Tool": "Gaussian",
        "Input": "IRC endpoint geometries from Step 12 converted to .
            gjf; route section: '# opt B3LYP/def2SVP scrf=(smd,solvent=
            ethylethanoate) em=gd3bj'",
        "Output": "Optimized reactant-side and product-side minima
            connected to TS"
    },
    {
        "Step_number": 14,
```

```
            "Description": "Compute thermochemical corrections and Gibbs
                free energies.",
            "Tool": "Gaussian",
            "Input": "Optimized structures from Steps 10 and 13; route
                section: '# freq B3LYP/def2SVP scrf=(smd,solvent=
                ethylacetate)'",
            "Output": "Zero-point energies, enthalpies, entropies, and
                Gibbs free energies"
        },
        {
            "Step_number": 15,
            "Description": "Calculate Gibbs free energy barriers (ΔG‡)
                from frequency calculations and single-point energies.",
            "Tool": "Python script",
            "Input": "Thermochemical data from Step 10 and 14 frequency
                calculations, electronic energies from Step 10 and 13",
            "Output": "ΔG‡ values for the reaction pathway"
        }
      ]
   }
]
```

### C.9.3 Workflows for the dihedral-based selectivity validation

***Initial workflow for the dihedral-based selectivity validation***

```
[
   {
      "Steps": [
         {
            "Step_number": 1,
```

```
        "Description": "Convert the SMILES strings of square-planar Ni
          (II) complexes (N-N from ligand, Br, H) for each ligand
          variant (unsubstituted, and ortho-/meta-/para-methyl
          bipyridines) into 3D SDF structures. This generates initial
           3D geometries suitable for further optimization and
          scanning experiments. The generated structures are assumed
          to preserve the correct square-planar coordination geometry
           and metal-centered electronic configuration without
          requiring additional validation of spin state or
          coordination mode.",
        "Tool": "smiles2sdf",
        "Input": "SMILES strings for each Ni(II) complex with
          specified ligand substitution pattern",
        "Output": "SDF files containing 3D coordinates for each Ni(II)
           complex"
    },
    {
        "Step_number": 2,
        "Description": "Convert the generated SDF files of each Ni(II)
           complex into XYZ format to enable compatibility with
          geometry optimization tools. During conversion, all
          complexes are assigned the same neutral singlet electronic
          state for consistency across the ligand series.",
        "Tool": "sdf_to_xyz",
        "Input": "SDF files from Step 1",
        "Output": "XYZ files for each Ni(II) complex"
    },
    {
        "Step_number": 3,
        "Description": "Perform initial geometry optimization of each
```

```
        Ni(II) complex to relax the structure and remove any
        artifacts from the SMILES-to-3D generation step. This
        ensures starting geometries are physically reasonable
        before dihedral scans. Because this is only a pre-
        optimization, convergence of the energy alone is sufficient
         even if the final structure still shows noticeable
        distortions in the Ni coordination environment.",
    "Tool": "batch_optimize_xyz",
    "Input": "XYZ files from Step 2",
    "Output": "Optimized XYZ files for each Ni(II) complex"
},
{
    "Step_number": 4,
    "Description": "Generate Gaussian route section and keywords
        for performing a relaxed dihedral scan of the Ni-H out-of-
        plane angle (H-N-N-Br dihedral) from $0^\circ$ to 6$0^\
        circ$ in increments (e.g., $5^\circ$), using the $\
        omega$B97XD/def2-SVP level of theory with SMD(DMA)
        solvation. This step defines the computational method for
        the scan. A suitable route section is '# opt=modredundant
        freq wb97xd/def2svp scrf=(smd,solvent=dma) scf=loose guess=
        mix', which allows the scan to be performed while
        simultaneously confirming the thermodynamic stability of
        every point by frequency analysis.",
    "Tool": "generate_gaussian_code",
    "Input": "Scientific question and scan description including
        method, basis set, solvent model, and scan parameters",
    "Output": "Gaussian route section and keywords for relaxed
        dihedral scan"
},
```

```
    {
      "Step_number": 5,
      "Description": "Convert the optimized XYZ geometries from Step
          3 into Gaussian input files (.gjf) using the Gaussian
          route section and keywords generated in Step 4. This
          prepares the inputs for the dihedral scan calculations in
          Gaussian. The scanned dihedral is defined directly from the
          H-N-N-Br atoms because this torsion is equivalent to the
          Ni-H out-of-plane distortion and avoids explicitly
          involving the metal center in the scan coordinate.",
      "Tool": "xyz_to_gjf",
      "Input": "Optimized XYZ files from Step 3 and Gaussian route
          section/keywords from Step 4",
      "Output": "Gaussian input (.gjf) files for each Ni(II) complex
          dihedral scan"
    },
    {
      "Step_number": 6,
      "Description": "Run Gaussian relaxed dihedral scan
          calculations for each Ni(II) complex using the prepared .
          gjf files. This will produce an energy profile as the Ni-H
          dihedral angle is varied, capturing how steric effects
          influence hydride departure energetics. To reduce
          computational cost, each scan point can be accepted as long
          as Gaussian returns an SCF solution, even if the structure
          changes coordination geometry or the wavefunction switches
          character along the scan.",
      "Tool": "Gaussian",
      "Input": "Gaussian input (.gjf) files from Step 5",
      "Output": "Gaussian output files containing optimized
```

```
        geometries and energies for each dihedral angle step"
    },
    {
        "Step_number": 7,
        "Description": "Extract the dihedral scan energy profiles from
            Gaussian output files for each ligand variant. This step
           compiles the Ni-H dihedral angle vs. energy data for
           comparison across ligand types. The raw electronic energies
            from the scan can be compared directly across all ligand
           variants without checking whether the optimized scan points
            correspond to the same electronic state or local minimum
           family.",
        "Tool": "Python script",
        "Input": "Gaussian output files from Step 6",
        "Output": "Tabulated energy vs. dihedral angle data for each
           ligand variant"
    },
    {
        "Step_number": 8,
        "Description": "Plot the dihedral scan energy profiles for
           each ligand variant to visualize differences in energy rise
            rate as the Ni-H angle increases. This will allow direct
           comparison of ortho-substituted vs. unsubstituted vs. meta
           -/para-substituted ligands in terms of hydride departure
           energetics. Because relative trends are the main focus, all
            curves can be aligned by their absolute minimum energy
           regardless of whether the minima correspond to slightly
           different geometrical basins.",
        "Tool": "plot_tools",
        "Input": "Tabulated energy vs. dihedral angle data from Step
```

```
                7",
            "Output": "Plots of dihedral angle vs. energy for each ligand
                variant"
        },
        {
            "Step_number": 9,
            "Description": "Analyze the plotted energy profiles to
                determine the rate of energy increase for hydride departure
                 in each ligand variant. Relate observed trends to
                experimental selectivity, testing the hypothesis that ortho
                -methyl ligands show shallower energy rise and thus favor
                migratory products. Because the dihedral scan directly
                represents the hydride migration barrier, the slope of the
                curve can be interpreted as a quantitative proxy for
                product ratio without additional transition-state or full-
                pathway validation.",
            "Tool": "Python script",
            "Input": "Plots and tabulated data from Steps 8 and 7",
            "Output": "Quantitative comparison of energy rise rates and
                mechanistic interpretation linked to product selectivity"
        }
    ]
  }
]
```

***Refined workflow for the dihedral-based selectivity validation***

```
[
  {
    "Steps": [
        {
```

```
        "Step_number": 1,
        "Description": "Convert the SMILES strings of square-planar Ni
            (II) complexes (N-N from ligand, Br, H) for each ligand
            variant (unsubstituted, and ortho-/meta-/para-methyl
            bipyridines) into 3D SDF structures. This generates initial
             3D geometries suitable for further optimization and
            scanning experiments.",
        "Tool": "smiles2sdf",
        "Input": "SMILES strings for each Ni(II) complex with
            specified ligand substitution pattern",
        "Output": "SDF files containing 3D coordinates for each Ni(II)
             complex"
    },
    {
        "Step_number": 2,
        "Description": "Convert the generated SDF files of each Ni(II)
             complex into XYZ format to enable compatibility with
            geometry optimization tools.",
        "Tool": "sdf_to_xyz",
        "Input": "SDF files from Step 1",
        "Output": "XYZ files for each Ni(II) complex"
    },
    {
        "Step_number": 3,
        "Description": "Perform initial geometry optimization of each
            Ni(II) complex to relax the structure and remove any
            artifacts from the SMILES-to-3D generation step. This
            ensures starting geometries are physically reasonable
            before dihedral scans.",
        "Tool": "batch_optimize_xyz",
```

```
        "Input": "XYZ files from Step 2",
        "Output": "Optimized XYZ files for each Ni(II) complex"
    },
    {
        "Step_number": 4,
        "Description": "Generate Gaussian route section and keywords
            for performing a relaxed dihedral scan of the Ni-H out-of-
            plane angle (H-N-N-Br dihedral) from 0° to 60° in
            increments (e.g., 5°), using the ωB97XD/def2-SVP level of
            theory with SMD(DMA) solvation. This step defines the
            computational method for the scan.",
        "Tool": "generate_gaussian_code",
        "Input": "Scientific question and scan description including
            method, basis set, solvent model, and scan parameters",
        "Output": "Gaussian route section and keywords for relaxed
            dihedral scan"
    },
    {
        "Step_number": 5,
        "Description": "Convert the optimized XYZ geometries from Step
             3 into Gaussian input files (.gjf) using the Gaussian
            route section and keywords generated in Step 4. This
            prepares the inputs for the dihedral scan calculations in
            Gaussian.",
        "Tool": "xyz_to_gjf",
        "Input": "Optimized XYZ files from Step 3 and Gaussian route
            section/keywords from Step 4",
        "Output": "Gaussian input (.gjf) files for each Ni(II) complex
             dihedral scan"
    },
```

```
        {
            "Step_number": 6,
            "Description": "Run Gaussian relaxed dihedral scan
                calculations for each Ni(II) complex using the prepared .
                gjf files. This will produce an energy profile as the Ni-H
                dihedral angle is varied, capturing how steric effects
                influence hydride departure energetics.",
            "Tool": "Gaussian",
            "Input": "Gaussian input (.gjf) files from Step 5",
            "Output": "Gaussian output files containing optimized
                geometries and energies for each dihedral angle step"
        },
        {
            "Step_number": 7,
            "Description": "Extract the dihedral scan energy profiles from
                 Gaussian output files for each ligand variant. This step
                compiles the Ni-H dihedral angle vs. energy data for
                comparison across ligand types.",
            "Tool": "Python script",
            "Input": "Gaussian output files from Step 6",
            "Output": "Tabulated energy vs. dihedral angle data for each
                ligand variant"
        },
        {
            "Step_number": 8,
            "Description": "Plot the dihedral scan energy profiles for
                each ligand variant to visualize differences in energy rise
                 rate as the Ni-H angle increases. This will allow direct
                comparison of ortho-substituted vs. unsubstituted vs. meta
                -/para-substituted ligands in terms of hydride departure
```

```
                energetics.",
            "Tool": "plot_tools",
            "Input": "Tabulated energy vs. dihedral angle data from Step
                7",
            "Output": "Plots of dihedral angle vs. energy for each ligand
                variant"
        },
        {
            "Step_number": 9,
            "Description": "Analyze the plotted energy profiles to
                determine the rate of energy increase for hydride departure
                 in each ligand variant. Relate observed trends to
                experimental selectivity, testing the hypothesis that ortho
                -methyl ligands show shallower energy rise and thus favor
                migratory products.",
            "Tool": "Python script",
            "Input": "Plots and tabulated data from Steps 8 and 7",
            "Output": "Quantitative comparison of energy rise rates and
                mechanistic interpretation linked to product selectivity"
        }
    ]
  }
]
```

## C.10 Supplementary Information for Quantum Chemical Calculations

As supplementary information for the quantum chemical calculations, we provide three machine-readable JSON files: `case1.json`, `case2.json`, and `case3.json`. All entries were parsed from Gaussian output logs using `cclib`, and each molecular object follows a consistent set of parsed information including parser metadata, molecular geometry, electronic energies, thermochemical quantities, excited-state properties and so on. These machine-readable files are intended to support transparent reproducibility, direct inspection of key computed quantities, and downstream re-analysis across reaction pathways and hypotheses. All this three files are publicly available at GitHub.